\documentclass{article} 
\usepackage[svgnames,usenames,dvipsnames,table]{xcolor}         

\usepackage{amsmath,amsfonts,bm}

\def\eqref#1{equation~\ref{#1}}

\def\1{\bm{1}}

\DeclareMathAlphabet{\mathsfit}{\encodingdefault}{\sfdefault}{m}{sl}
\SetMathAlphabet{\mathsfit}{bold}{\encodingdefault}{\sfdefault}{bx}{n}

\usepackage{booktabs}       
\usepackage[hyphens]{url}
\usepackage[colorlinks=true, citecolor=Navy, breaklinks=true]{hyperref}       
\usepackage{nicefrac}       
\usepackage{caption}
\usepackage{subcaption}
\usepackage{graphicx}
\usepackage{wrapfig}
\usepackage[most]{tcolorbox}
\usepackage{multirow}
\usepackage[capitalize,noabbrev]{cleveref}
\usepackage[inline]{enumitem}
\usepackage{mathtools}
\usepackage{amsthm}
\usepackage{pifont}
\usepackage{cereb}

\crefname{figure}{Fig.}{Figs.}
\Crefname{figure}{Fig.}{Figs.}
\crefname{equation}{Eq.}{Eqs.}
\Crefname{equation}{Eq.}{Eqs.}
\crefname{section}{Sec.}{Secs.}
\Crefname{section}{Sec.}{Secs.}
\crefname{subsection}{Sec.}{Secs.}
\Crefname{subsection}{Sec.}{Secs.}
\crefname{subsubsection}{Sec.}{Secs.}
\Crefname{subsubsection}{Sec.}{Secs.}

\theoremstyle{definition}
\newtheorem{definition}{Definition}

\definecolor{taborange}{RGB}{235,115,12}
\definecolor{tabblue}{RGB}{31,119,180}
\definecolor{tabgreen}{RGB}{44,160,44}
\definecolor{tabred}{RGB}{214,39,40}

\usepackage{tikz}
\usepackage{pgfplots}
\usetikzlibrary{arrows.meta,shapes.geometric}

\definecolor{cborange}{RGB}{255,127,0} 
\definecolor{cbblue}{RGB}{31,119,180}  

\newcommand{\twofighspace}{4mm}

\usepackage{enumitem}
\setlist[itemize]{left=0.5em}
\setlist[enumerate]{left=0.5em}

\newcommand{\batcht}{\mathcal{B}_t}
\newcommand{\thetaend}{\theta_T}
\newcommand{\reeval}{\mathcal{L}_{\mathrm{re}}}
\newcommand{\barreeval}{\mu_{\reeval}}
\newcommand{\hatreeval}{\hat{\mathcal{L}}_{\mathrm{re}}}
\newcommand{\barhatreeval}{\mu_{\hatreeval}}
\newcommand{\Dprime}{\mathcal{D}_{\mathrm{Orig}}}
\newcommand{\Dhigh}{\mathcal{D}_{\mathrm{HQ}}}
\newcommand{\pearsonr}{r_{\mathrm{p}}}
\newcommand{\Rtwo}{R^{2}}
\newcommand{\tpp}{\mathrm{TPP}}
\newcommand{\mstar}{m^{\ast}}
\newcommand{\mstarfit}{\mstar = C \cdot (\tpp)^{\mu_1} \cdot (\tema)^{\mu_2}}
\newcommand{\mycheck}{\text{\textcolor{ForestGreen}{\ding{51}}}}
\newcommand{\myx}{\text{\textcolor{BrickRed}{\ding{55}}}}
\newcommand{\bottomleftfig}{\text{\emph{bottom left}}}
\newcommand{\leftfig}{\text{\emph{left}}}
\newcommand{\middlefig}{\text{\emph{middle}}}
\newcommand{\rightfig}{\text{\emph{right}}}
\newcommand{\Leftfig}{\text{\emph{Left}}}
\newcommand{\Middlefig}{\text{\emph{Middle}}}
\newcommand{\Rightfig}{\text{\emph{Right}}}

\newcommand{\tenx}{\mbox{10}\times}
\newcommand{\dtoz}{\mbox{D2Z}}
\newcommand{\mup}{\mbox{$\mu$P}}
\newcommand{\titer}{\tau_\mathrm{iter}}
\newcommand{\tepochwang}{\tau_\mathrm{epoch}}
\newcommand{\tepoch}{\tau}
\newcommand{\tema}{\tepoch}  
\newcommand{\constant}{\mbox{\emph{Constant}}}
\newcommand{\linear}{\mbox{\emph{Linear}}}
\newcommand{\cosine}{\mbox{\emph{Cosine}}}
\newcommand{\step}{\mbox{\emph{Step}}}
\newcommand{\wsd}{\mbox{\emph{WSD}}}
\newcommand{\cyclic}{\mbox{\emph{Cyclic}}}
\newcommand{\dmodel}{d_{\mathrm{model}}}
\newcommand{\dproxy}{d_{\mathrm{proxy}}}
\newcommand{\nlayers}{n_{\mathrm{layers}}}
\newcommand{\dffn}{d_{\mathrm{ffn}}}
\newcommand{\dhead}{d_{\mathrm{head}}}
\newcommand{\hateta}{\tilde{\eta}}
\newcommand{\trainfrac}{\hat{t}}
\newcommand{\cbs}{B_{\mathrm{crit}}}
\newcommand{\bcrit}{\cbs}  
\newcommand{\nopt}{N_{\text{opt}}}
\newcommand{\dopt}{D_{\text{opt}}}

\newcommand{\maxlrdetail}{1.62\mbox{e-}02}

\tcbset{textmarker/.style={%
  enhanced,
  parbox=false,boxrule=0mm,boxsep=0mm,arc=0mm,
  outer arc=0mm,left=6mm,right=3mm,top=4pt,bottom=3pt,
  toptitle=1mm,bottomtitle=1mm,oversize,
  left skip=0mm,
  right skip=8mm
}}

\newcounter{fcounter}
\crefname{fcounter}{Finding}{Findings}

\newcounter{kcounter}
\newcommand\takeaway[1]{
        \refstepcounter{kcounter}\vspace{2pt}
        \begin{tcolorbox}[colback=green!10!white,colframe=green!80!black,boxsep=1pt,left=2pt,right=2pt,top=1pt,bottom=1pt]\noindent{\textbf{\sffamily Key takeaway \arabic{kcounter}}: \sffamily #1}
        \end{tcolorbox}\vspace{0pt}
}
\crefname{kcounter}{Takeaway}{Takeaways}

\newtcolorbox{hypothesisBox}{textmarker,
    borderline west={6pt}{0pt}{blue},
    colback=blue!10!white}
\newcounter{hcounter}
\newcommand\hypothesis[1]{
        \refstepcounter{hcounter}\vspace{1.0pt}
        \begin{hypothesisBox}\noindent{\textbf{\sffamily Hypothesis \arabic{hcounter}}: \sffamily #1}
        \end{hypothesisBox}\vspace{-0.5pt}
}
\crefname{hcounter}{Hypothesis}{Hypotheses}

\renewcommand{\cite}[1]{\PackageError{MyPackage}{Do not use \string\cite\space with natbib. Use \string\citet\space or \string\citep}{See the natbib package documentation for explanation.}}

\makeatletter
\AtBeginDocument{%
  \@ifpackageloaded{cleveref}{%

    \providecommand\cref@appendix@setup{%
      \crefname{appendix}{Appendix}{Appendices}%
      \Crefname{appendix}{Appendix}{Appendices}%
    }%

    \AddToHook{cmd/appendix/before}{%
      \cref@appendix@setup
      \@ifundefined{chapter}{%
        \crefalias{section}{appendix}%
      }{%
        \crefalias{chapter}{appendix}%
      }%
      \crefalias{subsection}{appendix}%
      \crefalias{subsubsection}{appendix}%
    }%

  }{}
}
\makeatother

\title{Predicting Training Re-evaluation Curves Enables Effective Data Curriculums for LLMs}

\author{Shane Bergsma, Nolan Dey \& Joel Hestness \\
  Cerebras Systems \\
  \texttt{\{shane.bergsma,nolan,joel\}@cerebras.net}
}

\begin{document}
\maketitle

\begin{abstract}
Data curriculums have become central to successful LLM training, yet
principles governing optimal data placement remain unclear. We
introduce the \emph{training re-evaluation curve (TREC)}, a diagnostic
that retrospectively evaluates training batches \emph{using the final
model weights}. The TREC characterizes how well a trained model
retains training data as a function of \emph{when} the data was
encountered during training.
Analyzing TRECs for models from 111M to 3.9B parameters, we show that
placing high-quality data at low points on the TREC significantly
improves performance. Importantly, while a TREC is initially
observable only after training, we demonstrate it can
be \emph{predicted in advance} from AdamW's implicit EMA coefficients,
enabling proactive curriculum design.
By predicting TRECs for published training recipes, we explain prior
ablations and reveal suboptimal data placements.
We also align high-quality data with TREC minima in order to improve
continual pre-training of a 3.9B-parameter LLM trained on 900B tokens.
All experiments were run on Cerebras CS-3 systems.

\end{abstract}

\begin{figure}[ht]
  \centering
  \begin{minipage}{0.33\textwidth}
    \includegraphics[trim={0.3cm 0.42cm 0.264cm 0.3cm}, clip, width=\linewidth]{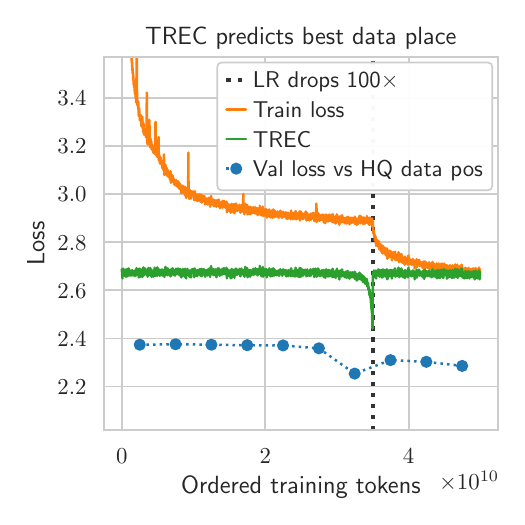}
  \end{minipage}\hfill
  \begin{minipage}{0.33\textwidth}
    \includegraphics[trim={0.3cm 0.42cm 0.264cm 0.3cm}, clip, width=\linewidth]{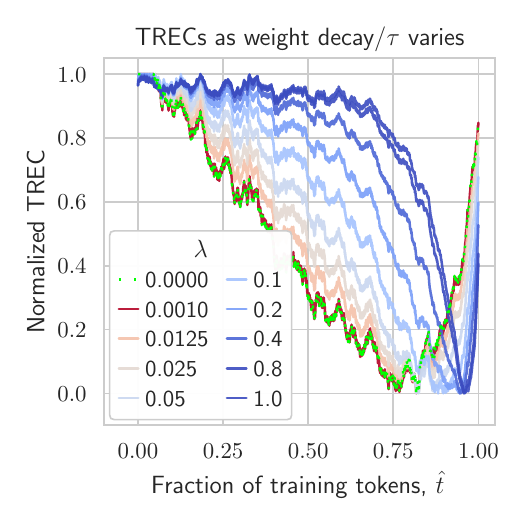}
  \end{minipage}\hfill
  \begin{minipage}{0.33\textwidth}
    \includegraphics[trim={0.3cm 0.42cm 0.264cm 0.3cm}, clip, width=\linewidth]{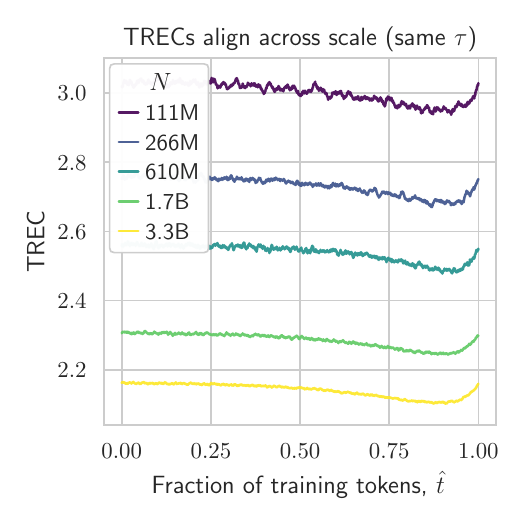}
  \end{minipage}
\caption{$\Leftfig$: (610M parameters): \textcolor{taborange}{train
    loss} steadily falls \emph{after} learning rate drops to
  $0.01\eta_{\max}$ (after 70\% of steps), but the optimal position
  for placing high-quality (HQ) data is \emph{before} the LR drop, in
  the \textcolor{tabgreen}{TREC valley}.
  $\Middlefig$: (610M, linear LR decay): TREC shape varies with AdamW
  timescale $\tema$ (varied via weight decay $\lambda$).
  $\Rightfig$: (size varies, linear LR decay-to-zero, all 20
  tokens-per-parameter): TRECs align across 1000$\times$ scaling of
  training compute, when $\tema$ matches.}
\label{fig:hook}
\end{figure}


\section{Introduction}\label{sec:intro}
LLM training now often includes \emph{mid-training}
or \emph{annealing} phases where special data is upsampled in the
final stages of pre-training. These data can be
\emph{high-quality}~\citep{shen2024jetmoe}, \emph{recent}~\citep{dubey2024llama}
or \emph{domain-specific}---e.g., math~\citep{olmo2024},
code~\citep{zhang2024map} or instructions~\citep{hu2024minicpm}. It is
assumed that placing such data \emph{at the end of training}, when
learning rate is near zero, maximizes its effectiveness. While some
work explores when this phase should
begin~\citep{feng2024maximize,parmar2024reuse,liu2025midtraining},
``many interesting questions remain around finding the optimal dataset
distribution for pre-training''~\citep{team2023gemini}.

We introduce the \emph{training re-evaluation curve (TREC)}, a
diagnostic for understanding LLM training and data placement. A TREC
measures how well a fully trained model performs on training batches
as a function of when those batches were seen. Defined over
homogeneous data, we construct it by evaluating the final model on the
i.i.d.\ training sequence, in order. If data were retained equally,
the TREC would be flat; in practice, models perform better on tokens
from specific points in training.

Under $\step$-decay learning rate (LR) schedules, e.g., as used (with
multi-step-decay) in DeepSeek LLM~\citep{bi2024deepseek},
the \textcolor{tabgreen}{TREC} may bottom-out well before the end of
training (\cref{fig:hook}, $\leftfig$), while
the \textcolor{taborange}{standard training curve} (on unseen batches)
continues to decline.
We hypothesize that high-quality (HQ) data should be placed where the
TREC is lowest---that is, where the final model would achieve its
lowest retrospective loss.
Indeed, when retraining different models, each with the same HQ data
inserted into a different 10\% segment of the training trajectory,
placing HQ data where the TREC is lowest yields the best validation
loss (\emph{loss-by-segment} plotted as \textcolor{tabblue}{blue
points} in \cref{fig:hook}, $\leftfig$).

Of course, it is impractical to train a very large model, measure its
TREC, and then \emph{re}-train with a TREC-informed
curriculum. Fortunately, we show the TREC is \emph{predictable},
enabling \emph{proactive} curriculum design.
For AdamW---the dominant optimizer in LLM pre-training---TRECs are
governed by the EMA timescale (\cref{sec:ema}): AdamW parameters can
be viewed as an EMA over weight updates (over data), so timescale
$\tema$ controls data influence across training, and thus where TREC
performance peaks (\cref{fig:hook}, $\middlefig$).
Sweeping learning rate, weight decay, or batch size with matching
timescales yields identical TREC shapes (\cref{fig:tema}).
Shape also persists across scale: despite 1000$\times$ more training
FLOPs, a 3.3B model matches a 111M model's shape (\cref{fig:hook},
$\rightfig$).
In \cref{sec:predict}, we formalize a predictive model of TRECs based
on an expanded view of the AdamW timescale---one that handles
arbitrary LR schedules, including $\step$ drops.

Based on our work, practitioners can use TRECs to guide data ordering,
avoiding flawed heuristics and costly ablations.
More specifically, our main contributions are:
\begin{itemize}
\item Introducing train re-evaluation curves, showing they \textbf{predict optimal
      data placement} for a given optimizer configuration
      (\cref{sec:placements}).
\item A large-scale study of 600 TRECs in models from
      111M to 3.9B parameters, and datasets from 20 to 1280
      tokens-per-parameter (TPP).\@ The study connects TRECs to the
      AdamW timescale (\cref{sec:ema}), and enables an analytical
      model for \textbf{predicting curves in advance of training}
      (\cref{sec:predict}).
\item Explaining findings in \textbf{sparse mixture-of-experts}
      (\cref{sec:moe}) and \textbf{prior data curriculum work}
      (\cref{sec:prior}).  For example, TREC prediction can explain
      why Llama-3~405B did not benefit, on GSM8k validation, from
      annealing on the GSM8k training set.
\item Leveraging TRECs to improve a
      \textbf{3.9B-parameter LLM} trained on 900B tokens
      (\cref{sec:demo}).
\end{itemize}

\section{TRECs predict effective data placement}\label{sec:placements}
We now formally define TRECs, state our key hypothesis, and describe
its evaluation.

\begin{definition}[TREC]
Let \( \mathcal{B}_1, \dots, \mathcal{B}_T \) denote the sequence of
batches used during training, drawn independently and identically from
data distribution \( \mathcal{D} \), and let \( \thetaend \) represent
the fully-trained model parameters. The \emph{training re-evaluation
curve (TREC)} is the sequence of scalar loss values: $\reeval(t) :=
\mathcal{L}(\mathcal{B}_t; \thetaend), \text{for } t = 1, \dots, T$,
where $\mathcal{L}(\mathcal{B}_t; \theta_T)$ denotes the loss (e.g.,
cross-entropy) evaluated on training batch $\mathcal{B}_t$ from step
$t$ using final parameters $\thetaend$. Intuitively, lower TREC loss
suggests greater alignment with $\thetaend$, and may indicate
$\mathcal{B}_t$ contributed more significantly to the final model.
\end{definition}

It is worth clarifying why re-evaluation loss may depend on order,
even when training batches are drawn i.i.d.\@ In any online
optimization process, each batch is encountered at a different point
along the parameter trajectory, with a different amount of subsequent
parameter evolution. Early batches are learned when the parameters are
far from their final values, and updates derived from these batches
lose their effect as the parameters subsequently evolve. The result is
higher re-evaluation loss on these earlier batches: the model performs
on these forgotten data similarly to how it performs on unseen
validation samples. This is related to catastrophic forgetting in
continual
learning~\citep{mccloskey1989catastrophic,kirkpatrick2017overcoming}:
earlier examples can be forgotten, even when the training distribution
does not change. In \cref{sec:theory}, we formalize this intuition and
show that TREC loss can be analytically predicted as a function of
parameter evolution under a simplified quadratic model.

\begin{definition}[High-Quality Data]
\label{def:hq_data}
Given training distribution $\Dprime$, a distribution $\Dhigh$ is
\emph{high-quality for a task} if replacing $\Dprime$ samples with
$\Dhigh$ samples improves task performance.
\end{definition}

The canonical TREC is defined under homogeneous i.i.d.\ sampling from
a base distribution $\mathcal{D}$.  We average across batches when
plotting TRECs in order to isolate the effect of training position
from any variation in intrinsic batch difficulty.
We hypothesize the resulting $\reeval(t)$ primarily reflects
optimization dynamics, rather than the specific $\mathcal{D}$ used to
compute it.  That is, given a fixed optimizer configuration, a TREC
characterizes how strongly data encountered at each \emph{training
position} is retained by the final model---regardless of what data is
actually inserted at those positions, or even if different data (e.g.,
high-quality data) is used heterogeneously across positions.  Our
belief in the effectiveness of this positional ranking leads to our
main hypothesis:

\hypothesis{Placing a fixed number of high-quality samples at steps
  where \( \reeval(t) \) is lowest maximizes performance on a target
  task.\label{hyp:placements}}

\paragraph{Experimental setup.}

\begin{table}[t]
\centering
\small
\renewcommand{\arraystretch}{1.15}
\begin{tabular}{p{0.385\linewidth} p{0.36\linewidth} p{0.165\linewidth}}
\toprule
{Schedule} & {Post-warmup decay $\eta(t)$} & {Used in} \\
\midrule

Step drop at 70\% of training ($\step$ decay) &
$\eta(t)=
\begin{cases}
\eta_{\max}, & t<0.7T \\
0.01\,\eta_{\max}, & t\ge0.7T
\end{cases}$ &
\cref{fig:hook} ($\leftfig$), \cref{fig:placements} ($\leftfig$) \\

Linear decay to $0.1\eta_{\max}$ ($\tenx$ decay) &
$\eta(t)=0.1\eta_{\max}
+0.9\eta_{\max}\!\left(1-\frac{t-w}{T-w}\right)$ &
\cref{fig:placements} ($\rightfig$) \\

Linear decay to zero ($\dtoz$ decay) &
$\eta(t)=\eta_{\max}
\!\left(1-\frac{t-w}{T-w}\right)$ &
\cref{fig:d2zplacements} ($\leftfig$) \\

\bottomrule
\end{tabular}
\caption{Learning-rate schedules used in placement tests. All
  schedules share linear warmup over $w=0.1T$ steps.  Appendix
  \cref{fig:d2zplacements} (right) visualizes the LR curves.}
\label{tab:lr_schedules}
\end{table}

\begin{figure}[ht]
  \centering
  \begin{minipage}[t]{0.33\textwidth}
    \vspace{0pt} 
    \centering
    \captionof{table}{SlimPajama mixes used in \cref{fig:placements} placement tests:
      \emph{General Blend} is the original distribution, \emph{Code Blend} is the HQ data.}
    \label{tab:blend}
    \small
\begin{tabular}{@{}cccc@{}}
\toprule
SlimPJ          & General     & Code                \\
Subset          & Blend       & Blend               \\ \midrule
Commcrawl       & 55.1        & 37.1                \\
C4              & 28.2        & 19.0                \\
GitHub          & 0.0         & ~~30.0$^{\uparrow}$                \\
Books           & 4.4         & 3.0                 \\
ArXiv           & 4.9         & 3.3                 \\
Wikipedia       & 4.0         & 2.7                 \\
StackExch.      & 3.5         & ~~5.0$^{\uparrow}$                 \\ \bottomrule
\end{tabular}

  \end{minipage}\hfill
  \begin{minipage}[t]{0.66\textwidth}
    \vspace{0pt} 
    \centering
    \begin{minipage}[t]{0.49\textwidth}
      \includegraphics[trim={0.3cm 0.4cm 0.264cm 0.3cm}, clip, width=\linewidth]{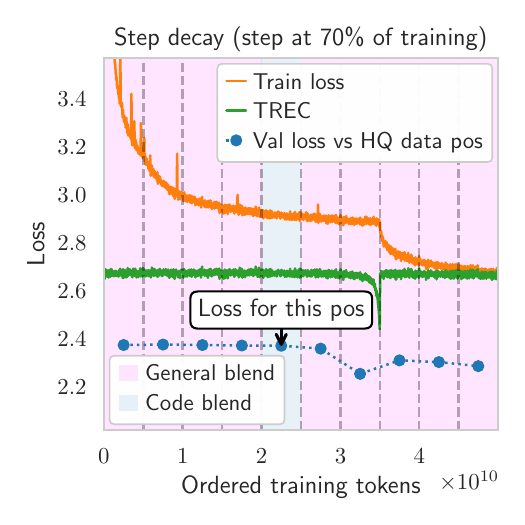}
    \end{minipage}\hfill
    \begin{minipage}[t]{0.49\textwidth}
      \includegraphics[trim={0.3cm 0.4cm 0.264cm 0.3cm}, clip, width=\linewidth]{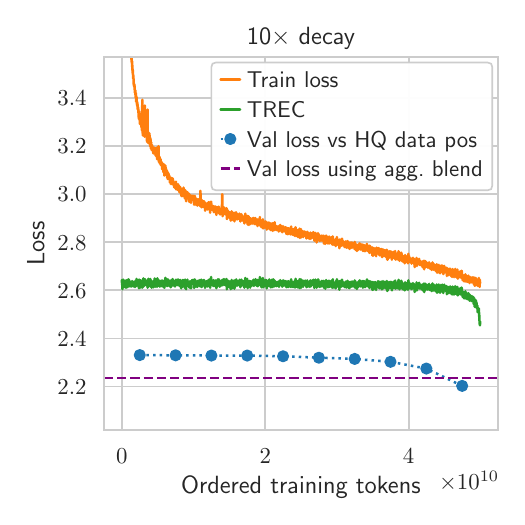}
    \end{minipage}
    \caption{\textbf{TRECs predict best placement.} $\Leftfig$:
      Example placement curriculum 5-of-10 for the $\step$ decay LR
      schedule. $\Rightfig$: Results for linear decay to
      $0.1\eta_{\max}$.
      \label{fig:placements}}
  \end{minipage}
\end{figure}

Experiments use 610M-parameter GPT2-style LLMs trained with AdamW,
$\mup$~\citep{yang2022mup}, and learning rate warmup over 10\% of
steps.
After warmup, LR follows one of the decay schedules in
Table~\ref{tab:lr_schedules}.
Full architecture and tuning details are in
\cref{sec:experimental_details}.
Models are trained on SlimPajama splits~\citep{cerebras2023slimpajama}
(blend weights in \cref{tab:blend}), using the \emph{general blend}
(GB) as $\Dprime$ and \emph{code blend} (CB) as $\Dhigh$, with CB
validation loss as the target task. We use 45B (90\%) GB tokens and 5B
(10\%) CB tokens.  We also construct an \emph{aggregate blend} (AB)
with a uniform 90/10 GB/CB mix. All models train on 50B tokens total
(82 TPP) with no repetition. Train and TREC loss curves are computed
on homogeneous AB batches.

To evaluate \cref{hyp:placements}, we conduct a data placement sweep:
for each LR schedule, we train 10 models, each with the 5B CB tokens
inserted into a different 10\% segment of the training data. An
example curriculum is shown in \cref{fig:placements}, $\leftfig$.
This figure also plots the standard \textcolor{taborange}{training
  curve} (loss on unseen batches during training). While the
\emph{train loss} shows a familiar decrease with LR
annealing~\citep{tissue2024scaling,schaipp2025surprising}, note the
\textcolor{tabgreen}{TREC} \emph{increases} after the LR drops.

\paragraph{Results.}

Results in \cref{fig:hook} ($\leftfig$, $\step$ decay schedule),
\cref{fig:placements} ($\rightfig$, linear decay to $0.1\eta_{\max}$),
and appendix \cref{fig:d2zplacements} (linear decay to zero) show, in
all cases, placing CB data at the step with lowest $\reeval(t)$ yields
the best validation loss.  Data placement also outperforms uniform
training on AB (appendix \cref{fig:moreplacements}).
TRECs do not merely rediscover the \emph{put HQ data at the end}
heuristic used in prior work (with varying effectiveness, see
\ref{sec:prior}). Looking only at the standard loss curve for $\step$
decay (\cref{fig:hook}, $\leftfig$), one might think it best to
``anneal'' on high-quality data during the low-LR region where
training loss drops rapidly. Our placement experiments show this
intuition is incorrect: placing data at the TREC minimum yields better
validation performance.

Under the decay-to-zero schedule (\cref{fig:d2zplacements}), the TREC
minimum is likewise not at the literal end of training; the curve
turns upward as the LR approaches zero. If arbitrary placement were
allowed (rather than fixed 10\% segments), the optimal window would
occur slightly before the final steps. In practice, HQ budgets are
typically far smaller than 10\% of pre-training tokens, making this
distinction even more consequential: sub-segments earlier within the
final decile can outperform end placement.
Consistent with this observation, \cref{sec:demo} shows that in CPT
under linear decay-to-zero, mid-CPT placement outperforms end
placement across three peak learning rates.

\takeaway{TRECs reliably indicate the best point for high-quality data
  insertion within a given optimizer configuration ($\eta(t)$,
  $\lambda$, $B$).}

Interestingly, we find that absolute $\reeval(t)$ values only weakly
generalize \emph{across different optimizer configurations} for
placement prediction.  \Cref{sec:correlation} investigates the
predictiveness of TREC loss for placement across configurations, and
discusses a grokking-vs-memorization view that may explain why
configurations with similarly low TREC loss can differ in validation
performance.

\section{TREC shape is governed by the AdamW timescale}\label{sec:ema}
To apply the insights of \cref{sec:placements} (without having to
train a model twice), we need to predict TRECs in advance. Before
presenting our predictive model (\cref{sec:predict}), we build
intuition for what controls TRECs (under a fixed LR schedule), finding
they are mainly governed by the AdamW timescale.

\paragraph{Background: the AdamW EMA and its timescale.}

AdamW~\citep{loshchilov2017decoupled} updates at step $t$ can be
expressed in terms of learning rate $\eta$ and weight decay $\lambda$
as: $\theta_t = (1 - \eta\lambda)\theta_{t-1}
- \eta \frac{\hat{m}_t}{\sqrt{\hat{v}_t} + \epsilon}$, where
$\hat{m}_t$ and $\hat{v}_t$ are bias-corrected exponentially-weighted
moving averages (EMAs) of gradients and squared gradients,
respectively~\citep{kingma2014adam}.

\citet{wang2024how} observed that AdamW parameters $\theta_t$ can also
be viewed as an EMA---of weight \emph{updates}. Specifically, the
standard EMA form \mbox{$y_t = (1 - \alpha)y_{t-1} + \alpha x_t$}
matches AdamW when \mbox{$y_t=\theta_t$}, \mbox{$\alpha=\eta\lambda$},
and \mbox{$x_t=-\frac{1}{\lambda}\frac{\hat{m}_t}{\sqrt{\hat{v}_t}
+ \epsilon}$}. The \emph{timescale} $\nicefrac{1}{\alpha} =
\nicefrac{1}{\eta\lambda}$, denoted $\titer$ by
\citeauthor{wang2024how}, represents the approximate number of
iterations over which updates are averaged.

When expressed in epochs as $\tepochwang = \titer / M$, where $M$ is
the number of iterations per epoch, \citeauthor{wang2024how} found the
optimal $\tepochwang$ remained stable under model and dataset scaling
in image tasks. Maintaining a constant $\tepochwang$ requires
decreasing $\lambda$ proportionally when $M$ increases.

Since LLM pre-training typically uses a single epoch, we follow
\citet{dey2025dont} and \citet{bergsma2025power} in defining a
normalized timescale $\tepoch = \titer/T$, where $T$ is the total
number of optimization steps. As $T = D/B$ (total tokens $D$ divided
by batch size $B$):
\begin{equation}\label{eq:tema}
\tema = \frac{1}{\eta\lambda T} = \frac{B}{\eta\lambda D}.
\end{equation}

\hypothesis{For a given learning-rate decay schedule (\linear, \cosine, \step,
  etc.), the TREC is controlled by the AdamW timescale
  $\tepoch$.\label{hyp:ema}}

In other words, because parameters in AdamW are implicitly weighted
averages over updates (derived from training data), the EMA timescale
$\tepoch$ governs the scope of data influence on the final model, and
thus the TREC.\@ Yet higher EMA weight alone may not lower TREC loss:
earlier updates can lose influence as our position on the loss
surface \emph{shifts} (discussed more in \cref{sec:predict}).

\paragraph{Training fraction.}

Viewing training in terms of discrete optimizer steps becomes limiting
when batch sizes and sequence lengths vary.  Instead, we view
optimization as a continuous stochastic process over a fixed dataset,
with different batch sizes yielding different discretizations.
In this perspective, we plot TRECs against \emph{training fraction}
$\trainfrac = t/T = tB/D$; this naturally facilitates comparing curves
across different batch sizes, step counts, and dataset sizes.
To reduce noise in small-batch settings (and align with large-batch
models), we also smooth TRECs using a moving-average filter (typically
over a window of 100 steps), smoothing curves without altering the
underlying trajectory.

\paragraph{Results.}

We follow the architecture and setup of \cref{sec:placements}, but
apply:
\begin{enumerate*}[label=(\alph*)]
\item a linear decay-to-zero LR schedule,
\item a context length of 2048, and
\item the original GPT2 vocabulary (50257 tokens).
\end{enumerate*}
We train on standard SlimPajama splits with original source weightings.
Plot axes indicate whether the TREC is shown in absolute loss or
normalized (min-max scaled) form.

\begin{figure}[ht]
    \centering
    \begin{minipage}{0.33\textwidth}
        \includegraphics[trim={0.3cm 0.4cm 0.264cm 0.3cm}, clip, width=\linewidth]{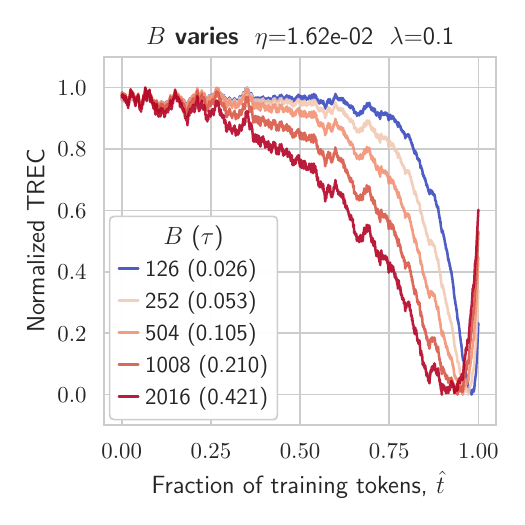}
    \end{minipage}\hfill
    \begin{minipage}{0.33\textwidth}
        \includegraphics[trim={0.3cm 0.4cm 0.264cm 0.3cm}, clip, width=\linewidth]{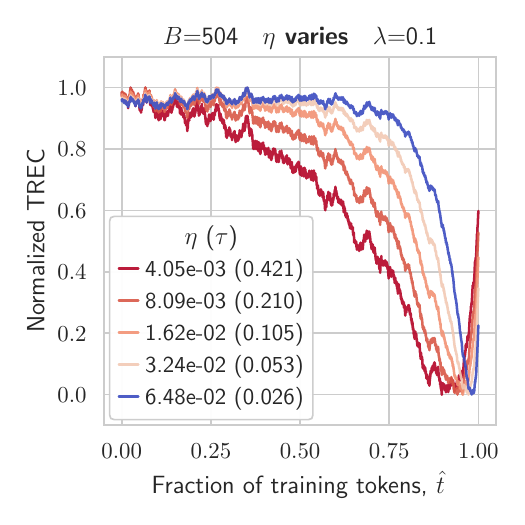}
    \end{minipage}\hfill
    \begin{minipage}{0.33\textwidth}
        \includegraphics[trim={0.3cm 0.4cm 0.264cm 0.3cm}, clip, width=\linewidth]{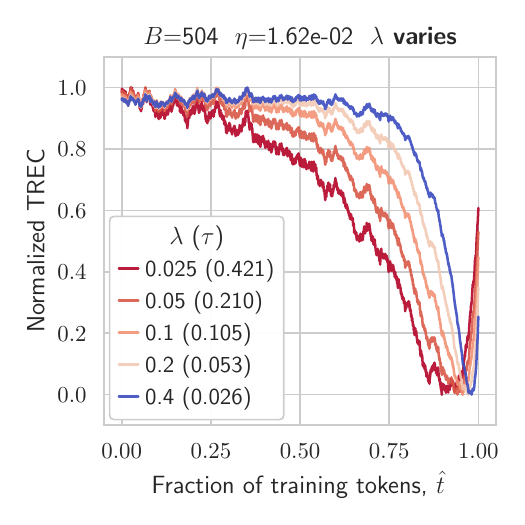}
    \end{minipage}
    \caption{\textbf{Timescale $\tema$ determines TREC shape (610M,
        80~TPP).} Sweeping $B$ ($\leftfig$), $\eta$ ($\middlefig$), or
      $\lambda$ ($\rightfig$) produces matching variations in TRECs
      when $\tema$ (\cref{eq:tema}) varies
      identically.\label{fig:tema}}
\end{figure}

\cref{fig:tema} shows normalized TRECs for 610M models
trained to 80~TPP, sweeping $\eta$, $\lambda$, or $B$ in each
subplot. Across hyperparameters, curves with matching $\tepoch$
exhibit very similar shapes, reflecting consistent timescale
control. This alignment is especially clear here since all models use
the same ordered training data. Similar patterns hold across other
scales and dataset sizes.
Generally, as the timescale expands ($\tema$ increases), the TREC
minimum (\emph{valley}) shifts earlier.\footnote{This shift has a
limit: as $\lambda \to 0$ ($\tepoch \to \infty$), the curves converge
(\cref{fig:hook}, $\middlefig$). This occurs because the \emph{shape}
of the EMA
\emph{coefficients} (\cref{sec:predict}) converges to the shape of the LR schedule (\cref{sec:adamw_limit}).}

\begin{figure}[t]
    \centering
    \begin{minipage}{0.33\textwidth}
        \includegraphics[trim={0.3cm 0.4cm 0.264cm 0.43cm}, clip, width=\linewidth]{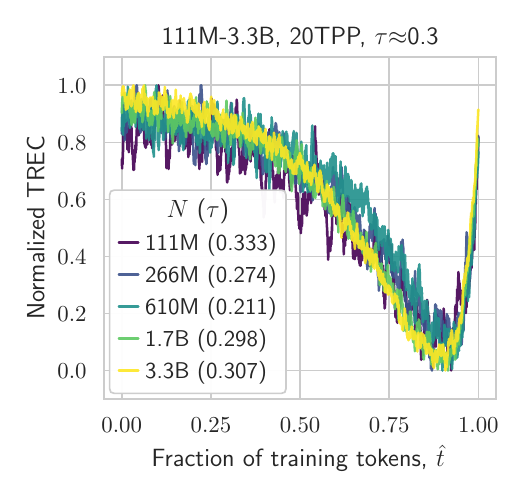}
    \end{minipage}\hfill
    \begin{minipage}{0.33\textwidth}
        \includegraphics[trim={0.3cm 0.4cm 0.264cm 0.43cm}, clip, width=\linewidth]{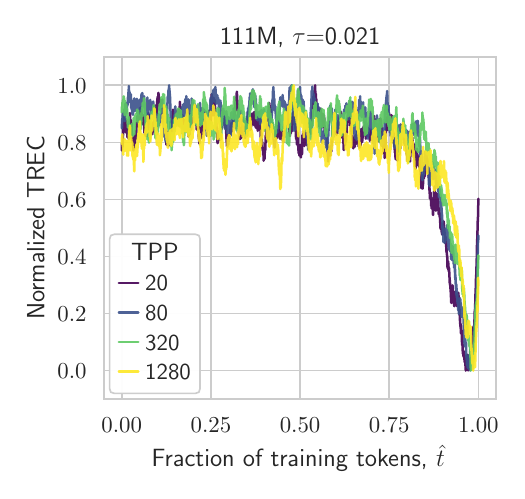}
    \end{minipage}\hfill
    \begin{minipage}{0.33\textwidth}
        \includegraphics[trim={0.3cm 0.4cm 0.264cm 0.43cm}, clip, width=\linewidth]{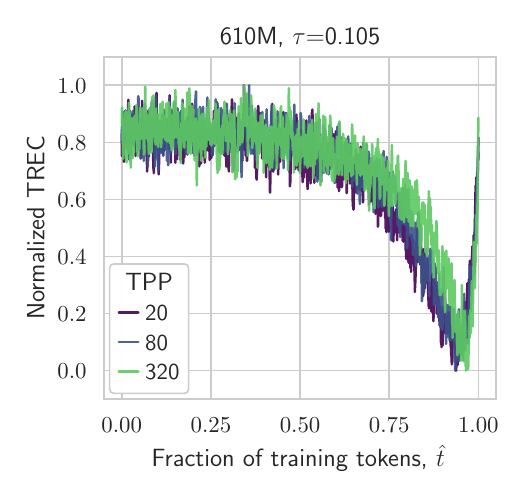}
    \end{minipage}
    \caption{\textbf{Timescale $\tema$ and TREC shape across model/dataset scales.}
      $\Leftfig$: Similar $\tema$ yields similar TREC shapes across
      model scales when training to 20~TPP ($\tema \approx 0.3$). At
      111M ($\middlefig$, $\tema = 0.021$) and 610M ($\rightfig$,
      $\tema = 0.105$), increasing TPP shifts TRECs slightly
      right.\label{fig:scaling}}
\end{figure}

We next examine TREC alignment as models and datasets scale. When
models share the same $\tema$ and TPP, their TRECs largely align
(\cref{fig:scaling}, $\leftfig$), with small $\tema$ differences
causing corresponding shifts.  The size of \emph{absolute} TREC drops
is also similar across scales when training to the same TPP
(\cref{fig:hook}, $\rightfig$).  As TPP increases, the TRECs move
slightly right (\cref{fig:scaling}, $\middlefig$, $\rightfig$); e.g.,
at 111M, increasing TPP by $64\times$ moves the TREC valley slightly
later in training.  Interestingly, as TPP increases, the size of the
TREC drop diminishes (appendix \cref{fig:morescaling}):
\emph{at lower TPP, training appears to emphasize memorization}.  This finding aligns
with a recent study of LLM memorization from an information theoretic
perspective~\citep{morris2025how}.

Finally, \cref{sec:betas} shows that large changes in AdamW's
$\beta_1$ and $\beta_2$ parameters do not significantly alter TREC
shape (even when $\beta_1 = 0$, i.e., no momentum). This suggests the
timescale imposed by weight decay is far more influential than the
timescales of momentum and velocity.

\takeaway{Overall, the data broadly supports \cref{hyp:ema}: the AdamW
  timescale ($\tema$) predominantly controls TREC shape, with TPP
  playing a secondary, smaller role.}

However, as seen in \cref{fig:placements}, the LR schedule itself
significantly affects TREC shape. To integrate LR schedules into our
analysis, we next adopt an expanded view of the AdamW timescale.

\section{Predicting TRECs: adjusting for training fraction}\label{sec:predict}
Following \cref{sec:ema}, one could in principle \textbf{predict a
TREC for any model by re-evaluating a smaller model with matching
$\tau$, TPP, and LR schedule}. More generally, we propose a functional
form that extends the AdamW EMA perspective to time-varying learning
rates, while also incorporating a training-fraction term to
capture \emph{minimizer drift}, described below.

\paragraph{Background: the extended AdamW EMA perspective.}

\citet{bergsma2025straight} extend \citet{wang2024how} by considering EMAs with time-varying
smoothing parameters $\alpha_t \in [0, 1]$. Setting $\alpha_1 = 1$ (so
$y_1 = x_1$), they show the recursion $y_2 = (1 - \alpha_2) \alpha_1 x_1 + \alpha_2 x_2$, and in general:
\begin{equation}
y_t = \sum_{i=1}^t \left( \prod_{j=i+1}^{t} (1 - \alpha_j) \right) \alpha_i x_i = \sum_{i=1}^t c_{t,i} x_i,
\label{eqn:extended_ema}
\end{equation}
where $c_{t,i}$ quantifies the contribution of input $x_i$ to the output $y_t$.

With AdamW, the LR/weight decay schedule defines the time-varying
smoothing $\alpha_t = \eta_t \lambda$ (\cref{sec:ema}). The EMA
operates over weight
updates \mbox{$x_t=-\frac{1}{\lambda}\frac{\hat{m}_t}{\sqrt{\hat{v}_t}
+ \epsilon}$}, with larger $c_{t,i}$ indicating the $i$th update
contributes more to model weights $y_t =
\theta_t$ at step $t$.
We focus on coefficients for final model weights $\theta_T$, dropping
subscript $T$ for clarity:
$c_i = \eta_i \lambda \prod_{j=i+1}^{T} (1 - \eta_j \lambda)$.
To connect with our continuous-time framing, we interpret $c_i$ over
$T$ steps as a continuous function, reparameterizing as
$c(\trainfrac)$ via $\trainfrac = i/T$, enabling direct comparison
with the TREC $\reeval(\trainfrac)$.

\begin{figure}[ht]
    \centering
    \begin{minipage}{0.33\textwidth}
        \includegraphics[trim={0.3cm 0.4cm 0.264cm 0.3cm}, clip, width=\linewidth]{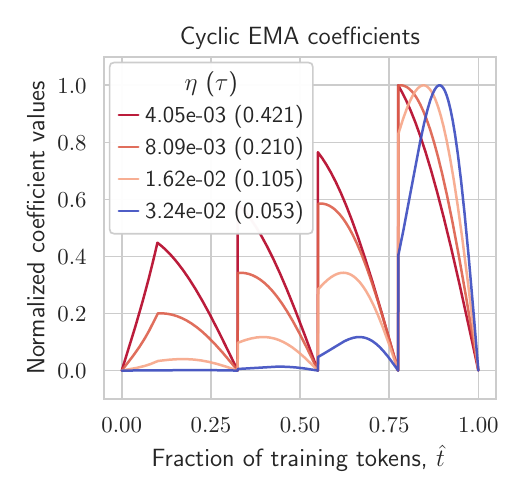}
    \end{minipage}\hfill
    \begin{minipage}{0.33\textwidth}
        \includegraphics[trim={0.3cm 0.4cm 0.264cm 0.3cm}, clip, width=\linewidth]{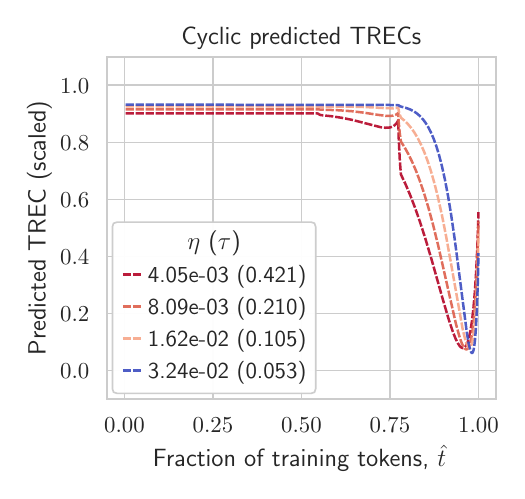}
    \end{minipage}\hfill
    \begin{minipage}{0.33\textwidth}
        \includegraphics[trim={0.3cm 0.4cm 0.264cm 0.3cm}, clip, width=\linewidth]{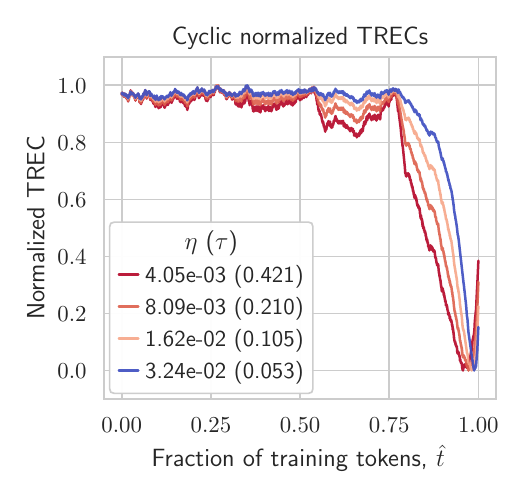}
    \end{minipage}
    \caption{\textbf{Predicting TRECs (610M, 80~TPP, Cyclic LR).}
      $\Leftfig$: Normalized EMA coefficients $c(\trainfrac)$ in our model.
      $\Middlefig$: Predicted TRECs $\hatreeval(\trainfrac)$ from \cref{eqn:prediction}.
      $\Rightfig$: True TRECs $\reeval(\trainfrac)$.
      Predictions match TREC dips, with early damping and late alignment with (inverted) EMA.
      \label{fig:preds}}
\end{figure}

\cref{fig:preds} ($\leftfig$) shows $c(\trainfrac)$ for a cyclic LR
schedule (schedule shown in \cref{fig:cyclic_wsd}). When
$c(\trainfrac)$ drops to zero (i.e., LR is zero), TRECs ($\rightfig$)
return to baseline (1.0); when $c(\trainfrac)$ is higher, TRECs dip
lower.  Yet $c(\trainfrac)$ influence fades earlier in training,
suggesting $c(\trainfrac)$ alone does not fully explain TREC shape.

\paragraph{Predicting TREC shape.}

EMA coefficients $c(\trainfrac)$ quantify each update's contribution
to the final weights, but an update's \emph{effectiveness} can fade if
the batch-specific loss surface shifts after the gradient was
computed, i.e., due to \emph{minimizer drift}.
Motivated by a simplified quadratic analysis
(Appendix~\ref{sec:theory}), we model drift on
a \emph{training-fraction clock} that is scale-invariant under $\mup$
yet LR-schedule-dependent, similar to the use of \emph{normalized
compute} in \citet{qiu2025scaling}.

\hypothesis{TREC shape can be predicted
using the AdamW EMA coefficients combined with an adjustment for
training fraction.\label{hyp:predict}}

Let $\reeval(\trainfrac)$ denote the TREC and $c(\trainfrac)$ the EMA
coefficients of the final weights $\thetaend$, both indexed by
training fraction $\trainfrac \in [0,1]$. We model the normalized TREC
shape using the simple form:
\begin{equation}\label{eqn:prediction}
  \hatreeval(\trainfrac) = 1 - c(\trainfrac)^p \cdot \trainfrac^m,
\end{equation}
where $p$ and $m$ are exponents to be fit.
Exponent $p$ controls the strength of the EMA contribution, while $m$
(the \emph{training-fraction exponent}) determines when the predicted
$\hatreeval(\trainfrac)$ begins to reflect $c(\trainfrac)$. For
example, with $m=1$, fluctuations in $c(\trainfrac)$ appear
immediately; with a larger $m$, $\hatreeval(\trainfrac)$ remains near
1 for most of training and only incorporates $c(\trainfrac)$ near the
end.

\cref{fig:preds} illustrates this formulation: EMA fluctuations
($\leftfig$) are dampened early in training in both predicted
($\middlefig$, using fits of \cref{eqn:prediction}) and true TRECs
($\rightfig$).
We focus on predicting \emph{shape}, not absolute values, and
use \cref{eqn:prediction} as a normalized functional form.  We find
tuning $p$ has minor impact on shape prediction compared to $m$, so we
fix $p=0.5$ across experiments and focus on fitting $m$.


\paragraph{Predicting the training-fraction exponent.}

For a given set of TRECs, we define the optimal $m$ as the one that
maximizes shape agreement between the predicted and true TRECs. We use
the Pearson correlation $\pearsonr$ as a scale- and shift-invariant
measure of this agreement, finding it aligns well with visual
assessments (\cref{sec:pred_details}).
Empirically, we find that optimal exponent $\mstar$ closely follows a
power-law relationship with tokens-per-parameter (TPP) and the AdamW
timescale $\tema$, which we express as:
\begin{equation}\label{eqn:optimal_m}
\mstarfit
\end{equation}
Fitting $C$, $\mu_1$, and $\mu_2$ at a small scale enables us to
predict $\mstar$ systematically across model/dataset sizes and
hyperparameter settings, completing the components needed for TREC
prediction.

\begin{figure}[ht]
    \centering
    \begin{minipage}{0.33\textwidth}
        \includegraphics[trim={0.3cm 0.4cm 0.264cm 0.3cm}, clip, width=\linewidth]{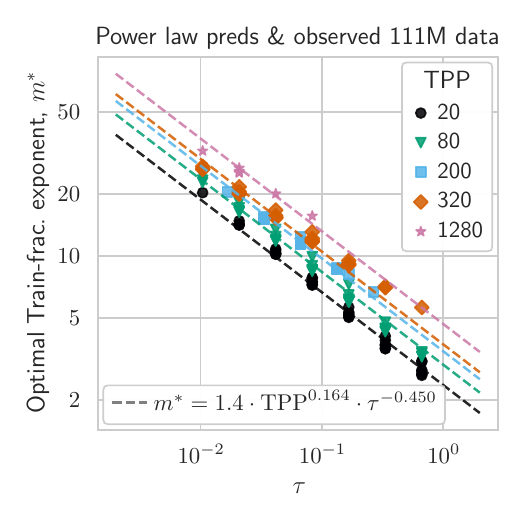}
    \end{minipage}\hfill
    \begin{minipage}{0.33\textwidth}
        \includegraphics[trim={0.3cm 0.4cm 0.264cm 0.3cm}, clip, width=\linewidth]{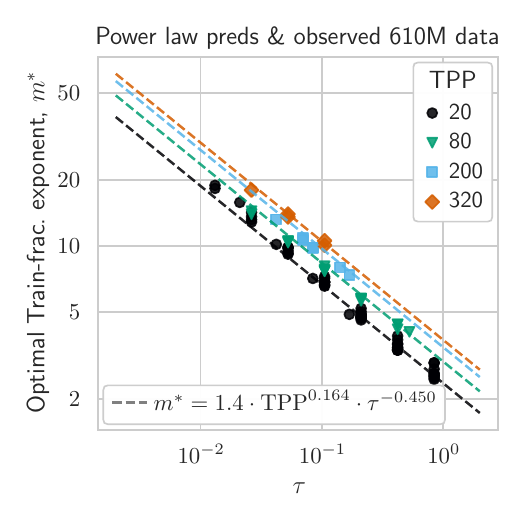}
    \end{minipage}\hfill
    \begin{minipage}{0.33\textwidth}
        \includegraphics[trim={0.3cm 0.4cm 0.264cm 0.3cm}, clip, width=\linewidth]{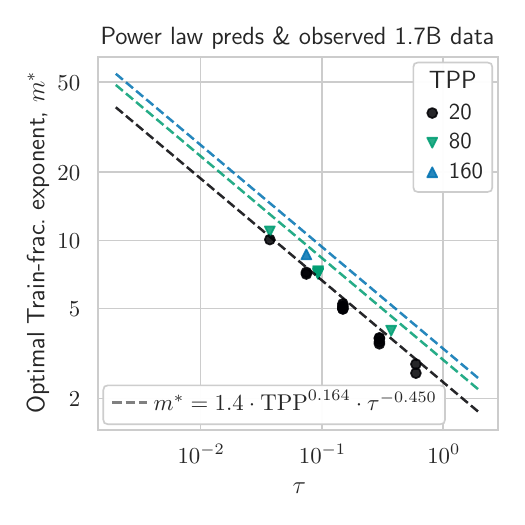}
    \end{minipage}
    \caption{\textbf{Fitted power law for $\mstar$ aligns with
        observed optima.}
      $\Leftfig$: 111M fit; $\middlefig$: 610M eval; $\rightfig$: 1.7B
      eval. Accuracy of 111M fit holds across scale, degrading
      slightly for larger models.\label{fig:fits}}
\end{figure}

\paragraph{Results.}

We follow the setup of \cref{sec:ema}, using a $\linear$ decay-to-zero
LR schedule unless noted. We evaluate prediction accuracy across
scales and datasets using: (i) $\Rtwo$ for predicted vs.\ true
$\mstar$ values, and (ii) Pearson $\pearsonr$ for predicted vs.\ true
TREC shape.
We fit the power law \cref{eqn:optimal_m} using $\mstar$ values from
small-scale 111M models, trained across varied TPPs and timescales
(\cref{fig:fits}, $\leftfig$).

Fits at 111M generalize to larger scales (\cref{fig:fits},
appendix \cref{fig:morefits}). As shown in
appendix \cref{tab:fit_scaling}, while fit $\Rtwo$ declines from 99\%
at 111M to 77\% at 3.3B, \textbf{TREC prediction accuracy remains high
across scales ($\pearsonr \sim 98\%$)}, confirming robust predictive
performance even when $\mstar$ fits are imperfect.
Ablation results (appendix \cref{tab:fit_ablation}, fitting at 111M,
evaluating at 610M scale) show that both TPP and especially $\tema$
are important for accurate $\mstar$ prediction and TREC shape
matching.

The observed dependence of $\mstar$ on $\tema$ and TPP, and also LR
schedule (\cref{sec:lr_schedule_generalization}), matches
the \emph{minimizer drift} account in \cref{sec:theory}. In the
quadratic view, drift accumulates with the schedule-weighted curvature
$\int_{\trainfrac}^{1}\eta(s)\,h(s)\,ds$; shorter EMA timescales
shorten memory and increase drift (larger $m$), higher TPP increases
the extent of curvature evolution and hence drift (larger $m$), and
different LR schedules directly change the cumulative integral,
yielding schedule-specific $m$.

\begin{figure}[ht]
  \centering
  \begin{minipage}[b]{0.33\textwidth}
    \centering
    \raisebox{0.25\height}{
      \resizebox{0.9\linewidth}{!}{
        \begin{tikzpicture}

\def\barheight{0.3}
\def\fullwidth{2.6}               
\def\cptwidth{0.06*\fullwidth}    
\def\barYgap{0.8}                 
\def\caseTwoYOffset{-0.4}         
\def\labelx{1.4}                  
\def\piex{3.1}                    
\def\radius{0.4}                 
\def\contentYOffset{-0.3}         

\def\toplabely{\barYgap + 1.0}
\node[align=center] at (\fullwidth/2,\toplabely) {\scriptsize ``CPT'' Training \\[-0.4em] \scriptsize Position};
\node[align=center] at (\piex,\toplabely) {\scriptsize ``CPT'' \\[-0.4em] \scriptsize Fraction};

\begin{scope}[yshift=\contentYOffset cm]

\node[above] at (\labelx,\barYgap+\barheight+0.1) {\scriptsize 5~TPP after PT};
\node[above] at (\labelx,\caseTwoYOffset+\barheight+0.1) {\scriptsize 5~TPP from scratch};

\draw[fill=lightgray] (0,\barYgap) rectangle (\fullwidth,\barYgap+\barheight);
\draw[fill=cborange] (\fullwidth-\cptwidth,\barYgap) rectangle (\fullwidth,\barYgap+\barheight);
\node[scale=0.8] at (\fullwidth/2, \barYgap + 0.5*\barheight) {234 TPP};

\draw[fill=cbblue] (\fullwidth-\cptwidth,\caseTwoYOffset) rectangle (\fullwidth,\caseTwoYOffset+\barheight);

\draw[->, line width=0.3pt] 
    (\labelx + 0.6, \barYgap + \barheight + 0.1) 
    -- 
    (\fullwidth - \cptwidth, \barYgap + 0.5*\barheight); 

\draw[->, line width=0.3pt] 
    (\labelx + 0.6, \caseTwoYOffset + \barheight + 0.1)
    -- 
    (\fullwidth - \cptwidth, \caseTwoYOffset + 0.5*\barheight);

\def\pieyone{\barYgap + 0.5*\barheight}
\def\pieytwo{\caseTwoYOffset + 0.5*\barheight}

\filldraw[lightgray] (\piex,\pieyone) circle (\radius);

\path[fill=cborange] (\piex,\pieyone)
  -- ++({\radius*cos(90)},{\radius*sin(90)})
  arc [start angle=90, end angle=97.2, radius=\radius]
  -- cycle;
\draw[dashed, dash pattern=on 1pt off 1pt, line width=0.3pt]
    (\piex,\pieyone) -- ++({\radius*cos(97.2)},{\radius*sin(97.2)});

\draw (\piex,\pieyone) circle (\radius);

\filldraw[cbblue] (\piex,\pieytwo) circle (\radius);
\draw (\piex,\pieytwo) circle (\radius);

\draw[dashed, dash pattern=on 1pt off 1pt, line width=0.3pt] (\piex,\pieytwo) -- ++(0,\radius);

\node at (\piex,\pieyone-0.1) {\tiny 0.98--1};
\node at (\piex,\pieytwo-0.1) {\tiny 0--1};

\end{scope}

\end{tikzpicture}
      }
    }
  \end{minipage}\hfill
  \begin{minipage}[t]{0.33\textwidth}
      \includegraphics[trim={0.3cm 0.4cm 0.264cm 0.3cm}, clip, width=\linewidth]{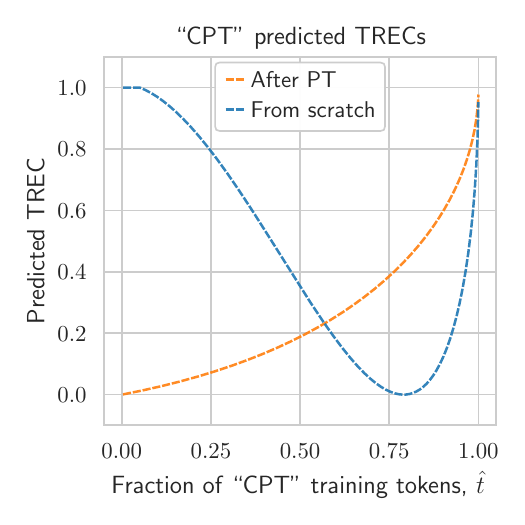}
  \end{minipage}\hfill
  \begin{minipage}[t]{0.33\textwidth}
      \includegraphics[trim={0.3cm 0.4cm 0.264cm 0.3cm}, clip, width=\linewidth]{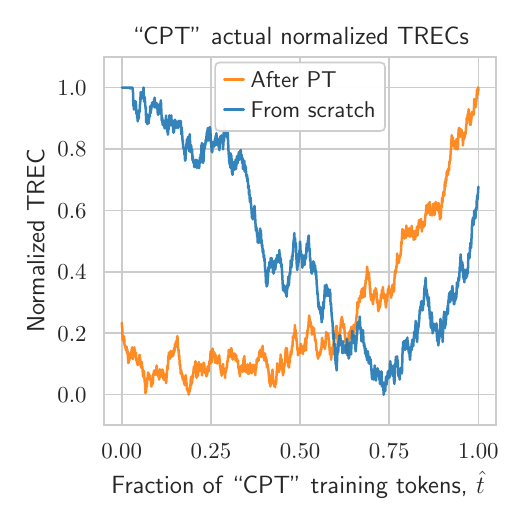}
  \end{minipage}
  \caption{\textbf{Impact of training fraction on TRECs (3.9B).}
    $\Leftfig$: Two 3.9B models trained with identical dataset, LR, weight decay, and batch size, differing only in
    init (scratch vs.\ checkpoint after 234~TPP) and thus $\trainfrac$.
    $\Middlefig$: Predicted TRECs using our framework.
    $\Rightfig$: True TRECs closely match predictions, showing training fraction determines shape under controlled conditions.
    \label{fig:pt_cpt_test}}
\end{figure}

\cref{fig:pt_cpt_test} illustrates the importance of the training
fraction term in \cref{eqn:prediction}: two 3.9B models trained with
identical dataset, learning rate, weight decay, and batch size have
the same $c(\trainfrac)$.  However, differences in initialization
(from scratch vs.\ pre-trained checkpoint) mean training evolves over
a different $\trainfrac^m$.  TREC predictions that differ solely on
the basis of $\trainfrac^m$ closely match actual TRECs.

\takeaway{TREC shape can be accurately predicted from EMA coefficients
and a training-fraction adjustment, enabling proactive curriculum
design.}

\section{Applications}
\subsection{Application to sparse mixture-of-experts (MoE)}\label{sec:moe}
We now apply TREC analysis to sparse MoE architectures, where only a
subset of parameters activate per
input~\citep{lepikhin2020gshard,fedus2022switch}. We replace each FFN
block in our 111M model with a sparse MoE layer, varying the number of
experts \( E \) from 1 (dense) to 32. Tokens are routed to experts via
hash routing~\citep{roller2021hash}, ensuring balanced usage. All
models train with identical total tokens and datasets. Yet each expert
receives only \(1/E\) of tokens, reducing an expert's \emph{effective}
tokens-per-parameter.
Timescale \(\tema\), however, is $E$-invariant: both batch size $B$
and total tokens $D$ scale identically with \(1/E\); since $\tau =
B/(\eta \lambda D)$, these reductions cancel.\footnote{Timescale
invariance may explain why correcting LR for the effectively
$1/E$-smaller per-expert \emph{batches} is
unnecessary~\citep[Appendix~C.1]{wei2024skywork}; MoEs might simply
have won the \emph{parameterization lottery}.}

\begin{figure}[ht]
    \centering
    \begin{minipage}{0.33\textwidth}
        \includegraphics[trim={0.3cm 0.4cm 0.264cm 0.3cm}, clip, width=\linewidth]{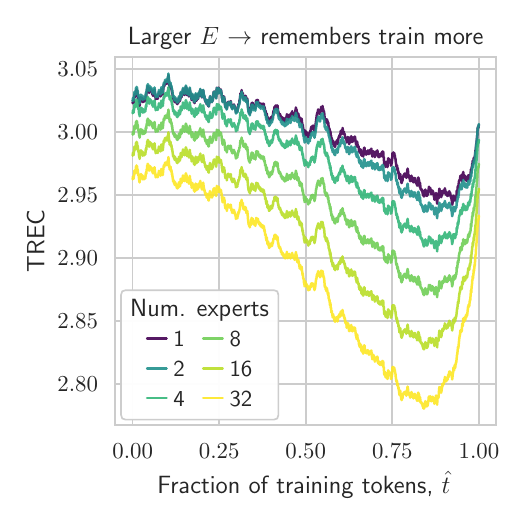}
    \end{minipage}\hfill
    \begin{minipage}{0.33\textwidth}
        \includegraphics[trim={0.3cm 0.4cm 0.264cm 0.3cm}, clip, width=\linewidth]{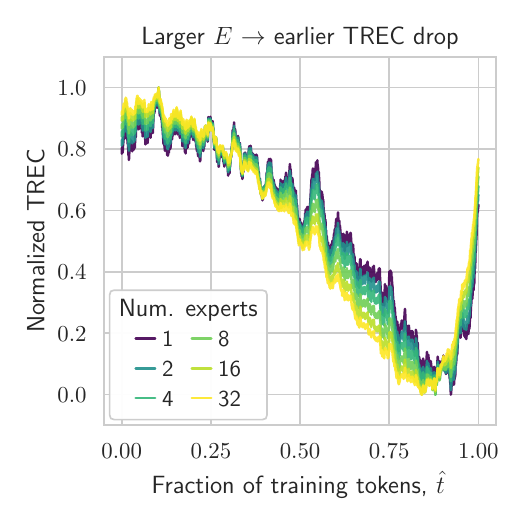}
    \end{minipage}\hfill
    \begin{minipage}{0.33\textwidth}
        \includegraphics[trim={0.3cm 0.4cm 0.264cm 0.3cm}, clip, width=\linewidth]{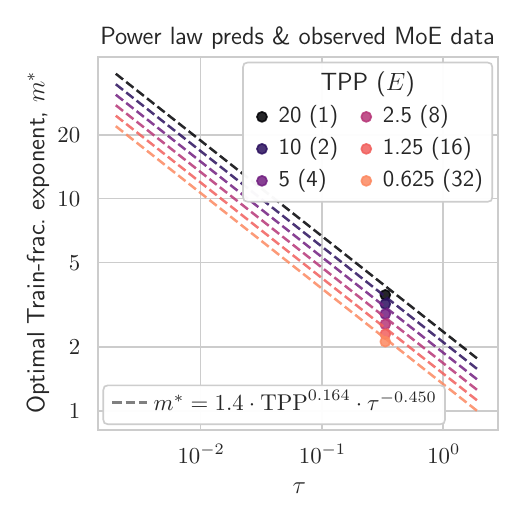}
    \end{minipage}
    \caption{\textbf{Sparse MoE TRECs reflect reduced effective TPP.\@}
      $\Leftfig$: Absolute TRECs for 111M models with increasing
      expert count $E$ (more sparsity) and increasing TREC drop.
      $\Middlefig$: Normalized curves show larger $E$ shifts the
      valley earlier.
      $\Rightfig$: $\mstar$ predictions from \cref{eqn:optimal_m}
      match true optima, confirming MoEs with more experts behave as
      if trained at reduced TPP.\@
      \label{fig:moe}
    }
\end{figure}

\cref{fig:moe} shows larger \(E\) produces greater
and earlier TREC drops, indicating stronger memorization: MoE layers
behave as if trained at their \emph{effective} TPP.\@
Indeed, empirically-optimal \(\mstar\) values for these curves align
well with predictions from our dense-data power-law model
(\cref{eqn:optimal_m}), when the power law uses the \emph{effective}
TPP (\(\Rtwo \approx 83\%\), \cref{fig:moe}-$\rightfig$). These
results complement prior observations by \citet{jelassi2024mixture},
who found that increasing experts boosts memorization more than
reasoning; our analysis suggests lower effective TPP may partly drive
this effect.

\subsection{Application to evaluating LLM recipes}\label{sec:prior}
While prior LLM recipes place their high-quality data at the end of
training, our TREC plots suggest this strategy is suboptimal. Still,
given this common practice, we ask whether the \emph{onset} of the HQ
phase---and its reported success or failure---is consistent with the
predicted TREC.\@

Llama~3~\citep{dubey2024llama} evaluated annealing on GSM8k and MATH
training sets and reported strong gains for Llama~3~8B but none for
their flagship 405B model. TRECs explain this outcome: the 405B model
annealed its LR from $8\text{e}{-7}$ to 0 over the final 40M
tokens. Since the batch size is 16M, this phase spans only $\sim$3
optimizer steps. With such a short window and vanishing LR, EMA
coefficients are essentially zero---the model retains little from this
final data.

\begin{figure}[ht]
  \centering
  \begin{minipage}[t]{0.66\textwidth}
    \vspace{0pt} 
    \centering
    \begin{minipage}{0.49\textwidth}
        \includegraphics[trim={0.3cm 0.4cm 0.264cm 0.3cm}, clip, width=\linewidth]{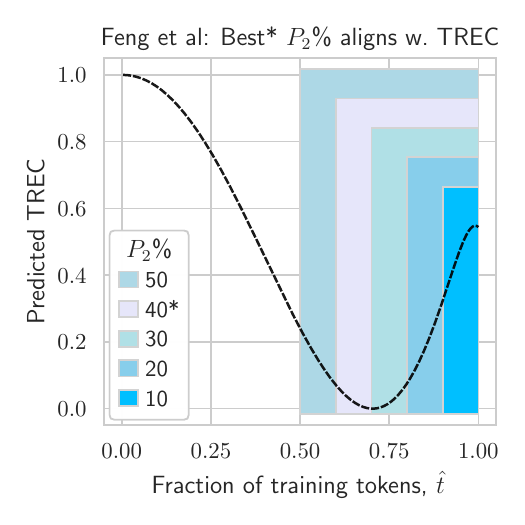}
    \end{minipage}\hfill
    \begin{minipage}{0.49\textwidth}
        \includegraphics[trim={0.3cm 0.4cm 0.264cm 0.3cm}, clip, width=\linewidth]{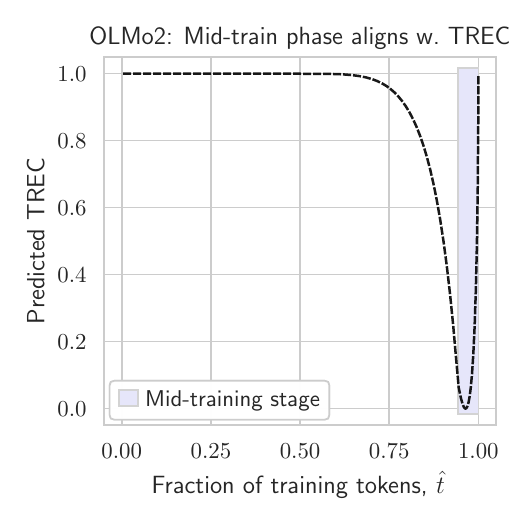}
    \end{minipage}
    \caption{\textbf{Prior work agrees with predicted TRECs.}
      $\Leftfig$: The optimal onset point for the HQ data phase in
      \citet{feng2024maximize} aligns with the TREC bottom.
      $\Rightfig$: Placement of HQ data in \citet{olmo2024} aligns
      with the much-narrower TREC valley.
      \label{fig:prior}}
  \end{minipage}\hfill
  \begin{minipage}[t]{0.32\textwidth}
    \vspace{0pt} 
    \centering
    \begin{minipage}{\textwidth}
        \includegraphics[trim={0.3cm 0.4cm 0.264cm 0.3cm}, clip, width=\linewidth]{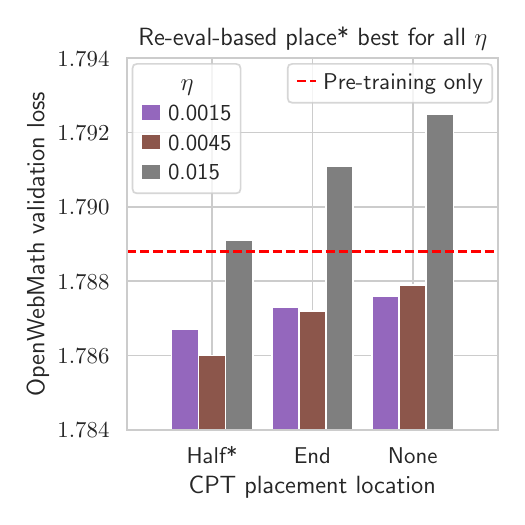}
    \end{minipage}
    \caption{\textbf{3.9B CPT results.} Placing data halfway (at
      TREC bottom) is most effective across all $\eta$ settings.
      \label{fig:celerity_barchart}}
  \end{minipage}
\end{figure}

\Cref{fig:prior} shows two other
examples. \citet{feng2024maximize} tested different onset points for
their HQ blend, finding best results when it was used for the final
40\%---a region aligning with the predicted TREC valley
($\leftfig$). In contrast, OLMo-2~13B~\citep{olmo2024} uses its HQ
blend for only the final 5.7\% of training, which again aligns with
the (narrower) predicted TREC dip ($\rightfig$).  TREC analysis
enables finding these optimal placement locations without costly
trial-and-error testing.

\subsection{Continual pre-training of a 3.9B LLM using TREC insights}\label{sec:demo}
We now test whether TREC-guided data placement improves outcomes in a
continual pre-training (CPT) setting. Prior placement experiments
in \cref{sec:placements} inserted high-quality data partway through
base-model training. In contrast, CPT typically refers to additional
training performed \emph{after} the base model is trained. To isolate
this setting, we define CPT strictly as continued
training \emph{after} learning rate decay to zero---excluding works
such as \citet{parmar2024reuse}, where the CPT LR continues from a
10\% decay value, thus arguably reflecting a mid-training strategy
(\cref{sec:nemotron}).  This definition matches intuition: CPT can be
performed multiple times, mid-training is only performed once.

Under this stricter definition, we take a 3.9B-parameter model trained
to 234 TPP (900B tokens) with learning-rate decay-to-zero, and
continue training for an additional 18B tokens (${\sim}$5 TPP;
illustrated earlier in
\cref{fig:pt_cpt_test}, $\leftfig$). During this CPT phase, we
continue training on the same data blend, but insert a 1.3B-token
segment of HQ data up-weighting \emph{math}, targeting improved
performance on OpenWebMath
validation~\citep{paster2023openwebmath}. We compare placing this HQ
data halfway through CPT, where we observed the empirical TREC for the
vanilla CPT run to be low (\textcolor{taborange}{orange line}
in \cref{fig:pt_cpt_test}, $\rightfig$), versus at the very end (as is
standard), where the TREC returns to baseline.
Full model, dataset, and experimental details are
in \cref{sec:app_demo}.

We test three CPT learning rates, as the optimal value is
unknown. While effects are necessarily small (HQ data represents only
0.14\% of total PT+CPT steps), the trend is consistent: TREC-guided
placement remains effective during CPT.\@
Placing HQ data at the TREC minimum outperforms end placement across
all LRs (\cref{fig:celerity_barchart}).  However, when LR is too high
(0.015), even correctly placed data fails to match the performance of
the original base model (dashed \textcolor{tabred}{red line} on
plot).\footnote{We also tested inserting HQ data at the predicted TREC
minimum during \emph{pre-training} ($\sim$97\%).  This mid-training
strategy matched the best CPT result, suggesting a rationale for why
mid-training is arguably supplanting CPT in practice: it achieves
comparable gains without requiring extra LR tuning or training
compute.}

\section{Related work}\label{sec:related}
Here we highlight the most relevant prior directions,
with \cref{sec:extended_related} providing full details.

\paragraph{Data curriculums, influence, and attribution.}
Curriculum learning explores strategies for effectively ordering
training data~\citep{bengio2009curriculum}. Recent LLM curriculums
often emphasize high-quality or domain-specific data in later training
phases~\citep{olmo2024,dubey2024llama}. While effective, crafting
these approaches typically relies on heuristics or expensive
experiments.

Meanwhile, quantifying influence of training points aids
interpretability, auditing, and
compensation~\citep{koh2017understanding,grosse2023studying}. Recent
scalable methods, such as data value
embeddings~\citep{wang2024capturing}, recognize the importance of
ordering but typically focus on retrospective attribution rather than
guiding training. Memorization research examines how models retain
training data, motivated by copyright or privacy
concerns~\citep{carlini2022quantifying,schwarzschild2024rethinking}.

\paragraph{Re-evaluation.}
Several works have analyzed loss on training data in order to probe
retention dynamics. \citet{pagliardini2024ademamix} examined memory
of \emph{specific} training batches as a function of optimizer type
and LR schedule.
\citet{lesci2024causal} estimate memorization profiles across training and
checkpoint steps for Pythia models, showing that retention depends on
data order and LR, and exhibits stable cross-scale trends.
\citet{bergsma2025straight} first linked retrospective
losses to AdamW's EMA dynamics.
In contrast, we systematically vary hyperparameters and LR schedules
in order to understand, predict, and exploit data retention in LLM
training.

\paragraph{Scaling collapse.} \citet{qiu2025scaling} show that when
\emph{training loss} is normalized appropriately (and training
progress is normalized similarly to $\trainfrac$), training loss
curves \emph{collapse} onto a universal trajectory across model
scales.  Deformations due to LR schedules are explained by a
noisy-quadratic analysis. Our theoretical analysis (\cref{sec:theory})
and empirical results are complementary: we study
retrospective \emph{re-evaluation} rather than initial
\emph{training} loss, account for both TPP and EMA timescale, and
translate re-eval structure into actionable curriculums.
In later work~\citep{bergsma2025scaling}, we found the same controls
that govern re-evaluation loss (TPP, $\tema$, LR schedule) govern
training loss curve shape.

\section{Conclusion}\label{sec:conclusion}
We introduced the \emph{training re-evaluation curve}, a simple
diagnostic that evaluates how well a trained model retains individual
training batches as a function of when they appeared in
training. Aligning high-quality data with TREC minima improves final
validation loss across models and training regimes (in both PT and
CPT), surpassing the default end-of-training placement. Crucially,
TRECs are largely determined by AdamW EMA coefficients and can be
predicted in advance, enabling proactive data placement.
We provide a theoretical account in which the training-fraction term
$\trainfrac^{m}$ captures, in phenomenological form, the
scale-invariant drift of batch-specific minimizers. In this view, the
fitted $m$ is directly linked to the cumulative influence of schedule
and curvature dynamics.
Our insights explain existing curriculum strategies, identify
suboptimal ones, and yield improved performance in large-scale
continual pre-training.
Taken together, our results position TREC-based placement as a
principled alternative to suboptimal heuristics and costly data-onset
ablations.

\bibliography{bib_reeval}
\bibliographystyle{cereb}

\newpage
\appendix
\section{Limitations and Future Work}\label{sec:limitations}
While TRECs offer a practical and predictive diagnostic
for guiding data placement in LLM training, several limitations and
opportunities for future work remain.

\paragraph{Optimizer scope.}  

The TREC itself is not optimizer-specific: TRECs can be computed for
any training run regardless of the optimization algorithm.
However, our \emph{predictive} analysis of TRECs is tailored to the
AdamW optimizer, drawing on its implicit EMA formulation and
corresponding timescale.
Update rules for other optimizers using weight decay (such as
Sophia~\citep[Algorithm~3]{liu2023sophia} and
MuonClip~\citep[Algorithm~1]{kimi2025k2}) can be directly converted to
an extended EMA form, exactly as was done with AdamW
(\cref{sec:predict}).
Moreover, the EMA perspective should also hold when AdamW is applied
in a different weight basis, e.g., as in SOAP~\citep{vyas2024soap},
where AdamW is applied in Shampoo's
eigenbasis~\citep{gupta2018shampoo}.

Extending predictive TREC models to optimizers without an implicit EMA
formulation---including Adagrad~\citep{duchi2011adaptive},
Adafactor~\citep{shazeer2018adafactor}, or SGD variants---remains an
important avenue for future exploration.
For optimizers that do not use weight decay, such as
Adam~\citep{kingma2014adam}, we can potentially view them as the limit
of weight-decay-enhanced versions, as weight decay goes to zero.  As
shown in \cref{sec:adamw_limit}, such optimizers may approach a
specific EMA \emph{shape}, even when the timescale is undefined.

\paragraph{Training setups and model scales.}  
Our experiments focus primarily on compute-optimal and overtrained
regimes, using training runs at or beyond 20 tokens-per-parameter
(TPP).
While this aligns with common practices in large-scale LLM
development, further work is needed to understand how TREC dynamics
behave in undertrained or data-scarce settings. We also primarily
study models in the 100M-4B range; exploring scaling trends for
smaller or larger models could refine our conclusions.

\paragraph{Data types, quality, and curriculums.}
Our work focuses on the placement of high-quality data presumed to be
limited in quantity, consistent with prior observations that ``truly
high-quality tokens are still scarce at this
moment''~\citep{wang2025octothinker}.  We test placing discrete
segments of HQ data, rather than continuously evolving data
distributions.  Moreover, although our findings generalize across
blends involving code, math, and web text, we do not explicitly
analyze how TRECs (or the reliability of TREC-guided placements) vary
for distinct data types---such as factual vs.\ reasoning, or
instruction vs.\ narrative content. A natural extension would be to
combine our placement framework with recent approaches for data
selection and weighting, including AutoScale~\citep{kang2024autoscale}
and RegMix~\citep{liu2024regmix}.

Optimizing placement to maximize retention and downstream task
performance raises important questions. For example, increasing
memorization may come at the expense of general reasoning
ability~\citep{jelassi2024mixture}, or may confound evaluation by
overly tailoring training to benchmark
tasks~\citep{dominguez2024training}. At the same time, high-quality
domain-specific pre-training has been shown to be essential for models
to benefit from downstream reinforcement
learning~\citep{wang2025octothinker}, or SFT datasets that differ
substantially from the PT distribution~\citep{liu2025midtraining}. The
TREC framework provides a tool for maximizing the effect of limited
training data---but whether learning from such data is
ultimately \emph{beneficial} remains an open, complex, and important
question in language model research.

\paragraph{Context length, vocabulary, and data diversity.}  
Our study is limited to two vocabulary configurations and context
lengths of 2048 and 8192 tokens. The interaction between TREC
dynamics and architectural choices such as tokenizer design or
sequence length remains underexplored. Similarly, we evaluate a modest
range of dataset blends, and further validation is needed to assess
generality across diverse languages or modalities.

\paragraph{Evaluating memorization and generalization.}  
TRECs measure how well the final model retains or forgets
data presented at different points in training.  However, we do not
attempt to quantify exact memorization as defined in prior work (e.g.,
substring
continuation~\citep{carlini2022quantifying,georgiev2024gemini,schwarzschild2024rethinking}).
Future studies could examine how TREC loss relates to
sequence-level memorization and whether schedules that maximize
retention may inadvertently encourage overfitting. In a similar
vein, \citet{biderman2023pythia} offered advice on where to ``place
sequences that are undesirable to memorize.''
It is worth studying whether deliberately placing biased or
undesirable data away from the TREC minimum may offer a new tool
for mitigating unwanted retention.  See also our notes on the
``memorization window'' in \cref{sec:extended_related}.

\paragraph{Predictive scope and validation.}  
Our predictive framework successfully anticipates TREC shape across
optimizer settings and learning rate schedules. However, predictions
of optimal placement for prior work are not verified via end-to-end
retraining.  Additionally, our CPT experiments suggest that predictive
placement is most reliable within a particular optimizer configuration
($\eta(t)$, $\lambda$, $B$), and may fail to generalize across
configurations.  Finding effective techniques that jointly choose both
placement and optimizer configuration is an open goal and an important
next step.  Doing so will likely require predicting when and why a
model transitions from memorization into grokking, as discussed
further in \cref{sec:correlation}.

Also, our predictive form focused on the (normalized) \emph{shape} of
the TREC loss (\cref{eqn:prediction}), rather than its absolute
\emph{magnitude}.  It would be valuable to explore other forms that can
predict both shape and magnitude.  It would also be valuable to expand
our theoretical model to encompass these extended forms
(\cref{sec:theory}).

\paragraph{Designing LR schedules that do not forget.}  
Given our predictive form for TREC loss (\cref{eqn:prediction}), it
is possible to design a learning rate schedule such that the EMA
coefficients ``cancel out'' the effects of the training fraction term.
That is, in theory, we may design a schedule in order to obtain a flat
TREC.\@

In practice, such a \emph{no-forgetting} schedule would have to
rapidly decrease the LR as a power law in training steps, in order to
offset the increasing power-law of the training fraction term.  In
prior work, such rapidly decreasing LR schedules do not perform as
well as more-gradual
decline~\citep{defazio2023when,bergsma2025straight}.  Based on the
discussion in \cref{sec:theory}, we may understand why: since
gradients lose their effectiveness, some forgetting is
actually \emph{desirable} in LLM pre-training.  Indeed, the optimal
EMA timescale has been shown to decrease as a power law in
TPP~\citep{bergsma2025power}, meaning that when training longer,
relatively more of the data should be forgotten.

However, there are contexts where avoiding forgetting may be important
(e.g., when performing CPT or SFT).  It would also be interesting to
investigate and mitigate sources of local optimizer drift, as a means
to reducing the need for forgetting.  Can we transition learning to a
regime where new knowledge can be added indefinitely, without
fundamentally changing the representation of such knowledge?  Recent
work using \emph{lazy learning} to avoid catastrophic forgetting may
provide valuable insights here for the TREC
perspective~\citep{graldi2025importance}.

\paragraph{Toward practical deployment.}  
Finally, while we offer actionable guidance for curriculum design
(e.g., predicting TRECs in advance, avoiding late placement
under step-drop or $\dtoz$ LR schedules, or leveraging homogeneous CPT
phases to measure TREC dips), wider adoption will depend on
usability.
We will explore mechanisms to make TREC tools more performant and
accessible to the community.
For example, for cases where TRECs are constructed through
explicit re-evaluation rather than prediction, compute can be saved
by \emph{sampling} a portion of the training batches to re-evaluate
on, rather than evaluating on every batch (as we do).  Determining an
acceptable fraction depends on loss variance, and should be
investigated systematically in future work.

\section{Additional related work}\label{sec:extended_related}
\paragraph{Optimizing data mixtures and quality.}
Recent work has explored methods for improving LLM training via better
data selection or mixture strategies.  One line of work focuses on
identifying \emph{what} constitutes high-quality data, including
weak-to-strong selection using attention
mechanisms~\citep{hua2025attentioninfluence}, filtering based on
scaling law deviations~\citep{li2024scalingfilter}, or assessments
from trained models~\citep{sachdeva2024train,li2024datacomp}.  Another
line of work addresses \emph{how} to mix datasets drawn from multiple
domains.  \citet{ye2024data} introduce \emph{data mixing laws},
showing that validation loss can be predicted as a function of mixture
proportions and proposing nested use of scaling laws to generalize to
larger model/data regimes.  \citet{liu2024regmix} similarly use
small-scale training runs to regress on mixture efficacy and
extrapolate to larger models, outperforming heuristic and
prior-optimized mixtures.  AutoScale~\citep{kang2024autoscale} takes a
more theoretical approach, modeling how optimal domain weights vary
with training scale, and deriving recipes that converge faster and
perform better than baselines.  While these efforts offer valuable
tools for \emph{which} data to include and \emph{how} to weight it,
our work focuses on \emph{when} to introduce data (possibly very
limited in size) during training.

\paragraph{Data curriculums and annealing phases.}
Our work shares conceptual motivation with curriculum
learning~\citep{bengio2009curriculum,mindermann2022prioritized}, which
aims to improve generalization by presenting examples in the most
effective order. In modern LLM training, this principle manifests as
staged or ``annealed'' data mixtures, where high-quality or
domain-specific datasets are introduced later in training.  In
Llama-3~\citep{dubey2024llama}, the authors explicitly report
following OpenAI's strategy~\citep{achiam2023gpt} of annealing on
in-domain datasets like GSM8k~\citep{cobbe2021training} and
MATH~\citep{hendrycks2021measuring}.\footnote{While the placement
details are often not reported (i.e., when exactly such data is
introduced during pre-training),
\citet{dominguez2024training} specifically identify November 2023 as a
turning point, after which technical reports ``start referencing
certain pre-training practices that may amount to training on the test
task,'' such as using instruction-tuning data or QA templates.}
Other state-of-the-art models using such practices include
OLMo-2~\citep{olmo2024}, JetMoE~\citep{shen2024jetmoe},
Phi-3~\citep{abdin2024phi}, Gemini~\citep{team2023gemini},
Gemma~\citep{team2024gemma}, MAP-Neo~\citep{zhang2024map},
Falcon-Mamba~\citep{zuo2024falcon} and
Yi-Lightning~\citep{wake2024yi}. These strategies are premised on the
assumption that exposing key data late in training improves downstream
task performance. Recent work has also used annealing phases to assess
data quality
efficiently~\citep{blakeney2024does,dubey2024llama,olmo2024}, enabling
comparisons without full-scale pre-training. However, placing the
highest-quality data at the very end is a flawed heuristic, and using
ablations to determine the optimal onset for the high-quality phase is
expensive.
Our TREC diagnostic addresses these issues by offering a scalable
method for identifying optimal data placement locations.

\paragraph{Measuring loss on training data post-hoc.}
As far back as \citet{graves2013generating}, it was observed that
generated text from a neural language model tends to inordinately
reflect the final training batches; from a TREC-perspective, this
manifests as lower TREC losses on the final training samples.
While not the primary focus of these works, previous studies have also
explicitly measured loss on previously seen training examples in order
to understand model retention dynamics. In a comparison of different
LR schedules, \citet{bergsma2025straight} qualitatively connected
training re-evaluation losses to the exponential moving average (EMA)
coefficients of AdamW updates. \citet{pagliardini2024ademamix} propose
AdEMAMix, an optimizer designed to forget more slowly.  They visualize
loss trajectories for individual batches across training and conclude
that learning rate decay is the dominant factor controlling
forgetting.

\citet{lesci2024causal} introduce a difference-in-differences estimator
for counterfactual memorisation, constructing a two-dimensional
memorisation profile $\tau_{g,c}$ that measures the causal effect of
\emph{training on batch $g$} on \emph{model performance at checkpoint $c$}.
Their analysis shows that memorisation strength in the Pythia model
suite~\citep{biderman2023pythia} depends on learning rate, data order,
and model scale. Conceptually, our TREC can be viewed as a specific
slice of such a memorisation profile: we fix $c = T$ (the final
checkpoint) and measure the retrospective loss as a function of
training position $g$.
By not relying on public checkpoints, but instead systematically
varying optimizer hyperparameters during pre-training, we develop a
deeper understanding of the factors that govern retention.  We also
differ from all these works in that we evaluate TRECs as a tool to
guide curriculum design.

\paragraph{Influence estimation and data attribution.}
Training data influence has long been studied for interpretability and
accountability. Influence function
approaches~\citep{koh2017understanding} and retraining-based
approximations~\citep{feldman2020neural,grosse2023studying} estimate
data value by measuring its effect on final model behavior. Recent
scalable methods such as DataInf~\citep{kwon2023datainf},
LESS~\citep{xia2024less} and LoGRA~\citep{choe2024what} leverage
gradient-based approximations (e.g., via LoRA) to approximate
influence at scale. \citet{wang2024capturing} break from
permutation-invariant assumptions and introduce trajectory-specific
data value embeddings that explicitly model training data order. Their
method uncovers distinct training phases in LLMs: a ``high-impact
warmup phase,'' followed by a ``low-impact basin,'' and then a
``gradual ascending'' region. While insightful, this pattern differs
from our findings: both EMA analysis and TREC diagnostics suggest that
specific batches in early training data have minimal influence on
final weights. In this sense, our work serves as a valuable
cross-check for attribution analyses---TRECs offer a simple,
forward-only sanity-check that can validate or challenge more complex
influence models.

\paragraph{Memorization and forgetting dynamics.}
A growing literature studies memorization in LLMs, often motivated by
copyright or privacy risks. Typical methods identify memorized
sequences by testing whether prompting with part of a training example
elicits exact or near-exact
continuations~\citep{carlini2022quantifying,georgiev2024gemini,schwarzschild2024rethinking}.
\citet{morris2025how} propose an information-theoretic definition of
memorization based on Kolmogorov complexity, and show how it varies
across dataset size, with conclusions that align with our own TREC
results (\cref{sec:ema}).

In a large-scale analysis, \citet{biderman2023pythia} found that
training order had little impact on memorization, with memorized
sequences distributed approximately as a Poisson process across
training. This contrasts sharply with our results: TRECs consistently
show strong order effects on loss, indicating meaningful differences
in retention and learning across training time. Exact sequence
reproduction may be too coarse a signal to capture the subtler,
gradient-based adaptations revealed by TREC analysis.

Seeking to avoid the known deleterious effects of data
repetition~\citep{hernandez2022scaling}, yet eager to make full use of
their high-quality data, \citet{tiifalconh1} report successful re-use
of such data, so long as one ``carefully estimates'' and avoids the
model's ``memorization window.'' Though no further details are
offered, their approach suggests another promising application of TREC
analysis: it could help identify when high-quality data can be safely
repeated.

While the prior works aim to avoid memorization, a separate line of
work focuses on the opposite failure mode: catastrophic
forgetting. This is a central concern in continual
learning~\citep{kirkpatrick2017overcoming}, where models forget
previously-learned knowledge when exposed to new data.
Closely related is the issue of \emph{loss of plasticity}, in which
extensive pre-training reduces a model's capacity to acquire new
information~\citep{ash2020warm,lyle2023understanding,dohare2024loss,kumar2024scaling,springer2025overtrained}.
Our TREC framework helps disentangle these effects. For example,
training to a higher TPP reduces the \emph{magnitude} of the TREC drop
(reflecting reduced plasticity), while mainly preserving
the \emph{shape} of the curve (indicating which segments are most
forgotten).

\paragraph{Continual pre-training (CPT) dynamics.}
Continual pre-training (CPT), or lifelong learning, involves adapting
models to new data distributions beyond the initial training
set. Earlier work in this area emphasized domain adaptation for
classification tasks~\citep{gururangan2020dont,qin2022elle}, while
recent LLM recipes apply CPT to full model continuation. However, such
practices often suffer from performance degradation, even when
continuing on the original domain, due to optimization
challenges~\citep{ibrahim2024simple,oncel2024adaptation}. CPT
strategies commonly repurpose pre-training heuristics---such as
annealing phases or late-stage data swaps---without principled
guidance. Our work introduces a TREC framework that directly informs
when and how to incorporate new data during CPT.\@

\paragraph{Scale-stable dynamics under $\mup$ and normalized compute.}
Feature-learning parameterizations such as $\mup$ can transfer
hyperparameters across scale and yield early-time consistency of
dynamics across widths \citep{vyas2023feature,kalra2023universal},
though finite-width deviations grow on harder tasks and later
epochs. Complementing this, \citet{noci2024super} provide evidence
of \emph{super-consistency} in curvature (e.g., Hessian eigenvalues)
along the training trajectory, supporting training fraction as a
natural coordinate (\cref{sec:tf_clock}). Building on these
observations, \citet{qiu2025scaling} show that when loss is indexed
by \emph{normalized compute} $x=t/t^*(p)$, where $t$ is the current
step and $t^*(p)$ is the compute-optimal step count for model size
$p$, then training-loss curves collapse across scales, with
LR-schedule-dependent deformations explained by a noisy-quadratic
analysis; collapse also holds for fixed multiples of $t^*(p)$.  Their
experiments, however, are limited to proof-of-concept models at
relatively small scale, whereas we validate TREC dynamics in LLMs up
to 3.9B parameters. Our setting also differs in target (re-evaluation
vs.\ training loss) and mechanism (AdamW EMA timescale + drift).

In later work~\citep{bergsma2025scaling}, we found the same controls
that govern re-evaluation loss (TPP, $\tema$, LR schedule) govern
training loss curve shape.

\section{Experimental details}\label{sec:experimental_details}

\begin{table}
  \centering
  \caption{Model architectures used in main experiments\label{tab:model_info}}
\begin{tabular}{@{}cccccc@{}}
\toprule
Model & $\dmodel$ & $\nlayers$ & $\dffn$ & $\dhead$ & Experiments \\ \midrule
111M  & 768    & 10 &  2048 & 64 & \cref{sec:ema,sec:moe}           \\
266M  & 768    & 32 &  2048 & 64 & \cref{sec:ema} \\
610M  & 2048   & 10 &  5461 & 64 & \cref{sec:placements,sec:ema}          \\
1.7B  & 2048   & 32 &  5461 & 64 & \cref{sec:ema}         \\
3.3B  & 2048   & 64 &  5461 & 64 & \cref{sec:ema}          \\
3.9B  & 2048   & 40 & 16384 & 128 & \cref{sec:predict} (\cref{fig:pt_cpt_test}), \cref{sec:demo} \\ \bottomrule
\end{tabular}
\end{table}

\begin{table}
  \centering
  \caption{Models, tokens-per-parameter (TPP) and corresponding
    dataset sizes (in tokens), number of model variants trained (LR
    schedule type, $\eta$, $\lambda$, $B$, or data placement
    strategy), and purpose of trained models.  In total, 41 models
    were trained with different data placements, and 578 TRECs were
    computed over different optimizer
    hyperparameters.\label{tab:train_steps}}
\begin{tabular}{@{}ccccc@{}}
  \toprule
Model & TPP & $D$ & Variants trained & Purpose \\ \midrule
111M  & 20  & 2.19B  & 61 & Fitting/evaluating TREC prediction \\
111M  & 80  & 8.76B  & 50 & Fitting/evaluating TREC prediction \\
111M  & 200 & 21.9B  & 21 & Fitting/evaluating TREC prediction \\
111M  & 320 & 35.0B  & 40 & Fitting/evaluating TREC prediction \\
111M  & 1280 & 140.1B & 11 & Fitting/evaluating TREC prediction \\
266M  & 20  & 5.31B  & 25 & Fitting/evaluating TREC prediction \\
266M  & 80  & 21.2B  & 19 & Fitting/evaluating TREC prediction \\
266M  & 320 & 85.0B  & 19 & Fitting/evaluating TREC prediction \\
266M  & 1280 & 339.8B & 3 & Fitting/evaluating TREC prediction \\
610M  & 20  & 12.1B  & 205 & Fitting/evaluating TREC prediction \\
610M  & 80  & 48.5B  & 53 & Fitting/evaluating TREC prediction \\
610M  & 82  & 50.0B  & 30 & Mid-training data placement tests (code blend) \\
610M  & 200 & 121.3B & 14 & Fitting/evaluating TREC prediction \\
610M  & 320 & 194.1B & 5 & Fitting/evaluating TREC prediction \\
1.7B  & 20  & 34.3B  & 31 & Fitting/evaluating TREC prediction \\
1.7B  & 80  & 137.2B & 11 & Fitting/evaluating TREC prediction \\
1.7B  & 160 & 274.3B & 1 & Fitting/evaluating TREC prediction \\
1.7B  & 320 & 548.6B & 1 & Fitting/evaluating TREC prediction \\
3.3B  & 20  & 66.5B  & 2 & Fitting/evaluating TREC prediction \\
3.3B  & 23  & 76.5B  & 1 & Fitting/evaluating TREC prediction \\
3.9B  & 234 & 909.2B & 2 & Mid-training data placement tests (math blend) \\
3.9B  & 239 & 923.4B & 9 & Continual PT data placement tests (math blend) \\
\bottomrule
\end{tabular}
\end{table}

\cref{tab:model_info} provides details on the architecture for models
used in the main experiments, while \cref{tab:train_steps} provides,
for each model scale and TPP, the dataset sizes used in training, and
the number of training variations explored at that scale (varying data
placement strategy, or LR schedule and hyperparameters $\eta$,
$\lambda$, $B$).
In total, 578 TRECs were computed for the main experiments.

Further details of the 3.9B model and settings for continual
pre-training experiments are in \cref{sec:app_demo}.  In the remainder
of this section, we discuss the main pre-training settings.

All trained models were GPT2-style LLMs~\citep{radford2019gpt2} with
ALiBi~\citep{press2022alibi} embeddings and
SwiGLU~\citep{shazeer2020glu} non-linearity.
We use the AdamW optimizer.  Following standard practice, we do not
apply weight decay or bias to LayerNorm layers.
AdamW settings are $\beta_1 = 0.9$, $\beta_2 = 0.95$, and $\epsilon = 1$e$-8$.
We report cross-entropy loss.
%
%
By default we parameterize with maximal update parameterization,
$\mup$~\citep{yang2022mup}, with hyperparameters set via proxy tuning,
as described below.

For a given TPP, all models have the exact same warmup phase: a linear
warmup of the learning rate from 0 to the maximum value.
In all runs (aside from training of the 3.9B model), warmup was 10\%
of the total steps.
Learning rate warmup is standard practice in LLM
training~\citep{brown2020language,rae2022scaling,biderman2023pythia,dubey2024llama,kosson2024analyzing}.

All models in the experiments were trained on Cerebras CS-3 systems.
610M-parameter 20~TPP models take roughly 6 hours each to train on a
single CS-3.

\paragraph{Proxy model hyperparameter tuning.}\label{sec:proxy_tuning}

\begin{table}
    \centering
    \caption{Tuned hyperparameters for $\mup$ proxy model\label{tab:mup_hps}}
    \begin{tabular}{cc}
         \toprule
         $\sigma_{W,\text{base}}$& $8.67$e-$02$ \\
         $\hateta$& $\maxlrdetail$\\
         $\alpha_{\text{input}}$& $9.17$\\
         $\alpha_{\text{output}}$& $1.095$\\
         \bottomrule
    \end{tabular}
\end{table}

We now describe how we tuned the $\mup$ base hyperparameters (HPs).
Our proxy model is a 39M-parameter LLM with a width $\dproxy$ of 256,
depth of 24 layers, and a head size of 64.  Tuning runs were conducted
on 800M tokens with $B = 256$ sequences and a context length of 2048
tokens.  Tuning was performed by randomly sampling 350 configurations
of base LRs, initialization standard deviations, and
embedding/output-logit scaling factors.  \Cref{tab:mup_hps} gives the
resulting top-performing values, which we used as our tuned HPs.

It is also worth noting that the LR values shown in \cref{fig:tema},
\cref{fig:celerity_barchart} and the appendix figures are the base
$\mup$ LRs \emph{before} $\mup$-adjustment.  Calculation of $\tema$
(\cref{sec:ema}) and the EMA coefficients (\cref{eqn:extended_ema})
requires the adjusted LR (i.e., multiplying by $\dproxy / \dmodel$).
Also, when LR decay is used, reported LR values always refer to the
peak/max LR of the LR schedule.

\subsection{Placement tests: experimental details}

For the placement tests in \cref{sec:placements}, we used the standard
training and validation splits of the SlimPajama
dataset~\citep{cerebras2023slimpajama}, but with different weighting
of subsets as given in \cref{tab:blend}.
For these experiments, we used a context length of 8192 tokens, batch
sizes of 126, base peak LR of $\eta = 1.62$e$-2$ (the MUP
proxy-model-tuned LR), and weight decay of $\lambda = 0.1$.
We used the Llama-3~\citep{dubey2024llama} vocabulary size of 128256.

\subsection{TREC fitting and prediction: experimental details}

For the experiments and analysis in \cref{sec:ema}, \cref{sec:predict}
and \cref{sec:moe}, we use a context length of 2048 tokens and the
GPT2~\citep{radford2019gpt2} vocabulary of size 50257.  For these
experiments, we use the default source weightings for the SlimPajama
dataset.

\section{Further data placement results}\label{sec:app_placements}

We present additional experimental results supporting our finding that
optimal data placement corresponds to the lowest point on the TREC.\@
These results cover additional learning rate schedules, summarize
placement effectiveness across blends, and extend validation to
alternative metrics.

\subsection{Within-schedule placement outcomes}

\begin{figure}[ht]
    \centering
    \begin{minipage}{0.33\textwidth}
        \includegraphics[trim={0.3cm 0.4cm 0.264cm 0.3cm}, clip, width=\linewidth]{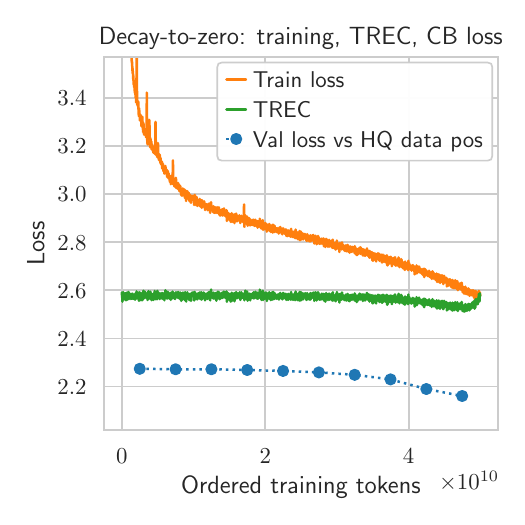}
    \end{minipage}
    \hspace{\twofighspace}
    \begin{minipage}{0.33\textwidth}
        \includegraphics[trim={0.3cm 0.4cm 0.264cm 0.3cm}, clip, width=\linewidth]{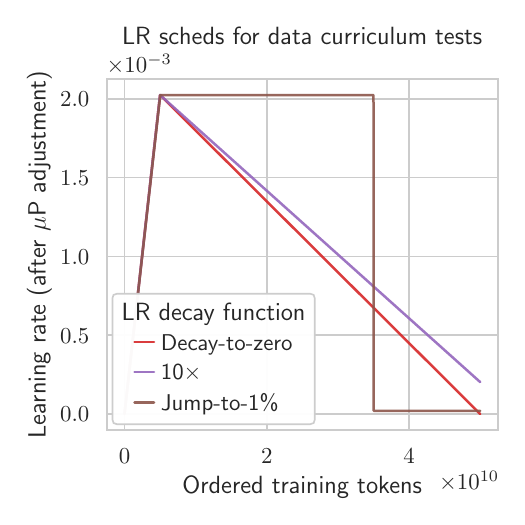}
    \end{minipage}
    \caption{\textbf{TREC-guided data placement: further details.}
      $\Leftfig$: TRECs predict best data placement in terms
      of resulting CB validation loss, for a linear decay-to-zero ($\dtoz$)
      LR schedule.
      $\Rightfig$: Plots of the $\dtoz$, $\tenx$, and $\step$ LR
      schedules used in \cref{sec:placements} (formulas in
      \cref{tab:lr_schedules}).
      \label{fig:d2zplacements}}
\end{figure}

\cref{fig:d2zplacements} provides the decay-to-zero ($\dtoz$) data
placement results, along with the LR schedules for $\dtoz$, $\tenx$,
and $\step$ drop (after 70\% of training); these are the three
schedules tested in the \cref{sec:placements} experiments.  For
$\dtoz$, although the TRECs bend back up to baseline at the end as
expected, note the tenth data placement position still obtains the
lowest average TREC loss, and consequently is the optimal data
placement location.

\begin{figure}[ht]
    \centering
    \begin{minipage}{0.33\textwidth}
        \includegraphics[trim={0.3cm 0.4cm 0.264cm 0.3cm}, clip, width=\linewidth]{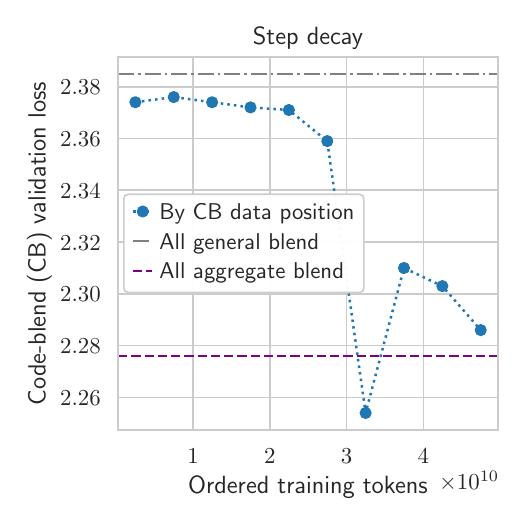}
    \end{minipage}\hfill
    \begin{minipage}{0.33\textwidth}
        \includegraphics[trim={0.3cm 0.4cm 0.264cm 0.3cm}, clip, width=\linewidth]{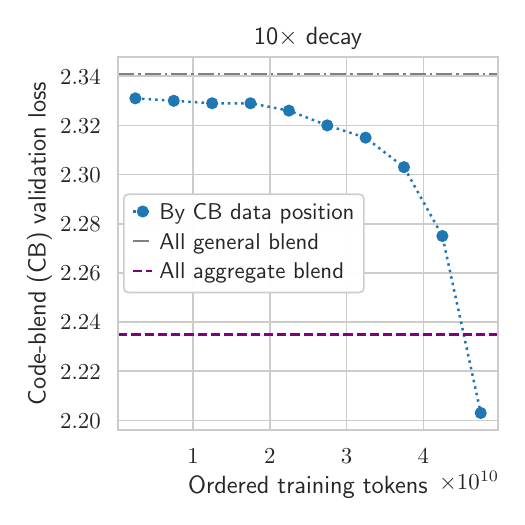}
    \end{minipage}\hfill
    \begin{minipage}{0.33\textwidth}
        \includegraphics[trim={0.3cm 0.4cm 0.264cm 0.3cm}, clip, width=\linewidth]{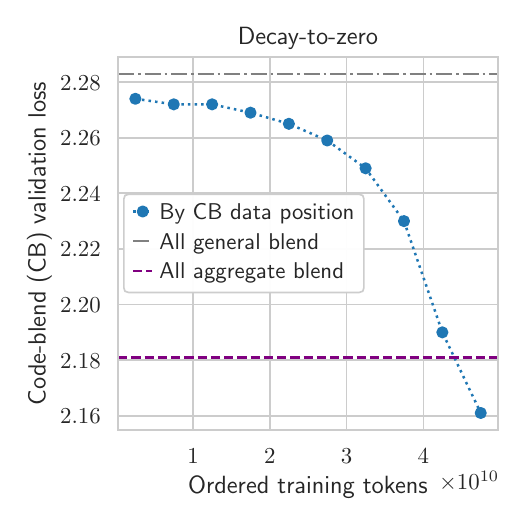}
    \end{minipage}
    \caption{\textbf{Code blend (CB) validation loss across different
        LR schedules and CB training-data placements.} Same result
      data as in \cref{fig:hook} ($\leftfig$), \cref{fig:placements},
      and \cref{fig:d2zplacements}. $\Leftfig$: $\step$ decay
      schedule, $\Middlefig$: $\tenx$ decay, $\Rightfig$: $\dtoz$.
      All 30 models that are trained with data placement improve over
      no placement at all (``All general blend'' line), while placing
      in the optimal TREC-guided position always improves over the
      aggregate blend (``All aggregate blend'' line, i.e., mixing the
      code blend uniformly across all training steps).
      \label{fig:moreplacements}}
\end{figure}

\begin{figure}[ht]
    \centering
    \begin{minipage}{0.33\textwidth}
        \includegraphics[trim={0.3cm 0.4cm 0.264cm 0.3cm}, clip, width=\linewidth]{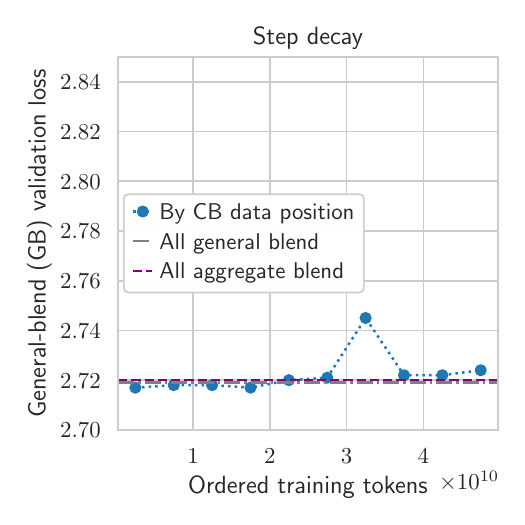}
    \end{minipage}\hfill
    \begin{minipage}{0.33\textwidth}
        \includegraphics[trim={0.3cm 0.4cm 0.264cm 0.3cm}, clip, width=\linewidth]{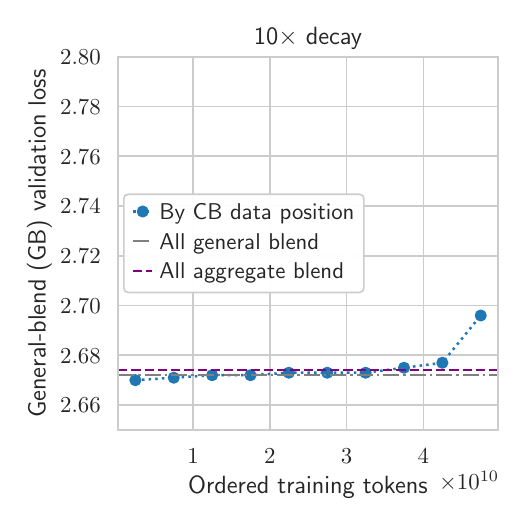}
    \end{minipage}\hfill
    \begin{minipage}{0.33\textwidth}
        \includegraphics[trim={0.3cm 0.4cm 0.264cm 0.3cm}, clip, width=\linewidth]{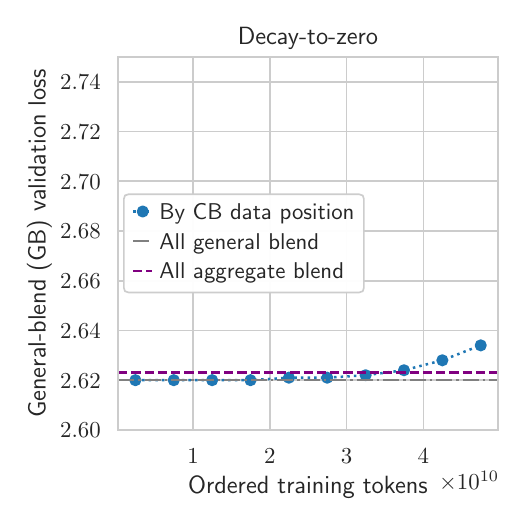}
    \end{minipage}
    \caption{\textbf{\emph{General blend} (GB) validation loss across
        different LR schedules and \emph{code-blend} training-data
        placements.} Counterpoint to \cref{fig:moreplacements} (same
      trained models), except now the y-axis gives \emph{loss on the
      \textbf{general blend}}. $\Leftfig$: $\step$ decay schedule,
      $\Middlefig$: $\tenx$ decay, $\Rightfig$: $\dtoz$.  Placing code
      blend (CB) data at the TREC minimum significantly impairs
      performance on the general blend.
      \label{fig:gbmoreplacements}}
\end{figure}

\cref{fig:moreplacements} isolates the results of the placement tests.
Here, the y-axis gives the code blend (CB) validation loss.  In each
case, the \emph{aggregate blend}, corresponding to a uniform mix of
the code and general data, is a strong baseline: it is only bested by
placing the code data at the optimal placement position.

\cref{fig:gbmoreplacements} presents results when all the above
trained models are evaluated on the \emph{general blend} (GB)
validation data, rather than the code blend.  This experiment can be
interpreted as a form of ablation: if we replace GB data with
\emph{non-GB} (i.e., CB) data at a given position during training,
which position leads to the greatest \emph{degradation} in performance
on GB validation (i.e., where \emph{not} to place)? In effect, we are
testing the \emph{necessity} of GB data at each position by observing
the impact of its omission. The greatest degradation consistently
occurs at the position of minimum TREC loss, thus validating the
placement hypothesis (\cref{hyp:placements}) through omission rather
than commission of task-relevant data.

\takeaway{The TREC minimum is also the most important placement
  position for GB data.}

\subsection{Cross-schedule-TREC placement hypothesis}\label{sec:correlation}

\hypothesis{Across \textbf{different} LR schedules, position-wise TREC
  loss predicts the effectiveness of high-quality data placement at
  corresponding positions.\label{hyp:crossschedule}}

This hypothesis extends \cref{hyp:placements}, proposing that TRECs
are not only useful \emph{within} a single training run but that their
\emph{absolute values} are meaningful and comparable \emph{across}
optimization schedules. For example, in the context of
\cref{sec:placements}, the question is: does absolute TREC loss
predict CB validation loss \emph{across} $\step$, $\tenx$, \emph{and}
decay-to-zero LR schedules?

If true, one could use predicted TRECs from multiple learning rate
schedules (or weight decay/batch size configurations) to identify the
globally best HQ data placement \emph{and} optimization
settings---i.e., the position, LR schedule, and other hyperparameters
yielding the lowest TREC loss.
Consider a 5B HQ token budget alongside 45B baseline tokens. One could
scan different schedules, find the decile with the lowest predicted
$\reeval(t)$, and insert HQ data there.  Similarly, if HQ data arrives
late (e.g., recent web data to ``advance the model's knowledge
cut-off''~\citep{dubey2024llama}), one might select the schedule with
the lowest TREC loss in the \emph{final} decile, knowing this will
maximize task performance.

\begin{figure}[ht]
    \centering
    \begin{minipage}{0.66\textwidth}
        \includegraphics[trim={0.3cm 0.4cm 0.264cm 0.3cm}, clip, width=\linewidth]{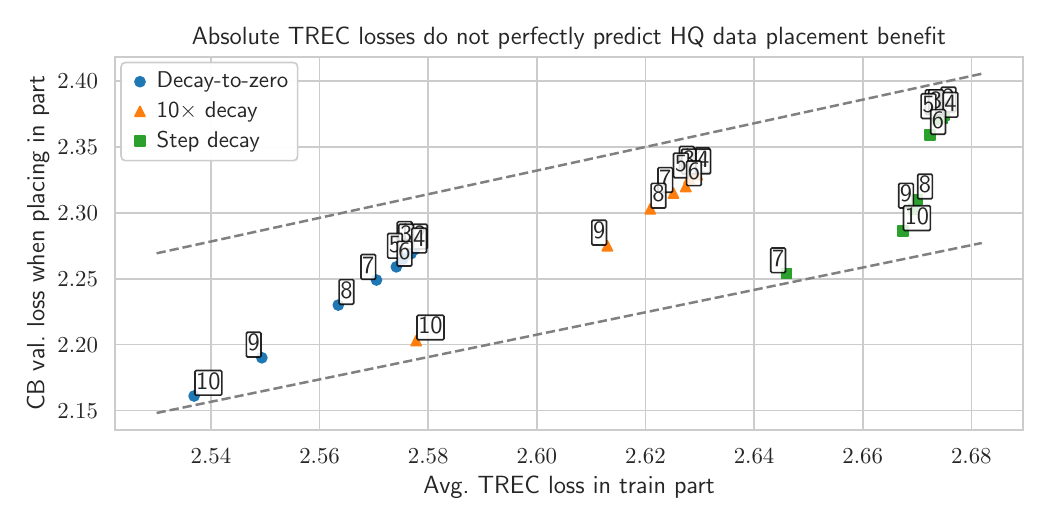}
    \end{minipage}
    \caption{\textbf{Correlation between CB validation loss and
        TREC loss, by placement position.} CB validation loss from
      placement in each decile of training, versus the actual absolute
      TREC loss measured in that decile.  Markers labeled with
      their deciles (1-10), dashed lines to show linear trends.  While
      CB validation varies monotonically with TREC loss
      \emph{within} one LR schedule, the correlation does not hold
      well \emph{across} schedules.
      \label{fig:correlations}}
\end{figure}

\cref{fig:correlations} illustrates this idea. Each marker plots CB
validation loss (y-axis) from placing HQ data in a training decile
against the TREC loss (x-axis) in that decile. Markers are grouped by
LR schedule, with dashed lines showing idealized linear trends.

Within each schedule, TREC loss monotonically predicts validation
performance, consistent with \cref{hyp:placements}. But across
schedules, the alignment breaks down. For example, placing HQ data in
the final decile of the $\tenx$ schedule yields lower validation loss
than many decay-to-zero placements, despite higher TREC loss.
Furthermore, large TREC differences sometimes translate to minor
validation gains, and vice versa, indicating TREC loss alone does
not capture all relevant dynamics.

We revisit this in \cref{sec:app_demo}, where our 3.9B experiments
show a similar disconnect: CPT segments with the lowest TREC loss
(typically under high LR) do not always yield the best validation loss
when HQ data is inserted.  Interestingly, while TREC loss \emph{on the
placed data itself} matches TREC predictions (from the general blend),
the gains do not translate to the held-out validation sets.
It is worth re-iterating, however, that this generalization gap only
occurs across optimizer configurations (e.g., different LRs or LR
schedules), not within a given configuration.

\takeaway{While TRECs reliably guide HQ placement within a given
  optimizer configuration, their absolute values do not consistently
  predict optimal placement \emph{across} configurations. Further work
  is needed to guide schedule selection for data placement.}

\paragraph{Toward an explanation of the generalization gap.}

We can summarize the gap as follows: two schedules A and B may attain
the same TREC loss at certain positions, yet A may generalize better
than B.\@ One possible explanation is that A and B can achieve similar
TREC loss via qualitatively different mechanisms. Recent grokking
studies~\citep{power2022grokking,morris2025how} show that models
initially reduce training loss through memorization, and only later
transition into grokking, where they learn structure that
generalizes. Both mechanisms can yield equally low TREC loss, but
grokking will produce stronger validation performance.

This provides a natural interpretation of why placement at position 10
of the $\tenx$ schedule generalizes better than early decay-to-zero
placement in \cref{fig:correlations}, even though both achieve similar
TREC losses: in the tenth decile of training, the $\tenx$ schedule
evidently leverages more grokking and less memorization. Likewise, in
the CPT LR sweeps, high LR appears to push the model back into a
memorization regime, while low LR preserves grokked structure and
yields better validation loss.

These observations connect to classical factors known to influence
generalization (such as curvature, SGD noise, and basin geometry),
which may vary systematically across LR schedules. If we can identify
signals that distinguish memorization-driven vs.\ grokking-driven TREC
improvements (e.g., the width of the TREC valley, or deviations
between the TREC and the standard training-loss curve), we could
potentially correct for these generalization differences and enable
cross-schedule prediction.

\section{Further TREC results}

This section provides supporting analysis for the TREC behavior
discussed in \cref{sec:ema} of the main paper. We expand on trends
observed across model scale, dataset size, and tokens-per-parameter
(TPP), and include additional plots that quantify absolute loss drops
and timescale effects. Our findings reinforce the role of the AdamW
timescale $\tema$ in shaping TRECs and offer deeper insight into how
training dynamics evolve across compute regimes.

\subsection{Further scaling results}\label{sec:app_scaling}

\begin{figure}[ht]
    \centering
    \begin{minipage}{0.33\textwidth}
        \includegraphics[trim={0.3cm 0.4cm 0.264cm 0.3cm}, clip, width=\linewidth]{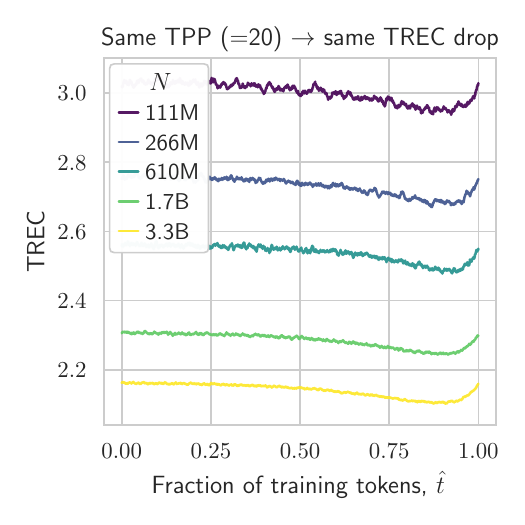}
    \end{minipage}\hfill
    \begin{minipage}{0.33\textwidth}
        \includegraphics[trim={0.3cm 0.4cm 0.264cm 0.3cm}, clip, width=\linewidth]{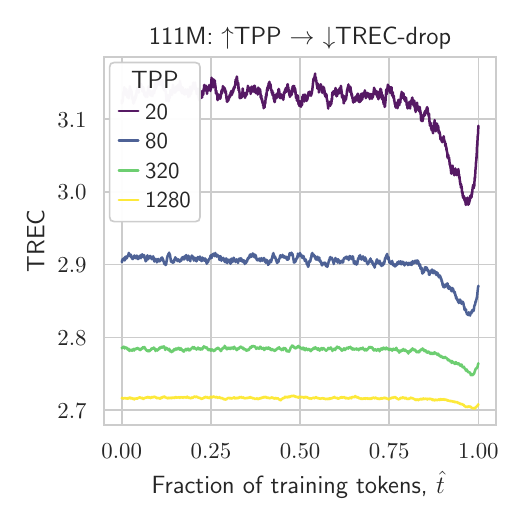}
    \end{minipage}\hfill
    \begin{minipage}{0.33\textwidth}
        \includegraphics[trim={0.3cm 0.4cm 0.264cm 0.3cm}, clip, width=\linewidth]{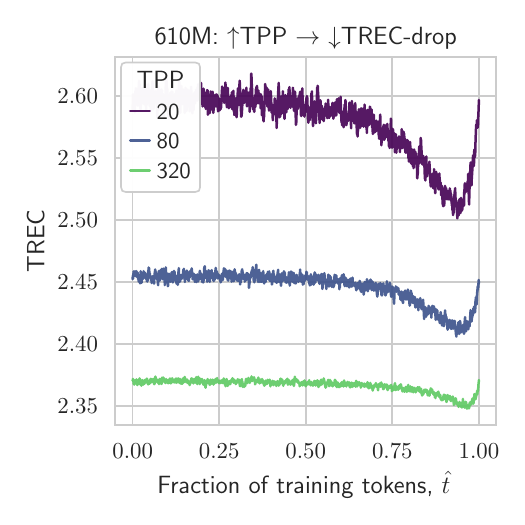}
    \end{minipage}
    \caption{\textbf{Absolute magnitude of TREC drops decrease with TPP.} Plots show absolute unnormalized TRECs, same data as in \cref{fig:scaling}:
      $\Leftfig$: $\tepoch \approx 0.3$; TPP is 20 for all model scales, and absolute magnitudes of drops are \emph{similar}.
      $\Middlefig$: $\tepoch = 0.021$, all models 111M, and magnitude of drop \emph{decreases} with TPP.\@ 
      $\Rightfig$: $\tepoch = 0.105$, all models are 610M, and magnitude of drop again decreases with TPP.\@
      \label{fig:morescaling}}
\end{figure}

The main paper mainly focused on the \emph{shape} and \emph{position}
of TREC valleys, as these are most pertinent for optimal data
placement.  As part of those findings, we found that $\tema$ and
$\tpp$ both modulate the shape of the TRECs.  We now examine the
\emph{absolute magnitude} of the TREC loss drops. As shown in
\cref{fig:morescaling}, when models are trained at constant TPP (e.g.,
20), the overall TREC trajectories exhibit similar total drops
across scales ($\leftfig$). However, when we increase TPP while
holding model scale fixed ($\middlefig$ and $\rightfig$ panels), the
absolute TREC drop shrinks.

This behavior is notable because it raises the hypothesis that at high
TPP, models become more inertial or \emph{rigid}---possibly due to
saturation or reduced plasticity---absorbing less signal per training
fraction. This aligns with recent findings on overtraining and reduced
update effectiveness at high compute budgets (e.g.,
\citet{kumar2024scaling,springer2025overtrained}). From another
perspective, it could also mean that lower-TPP training focuses more
on memorization, and fitting particular training examples, consistent
with other recent findings~\citep{morris2025how,jelassi2024mixture},
as discussed in \cref{sec:ema} and \cref{sec:moe}.

\begin{figure}[ht]
    \centering
    \begin{minipage}{0.33\textwidth}
        \includegraphics[trim={0.3cm 0.4cm 0.264cm 0.3cm}, clip, width=\linewidth]{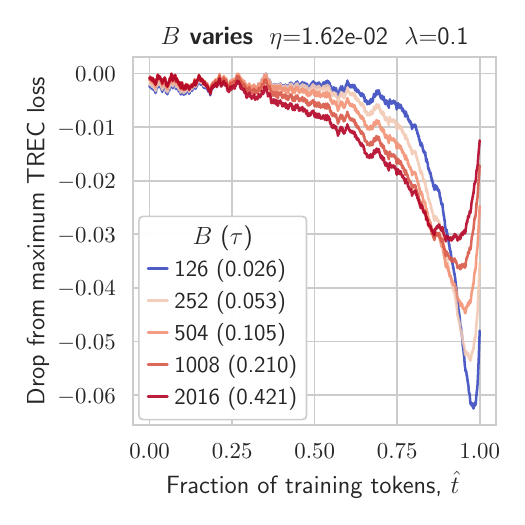}
    \end{minipage}\hfill
    \begin{minipage}{0.33\textwidth}
        \includegraphics[trim={0.3cm 0.4cm 0.264cm 0.3cm}, clip, width=\linewidth]{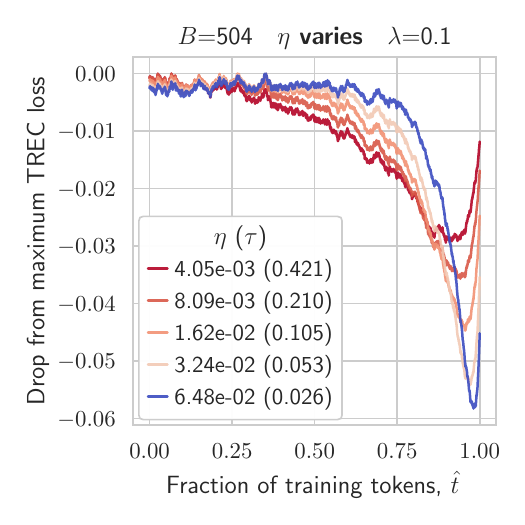}
    \end{minipage}\hfill
    \begin{minipage}{0.33\textwidth}
        \includegraphics[trim={0.3cm 0.4cm 0.264cm 0.3cm}, clip, width=\linewidth]{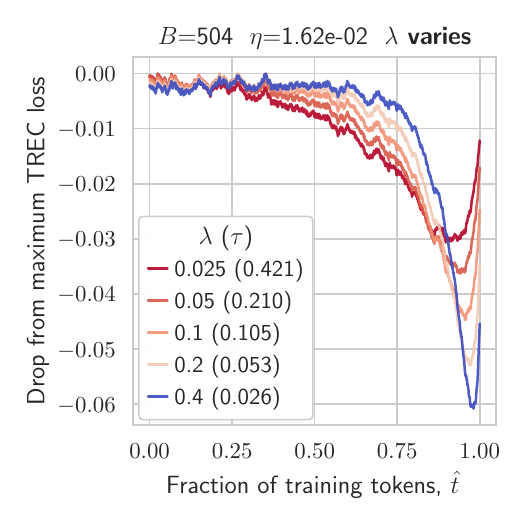}
    \end{minipage}
    \caption{\textbf{Total summed TREC drop invariant to $\tema$
        (610M, 80~TPP).} Same data as in \cref{fig:tema}, but showing
      absolute differences between the per-step and overall-maximum
      TREC loss.  Within and across sweeps of $B$ ($\leftfig$), $\eta$
      ($\middlefig$), or $\lambda$ ($\rightfig$) the total summed TREC
      drop is fairly consistent. Thus the more narrow the timescale,
      the larger the drop.
      \label{fig:topdrop}}
\end{figure}

\cref{fig:topdrop} shows that even though the shape of the TREC shifts
with $\tema$, the \emph{total} (summed) TREC drop across training
steps is fairly consistent (for a given TPP regime). Together with the
above findings, this suggests that $\tema$ governs the \emph{width} of
the valley (how long data influences the model), while TPP sets its
average \emph{depth} (how strongly the model responds).

The upshot is that compute-efficient training regimes---like 20 TPP,
where TREC valleys are sharp---may benefit most from intelligent
data placement.\footnote{\citet{hoffmann2022empirical} found the
optimal model size $\nopt$ and dataset size $\dopt$ to scale roughly
equally as compute increases, with the optimal $\nicefrac{D}{N}$ ratio
around 20 TPP\@.  Further studies have found similar
results~\citep{besiroglu2024chinchilla,porian2024resolving}, and 20
TPP has become synonymous with compute-optimal
training~\citep{dey2023cerebras,zhang2024how}.  Starting with
Llama~\citep{touvron2023llama}, released models are often trained for
more than 20~TPP because smaller, overtrained models are more
efficient for \emph{inference}.}
Note such regimes are often used for frontier-scale training due to
their compute-efficiency. MoEs are also now often used for
frontier-scale efforts, and in MoEs, expert parameters can also see
relatively-low \emph{effective} TPP (\cref{sec:moe}).
TREC-guided data curriculums are therefore likely to offer benefits to
frontier-scale training going forward.

\subsection{Batch size}\label{sec:app_batchsize}

We now investigate how TRECs behave across a wide range of batch
sizes.  In particular, we study regimes well above the \emph{critical
batch size}
$\bcrit$~\citep{mccandlish2018empirical,shallue2019measuring,merrill2025critical}---the
point beyond which increasing batch size significantly degrades loss
as a function of total tokens trained.

Following \citet{bergsma2025power}, we estimate $\bcrit \approx 2150$
tokens for 610M models trained to 20 TPP (via a fitted power law in
training tokens), and we sweep batch sizes from 63 to 8064 (over two
orders of magnitude). In each case, we adjust weight decay to maintain
a constant AdamW timescale of $\tepoch = 0.421$, allowing us to
isolate batch size effects from timescale variation.

\begin{figure}[ht]
    \centering
    \begin{minipage}{0.33\textwidth}
        \includegraphics[trim={0.3cm 0.4cm 0.264cm 0.3cm}, clip, width=\linewidth]{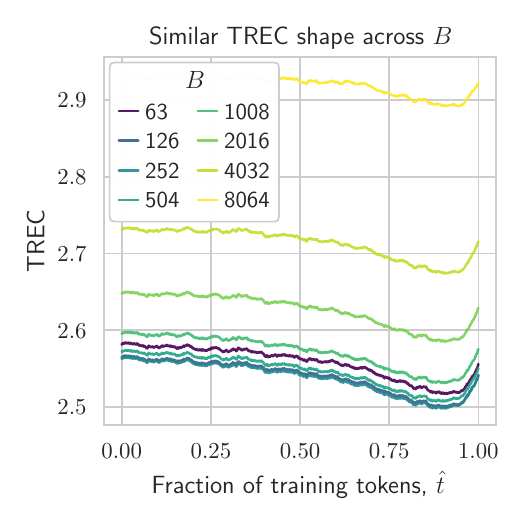}
    \end{minipage}
    \hspace{\twofighspace}
    \begin{minipage}{0.33\textwidth}
        \includegraphics[trim={0.3cm 0.4cm 0.264cm 0.3cm}, clip, width=\linewidth]{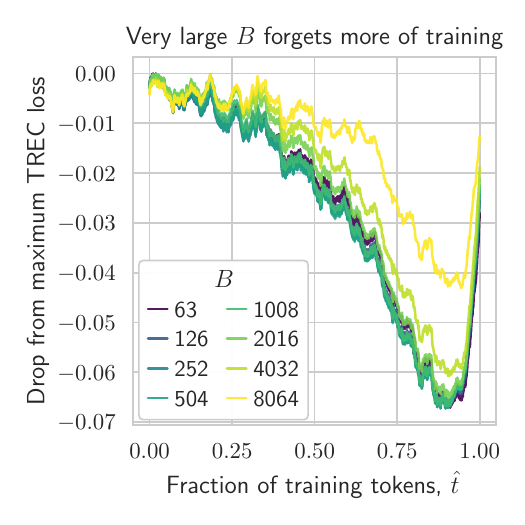}
    \end{minipage}
    \caption{\textbf{TRECs and batch size} (610M, 20~TPP, $\tepoch =
      0.421$).
      $\Leftfig$: Absolute TREC losses across batch sizes; very large
      batches achieve worse loss, but patterns are somewhat similar.
      $\Rightfig$: Absolute TREC losses, but normalized so they have
      same maximum value; $B=8064$ behaves quite differently than all
      other settings.
      \label{fig:batches}}
\end{figure}

\cref{fig:batches} presents the resulting TRECs. The left panel shows
the absolute loss curves: as batch size increases, the total TREC drop
seems to become somewhat shallower, especially once batch size exceeds
$\bcrit$. The right panel aligns curves by their maximum values in
order to compare shapes directly. For $B \leq 2016$, the curves remain
remarkably similar in shape and position. However, for $B = 8064$
(well beyond $\bcrit$), the curve diverges notably, indicating a
qualitatively different training dynamic.

This divergence aligns with theoretical expectations: as $B$ exceeds
$\bcrit$, gradient estimates become increasingly redundant, reducing
the marginal utility of each new example. From a TREC perspective,
gradients from individual samples have less impact on the overall
update, which focuses more on common features than idiosyncrasies of
individual batches.

Viewed through the lens of TRECs, these results offer a novel
diagnostic perspective on the batch size scaling frontier. These
findings also suggest that models trained far beyond $\bcrit$ may
benefit less from careful data placement, as individual batches
contribute less distinct signal to model updates.  On the other hand,
in situations where we might wish to avoid memorization
(\cref{sec:limitations}), using larger batches could help accomplish
this objective.

\takeaway{TRECs remain stable across batch sizes up to $\bcrit$, but
  diverge significantly beyond it, highlighting diminishing marginal
  data influence in large-batch regimes.}
       
\subsection{Adam momentum $\beta_1$ and velocity $\beta_2$}\label{sec:betas}

\begin{figure}[ht]
    \centering
    \begin{minipage}{0.33\textwidth}
        \includegraphics[trim={0.3cm 0.4cm 0.264cm 0.3cm}, clip, width=\linewidth]{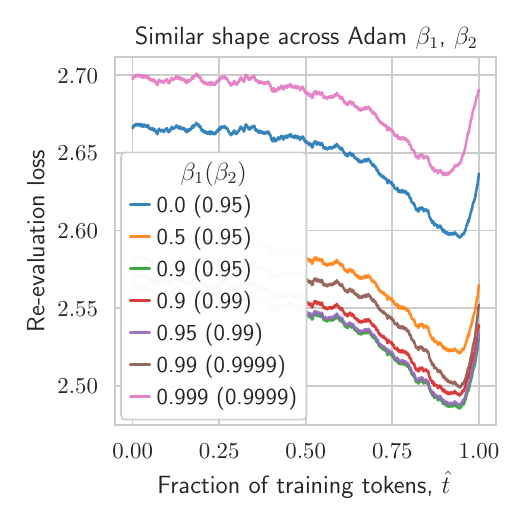}
    \end{minipage}
    \hspace{\twofighspace}
    \begin{minipage}{0.33\textwidth}
        \includegraphics[trim={0.3cm 0.4cm 0.264cm 0.3cm}, clip, width=\linewidth]{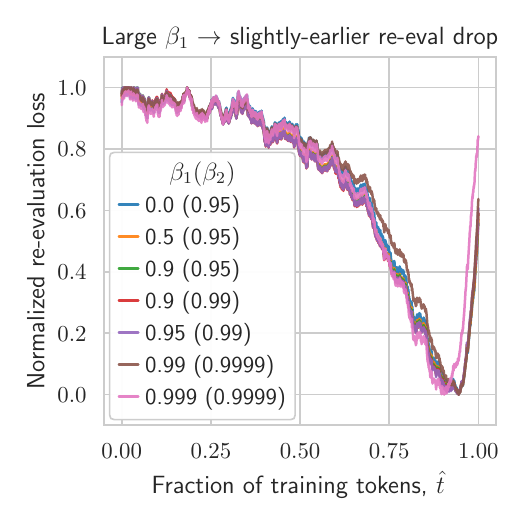}
    \end{minipage}
    \caption{\textbf{TRECs and Adam $\beta_1$ and $\beta_2$} (610M,
      20~TPP, $\tepoch = 0.210$).
      $\Leftfig$: Absolute TREC losses for a range of $\beta_1$ and
      $\beta_2$ settings.  Both no momentum ($\beta_1 = 0.0$) and too
      much momentum ($\beta_1 = 0.999$) have highest absolute loss.
      $\Rightfig$: Normalized TREC losses; despite wide variation in
      absolute loss, normalized TREC shape is remarkably similar
      across all settings, with only very large momentum beginning to
      slightly shift the TREC minimum to earlier in training.
      \label{fig:betas}}
\end{figure}

To assess whether TREC shape is driven by other timescales in the
optimizer, we sweep the AdamW momentum ($\beta_1$) and velocity
($\beta_2$) parameters while holding other hyperparameters fixed
($\tepoch = 0.210$, 610M model, 20 TPP).  We tried a variety of
settings, and report those that completed training successfully (i.e.,
without failure due to numerical instabilities).

\cref{fig:betas} shows that, although absolute TREC losses vary with
$\beta_1$ and $\beta_2$ (especially at extremes such as $\beta_1 = 0$
or $\beta_1 = 0.999$), TREC \emph{shape} remains remarkably consistent
across settings. This is especially evident in the right panel, where
curves are normalized. Only very large momentum begins to shift the
TREC minimum to slightly earlier in training.

These results reinforce the conclusion from the main text: while
standard settings of $\beta_1$ and $\beta_2$ affect training dynamics
and absolute loss, they do not significantly alter TREC shape. The
AdamW timescale (via $\tepoch$) appears to be the dominant factor
shaping TREC trajectories.

\takeaway{TREC shape is largely invariant to changes in $\beta_1$ and
  $\beta_2$, underscoring the dominant role of the AdamW timescale in
  shaping learning dynamics.}

\section{Further prediction results}

In this section, we provide additional details on our predictive
framework for TRECs (\cref{sec:predict}), including the specific
evaluation metrics used, derivations underlying key equations, and
further analysis across model scales and LR schedules.

\subsection{Further prediction details}\label{sec:pred_details}

When computing predictions from our analytical framework
(\cref{eqn:prediction}), we discard the initial EMA coefficient $c_0$,
as it corresponds to the influence of the model's random
initialization rather than any datapoint observed during training. We
focus on the remaining coefficients $c_i$ for $i \geq 1$, which
quantify the contribution of training updates to the final model
weights.

\paragraph{Pearson correlation ($\pearsonr$).}

Since we aim to predict the \emph{shape} of TRECs
$\reeval(\trainfrac)$ rather than absolute scale, we evaluate
prediction quality using a scale- and shift-invariant metric: the
Pearson correlation $\pearsonr$ between the predicted curve
$\hatreeval(\trainfrac)$ and the true curve $\reeval(\trainfrac)$. All
plotted curves are similarly normalized to emphasize shape agreement.

Empirically, we found that Pearson correlation better aligned with
human judgments of prediction quality than alternatives like $\ell_2$
loss, $\Rtwo$, or MSE. It is computed as follows:
\begin{equation}
\pearsonr = \frac{\sum_t (\reeval(t) - \barreeval) (\hatreeval(t) - \barhatreeval)}{\sqrt{\sum_t (\reeval(t) - \barreeval)^2} \cdot \sqrt{\sum_t (\hatreeval(t) - \barhatreeval)^2}},
\end{equation}
where $\barreeval$ and $\barhatreeval$ are the means of the true and
predicted TREC values, respectively.

This measure ranges from $-1$ (perfect inverse correlation) to $+1$
(perfect match), with $0$ indicating no correlation.

\paragraph{Illustrative example.}
The middle panel of \cref{fig:preds} illustrates the full prediction
process for a $\cyclic$ LR schedule. (For reference,
\cref{fig:cyclic_wsd}, $\bottomleftfig$, shows the full LR
trajectory.)  We compute EMA coefficients $c(\trainfrac)$ from the LR
and weight decay schedule (\cref{fig:preds}, $\leftfig$), apply
\cref{eqn:prediction} with a fitted exponent $m$ to obtain
$\hatreeval(\trainfrac)$ (\cref{fig:preds}, $\middlefig$), and compare
it to the actual TREC (\cref{fig:preds}, $\rightfig$). The predicted
curve tracks the shape of the true TREC, particularly in later
training stages where the EMA contribution becomes more pronounced.

\subsection{Schedule history and TRECs: \cyclic{} vs.\ \wsd{}}\label{sec:wsd_cyclic_alignment}

In this section, we study the question: does the full LR schedule
history matter, or is late-stage alignment of EMA and $\trainfrac$
sufficient to determine TREC shape?

\paragraph{Motivation.}
Our predictive framework assumes that TREC shape is primarily
determined by the EMA coefficients and the training fraction
(\cref{eqn:prediction}). However, real-world schedules can differ
substantially in their early phases while converging later. To test
the extent to which \textbf{schedule history leaves residual effects}
on TRECs, we directly compare $\cyclic$ and $\wsd$ schedules under
controlled conditions.

\paragraph{Experimental setup.}
We train 610M-parameter models to 80 TPP on SlimPajama using identical
peak LR, batch size, weight decay, and dataset. The only difference is
the LR schedule:
\begin{itemize}
    \item The $\wsd$ schedule warms up, then maintains a long flat LR phase, before decaying to zero.
    \item The $\cyclic$ schedule oscillates but aligns with $\wsd$ in the final 20\% of training.
\end{itemize}
Because the batch size and dataset are fixed, the training fraction
$\trainfrac$ at each step matches across schedules.

\begin{figure}[ht]
    \centering
    \begin{minipage}{0.33\textwidth}
        \includegraphics[trim={0.3cm 0.4cm 0.264cm 0.3cm}, clip, width=\linewidth]{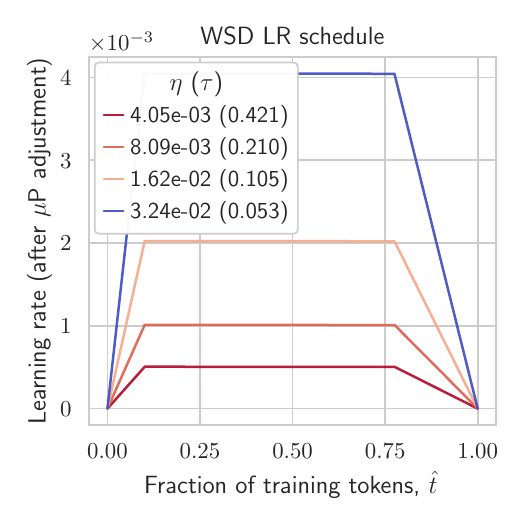}
    \end{minipage}\hfill
    \begin{minipage}{0.33\textwidth}
        \includegraphics[trim={0.3cm 0.4cm 0.264cm 0.3cm}, clip, width=\linewidth]{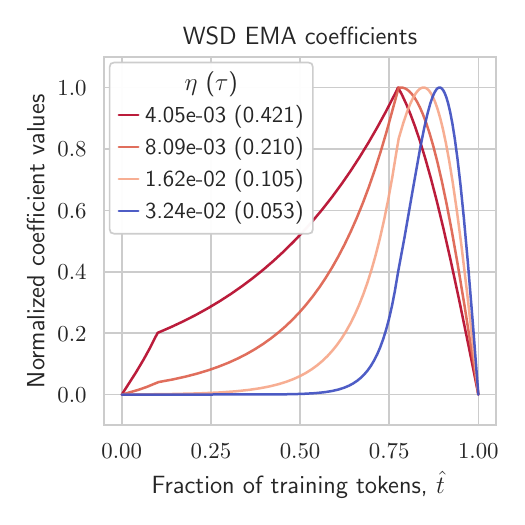}
    \end{minipage}\hfill
    \begin{minipage}{0.33\textwidth}
        \includegraphics[trim={0.3cm 0.4cm 0.264cm 0.3cm}, clip, width=\linewidth]{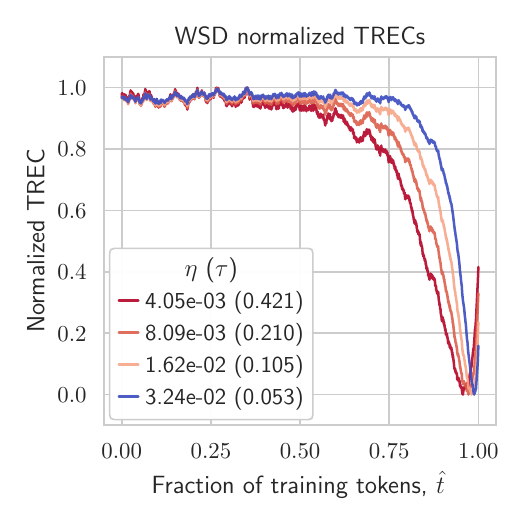}
    \end{minipage} \\
    \begin{minipage}{0.33\textwidth}
        \includegraphics[trim={0.3cm 0.4cm 0.264cm 0.3cm}, clip, width=\linewidth]{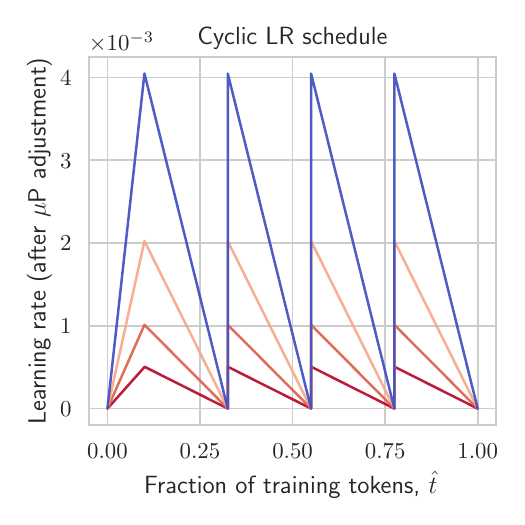}
    \end{minipage}\hfill
    \begin{minipage}{0.33\textwidth}
        \includegraphics[trim={0.3cm 0.4cm 0.264cm 0.3cm}, clip, width=\linewidth]{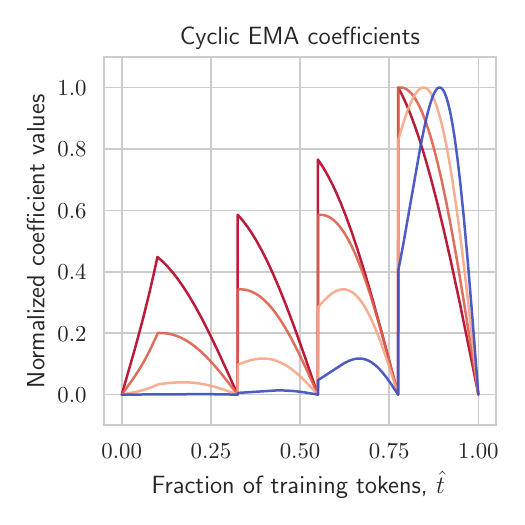}
    \end{minipage}\hfill
    \begin{minipage}{0.33\textwidth}
        \includegraphics[trim={0.3cm 0.4cm 0.264cm 0.3cm}, clip, width=\linewidth]{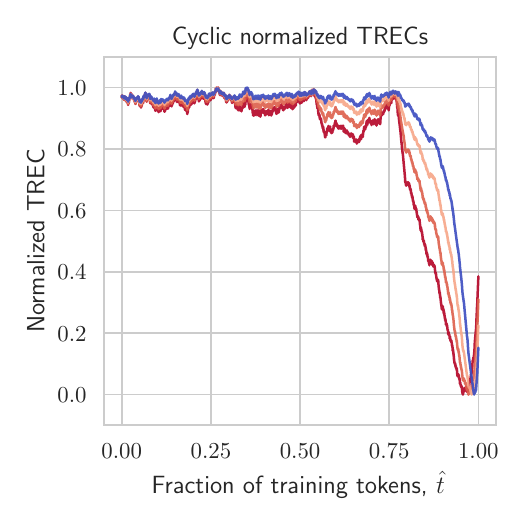}
    \end{minipage}
    \caption{\textbf{Comparison of \cyclic{} and \wsd{} schedules (610M, 80~TPP).}
      Top row: Using \wsd{} schedules; bottom row: Using \cyclic{} schedules.
      $\Leftfig$: \emph{LR schedule}. Training shares the
      same batch size, weight decay, and dataset, differing
      only in LR function (cyclic vs.\ WSD).
      $\Middlefig$: \emph{Corresponding EMA coefficients} $c(\trainfrac)$. In
      the final 20\% of training, schedules align in LR decay,
      resulting in identical EMA coefficients.
      $\Rightfig$: \emph{TRECs}. In the final portion of training
      where EMA and training fraction align, TRECs also
      align closely, indicating that \textbf{EMA
        coefficients and training fraction, rather than prior LR
        schedule history, predominantly determine TREC shape.}
      \label{fig:cyclic_wsd}}
\end{figure}

\paragraph{Results.}
\cref{fig:cyclic_wsd} shows LR schedules ($\leftfig$), EMA
coefficients ($\middlefig$), and true TRECs ($\rightfig$) for both
schedules (top: \wsd{}, bottom: \cyclic{}). In the final 20\% of
training---where LR schedules align---the \emph{EMA coefficients}
converge exactly. Correspondingly, after a brief transient period, the
\emph{TRECs} also align, despite the differences in earlier schedule
history.

In summary:
\begin{itemize}
    \item When EMA and $\trainfrac$ align, TRECs align---even across
      different LR histories.
    \item Prior LR fluctuations leave negligible residual effect on
      TREC shape once EMA and $\trainfrac$ match.
\end{itemize}

\takeaway{These results reinforce that given similar amounts of
  pre-training, \textbf{EMA coefficients and training fraction are
    sufficient to determine TREC shape}, supporting the generality of
  our predictive framework across schedules.}

\subsection{Adam as the limit of AdamW when $\lambda \rightarrow 0$}\label{sec:adamw_limit}

We now explain why TRECs converge as $\lambda \to 0$, or equivalently,
as the AdamW timescale $\tepoch = 1/\lambda \to \infty$. In this
regime, the effect of weight decay vanishes, and AdamW behavior
approaches that of standard Adam.

\cref{fig:hook} ($\middlefig$) illustrates this convergence at 20~TPP:
the TREC for $\lambda = 0$ (vanilla Adam) closely resembles that for
$\lambda = 0.001$. Since both TPP and $\tepoch$ are fixed, the
training-fraction term $\trainfrac^m$ in \cref{eqn:prediction} remains
unchanged across settings, so convergence must arise from changes in
the EMA coefficients $c(\trainfrac)$.

\begin{figure}[ht]
    \centering
    \begin{minipage}{0.33\textwidth}
        \includegraphics[trim={0.3cm 0.4cm 0.264cm 0.3cm}, clip, width=\linewidth]{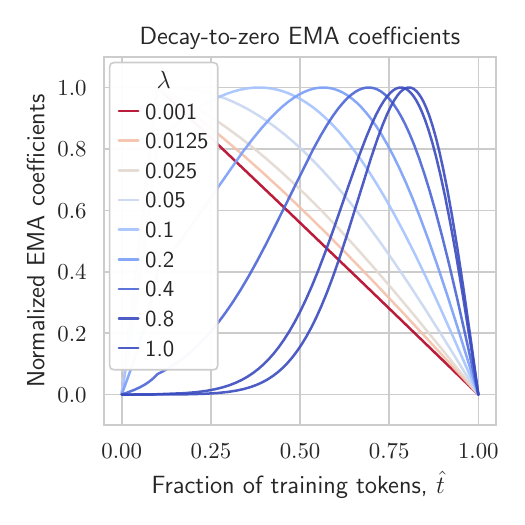}
    \end{minipage}\hfill
    \begin{minipage}{0.33\textwidth}
        \includegraphics[trim={0.3cm 0.4cm 0.264cm 0.3cm}, clip, width=\linewidth]{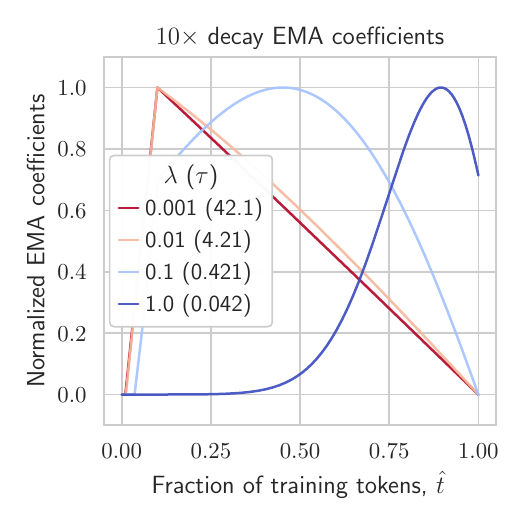}
    \end{minipage}\hfill
    \begin{minipage}{0.33\textwidth}
        \includegraphics[trim={0.3cm 0.4cm 0.264cm 0.3cm}, clip, width=\linewidth]{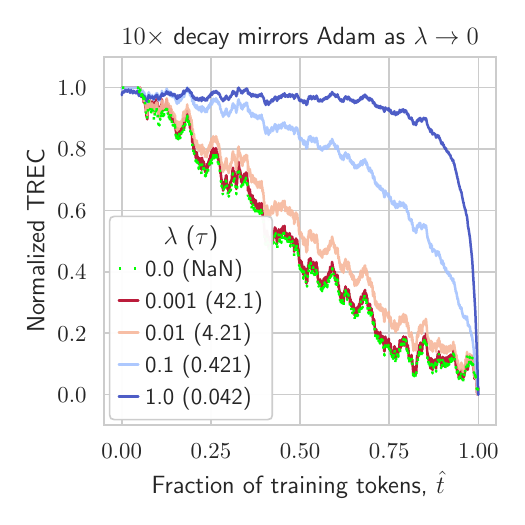}
    \end{minipage}
    \caption{\textbf{Effect of weight decay on EMA coefficients and TRECs.}
      $\Leftfig$: Normalized EMA coefficients for the decay-to-zero LR
      schedule across decreasing $\lambda$ values. As $\lambda
      \rightarrow 0$, the EMA coefficients flatten toward the shape of
      the LR schedule (coefficients undefined at $\lambda =
      0$). Corresponding TRECs are shown in \cref{fig:hook},
      $\middlefig$.
      $\Middlefig$: EMA coefficients for the $\tenx$-decay
      schedule. As weight decay decreases, the curve approaches the
      $\lambda = 0$ (Adam) case.
      $\Rightfig$: TRECs under the $\tenx$-decay schedule. As
      $\lambda \rightarrow 0$, TRECs converge to the $\lambda = 0$
      baseline, confirming that TREC shape becomes increasingly
      determined by the LR schedule alone.
      \label{fig:wd}}
\end{figure}

\cref{fig:wd} provides direct evidence. As shown in the $\leftfig$,
decreasing $\lambda$ causes the EMA coefficients to flatten toward the
shape of the LR schedule. We observe this convergence behavior both
for the decay-to-zero schedule ($\leftfig$) and the $\tenx$ decay
schedule ($\middlefig$), with corresponding TRECs for $\tenx$ shown in
the $\rightfig$ panel.

In fact the EMA convergence can be derived mathematically.

\paragraph{Derivation: $\boldsymbol{c_i \rightarrow \eta_i}$ as $\boldsymbol{\lambda \rightarrow 0}$.}
\label{subsec:wd_derivation}

To clarify the behavior of the EMA coefficients as $\lambda
\rightarrow 0$ (and thus $\tau \rightarrow \infty$), it is helpful to
separate the \emph{scale} of the learning rate from its \emph{shape}.
We write the learning-rate schedule as
\[
\eta_t = \gamma \, \hat{\eta}_t,
\]
where $\gamma$ captures the overall scale (e.g., the peak learning
rate) and $\hat{\eta}_t \in [0,1]$ denotes the normalized schedule
shape.

Under this parameterization, the EMA coefficients can be written (up
to an overall scale factor) as
\[
\hat{c}_i
=
\hat{\eta}_i
\prod_{j=i+1}^{T}
\left(1 - \gamma \lambda \hat{\eta}_j \right),
\]
where $\hat{c}_i$ denotes the coefficients with scale factors divided
out.

This form makes the limiting behavior transparent: as $\lambda
\rightarrow 0$, the product term approaches $1$, and therefore
\[
\hat{c}_i \;\longrightarrow\; \hat{\eta}_i.
\]
Thus, in the zero--weight-decay limit, the temporal structure of the
coefficients is governed entirely by the learning-rate schedule shape.

An analogous limiting behavior occurs as $\gamma \to 0$ (i.e., as the
overall learning-rate scale $\eta_{\max}$ vanishes), since the
multiplicative term again approaches $1$, yielding $\hat{c}_i \to
\hat{\eta}_i$.

Moreover, the same structural behavior also emerges in the large-batch
regime $B \to \infty$, with $\gamma$ and $\lambda$ held
fixed. Increasing $B$ reduces the total number of optimizer steps $T$,
so the product spans fewer terms and has less opportunity to compound
decay. In the limit, the cumulative effect becomes negligible, and the
coefficients inherit the learning-rate schedule shape.

We thank Fabian Schaipp for suggesting this scale--shape
reparameterization, which clarifies these limiting regimes.

\section{Fitting the Training-Fraction Exponent $\mstar$}\label{sec:fit_mstar}

This section provides additional details about our methodology for
fitting and evaluating the power-law functional form for the
training-fraction exponent $\mstar$ (\cref{eqn:optimal_m}). We
describe how we select fitting data, evaluate prediction quality,
validate generalization across model scales, and assess how the fit
transfers across learning rate schedules.

\subsection{Data Filtering and Fit Criteria}

To ensure robust and meaningful fits, we restrict our fitting dataset
to regimes exhibiting well-behaved training and stable TRECs:

\begin{itemize}
    \item \textbf{Effective $\tema$ range:} We include only runs with
      $\tema$ values between 0.001 and 1.0, corresponding to training
      runs where learning was stable and meaningful signal is present.
    \item \textbf{Excluding unstable hyperparameters:} We discard
      configurations with extremely high or low learning rates or
      weight decay values, which frequently result in divergence, loss
      spikes, or poor convergence.
    \item \textbf{Batch size filtering:} We remove runs where batch
      size exceeds the estimated critical batch size $\bcrit$, beyond
      which training enters a distinct large-batch regime (see
      \cref{sec:app_batchsize}).
\end{itemize}

These filters ensure that the resulting fit captures relationships in
the regime of effective optimization, avoiding pathological outliers.

\subsection{Evaluation Metric: $\Rtwo$ in Log Space}

We quantify the accuracy of our predicted $\mstar$ values using the
coefficient of determination, $\Rtwo$, computed in log space.

Let $\{m_i\}$ denote the true optimal training-fraction exponents,
i.e., those where the prediction has the highest Pearson $\pearsonr$
agreement with the true TRECs. Let $\{\hat{m}_i\}$ be the values
predicted by our power-law fit. Then:

\begin{equation}
\Rtwo = 1 - \frac{\sum_i \left( \log m_i - \log \hat{m}_i \right)^2}
                  {\sum_i \left( \log m_i - \overline{\log m} \right)^2}
\end{equation}

This log-space evaluation accounts for multiplicative relationships
and downweights the influence of very large or small outliers.

\subsection{Fits at other model scales}

\begin{figure}[ht]
  \centering
  \begin{minipage}[t]{0.38\textwidth}
    \vspace{0pt} 
    \centering
    \captionof{table}{\textbf{TREC $\pearsonr$ rises with scale}: even as $\Rtwo$ with true $\mstar$ declines.}
    \label{tab:fit_scaling}
    \small
    \begin{tabular}{@{}ccc@{}}
      \toprule
      Eval scale     & $\mstar$: $\Rtwo$    & TREC: $\pearsonr$ \\ \midrule
      111M           & \textbf{98.9\%} & 96.6\%          \\
      266M           & 97.2\%          & 97.5\%          \\
      610M           & 98.7\%          & 98.4\%          \\
      1.7B           & 89.0\%          & \textbf{98.7\%} \\
      3.3B           & 76.7\%          & 98.6\%          \\ \bottomrule
    \end{tabular}    
  \end{minipage}\hfill
  \begin{minipage}[t]{0.58\textwidth}
    \vspace{0pt} 
    \centering
    \captionof{table}{\textbf{Both $\tpp$ and $\tema$ improve estimate
        of $\mstar$}: fit at 111M scale, evaluation (TRECs \&
      $\mstar$ fits) at 610M.}
    \label{tab:fit_ablation}
    \small
    \begin{tabular}{@{}cccc@{}}
      \toprule
      \multicolumn{2}{@{}c}{Fit of $\mstarfit$}    & \multirow{2}{*}{$\mstar$: $\Rtwo$} & \multirow{2}{*}{TREC: $\pearsonr$} \\
      $\rightarrow$ uses $\tema$   & $\rightarrow$ uses $\tpp$   &  & \\ \midrule
      $\myx$        & $\myx$       & -45.9\%         & 84.1\%         \\
      $\myx$        & $\mycheck$   & 15.1\%          & 90.6\%          \\
      $\mycheck$    & $\myx$       & 90.9\%          & 97.7\%          \\
      $\mycheck$    & $\mycheck$   & \textbf{98.7\%} & \textbf{98.4\%} \\ \bottomrule
    \end{tabular}
  \end{minipage}
\end{figure}

\begin{figure}[ht]
    \centering
    \begin{minipage}{0.33\textwidth}
        \includegraphics[trim={0.3cm 0.4cm 0.264cm 0.3cm}, clip, width=\linewidth]{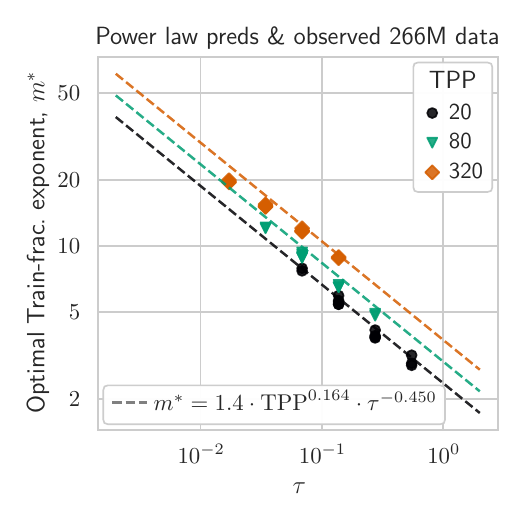}
    \end{minipage}
    \hspace{\twofighspace}
    \begin{minipage}{0.33\textwidth}
        \includegraphics[trim={0.3cm 0.4cm 0.264cm 0.3cm}, clip, width=\linewidth]{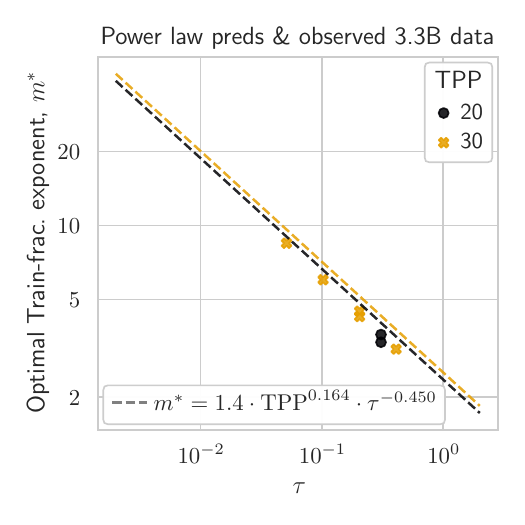}
    \end{minipage}
    \caption{\textbf{Generalization of $\mstar$ power-law fits to
        larger models.}  We evaluate the power-law fit made at 111M
      scale: $\Leftfig$: Fit applied to 266M model; $\rightfig$: Fit
      applied to 3.3B model.  While fit accuracy declines slightly at
      larger scale, TREC predictions remain strong overall.
      \label{fig:morefits}}
\end{figure}

\cref{fig:morefits} extends the evaluation of our $\mstar$ fit to
additional model scales not included in the primary figure
(\cref{fig:fits}). Fit quality slightly degrades at the 3.3B scale.

However, even though the optimal $\mstar$ values exhibit growing
prediction error (\cref{tab:fit_scaling}, middle column $\Rtwo$
results), the predicted TRECs still closely match the true TREC shapes
(\cref{tab:fit_scaling}, right column $\pearsonr$ results)---which is
what we ultimately care about.  We attribute this resilience to
(1)~TREC predictions being robust to exact training fraction
exponents, and (2)~reduced noise in the reference TRECs at larger
model scales, which compensates for slightly-worse $\mstar$
prediction.

\cref{tab:fit_ablation} provides the ablation results mentioned in the
main paper, showing how the power law performs when only a function of
$\tema$, only a function of $\tpp$, or when a function of
neither---i.e., using a constant $m$ tuned over all the fitting data.

\subsection{Fit Generalization Across Learning Rate Schedules}\label{sec:lr_schedule_generalization}

All $\mstar$ power-law fits in the main paper are derived from
training runs using a $\linear$ decay-to-zero LR schedule. We assess
whether these coefficients generalize to other schedules by applying
them to TREC predictions on models trained with a $\cosine$ decay
schedule.

\begin{itemize}
    \item Using the $\linear$-based fit, we achieve a TREC
      prediction accuracy of $\pearsonr = 94.2\%$ on
      $\cosine$-schedule runs.
    \item Fitting a dedicated power-law using $\cosine$ runs improves
      the prediction score to $\pearsonr = 97.8\%$.
\end{itemize}

These results indicate that while the predictive framework retains
strong performance across related schedules, a schedule-specific fit
yields better accuracy. Accordingly, when evaluating prior work that
uses $\cosine$ pre-training schedules (e.g.,
\cref{sec:prior,sec:further_prior}), we employ our $\cosine$-specific
$\mstar$ fit in order to optimize our TREC predictions.

In summary, if a practitioner wishes to use a more complex schedule
(e.g., WSD, or a custom multi-stage schedule), the $\linear$-based fit
can provide strong out-of-the-box guidance.  In particular, this
guidance is far stronger than the default end-of-training heuristic.
For large-scale production training, hyperscalers could re-fit
schedule-specific TREC predictors at small scale (analogous to how
they re-fit scaling laws), and use these instead.  Or, more directly,
obtain a TREC prediction by simply re-evaluating a smaller model with
matching $\tau$, TPP, and LR schedule.

\takeaway{Power-law prediction of $\mstar$ remains robust across model
  scales and related schedules, though schedule-specific fits can
  improve accuracy. Even when $\mstar$ prediction slightly degrades,
  the resulting TREC predictions remain highly accurate.}

\section{Evaluating prior LLM recipes: further details}\label{sec:further_prior}

This section provides methodological details for how we predicted
TRECs for prior LLM training recipes, including Llama-3 (405B), OLMo-2
(13B), Feng et al. (8B), Pangu-Ultra (135B), and Nemotron-4 (15B).

\subsection{Methodology for predicting TRECs of prior LLMs}

\begin{figure}[ht]
    \centering
    \begin{minipage}{0.33\textwidth}
        \includegraphics[trim={0.3cm 0.4cm 0.264cm 0.3cm}, clip, width=\linewidth]{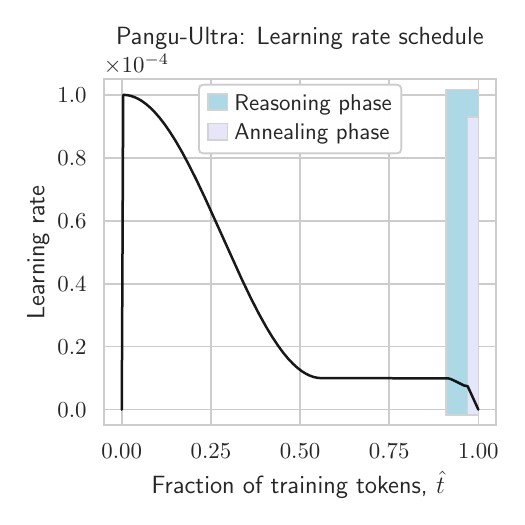}
    \end{minipage}\hfill
    \begin{minipage}{0.33\textwidth}
        \includegraphics[trim={0.3cm 0.4cm 0.264cm 0.3cm}, clip, width=\linewidth]{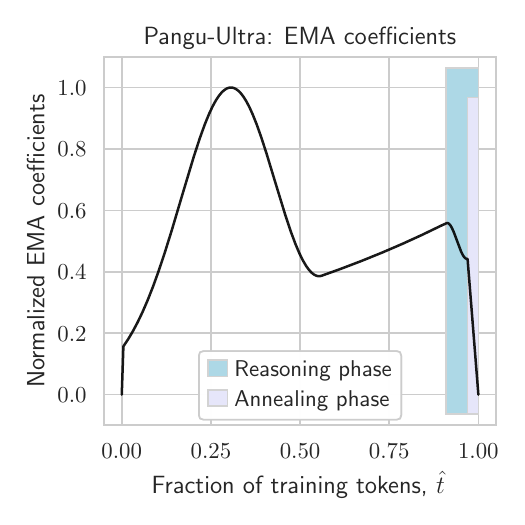}
    \end{minipage}\hfill
    \begin{minipage}{0.33\textwidth}
        \includegraphics[trim={0.3cm 0.4cm 0.264cm 0.3cm}, clip, width=\linewidth]{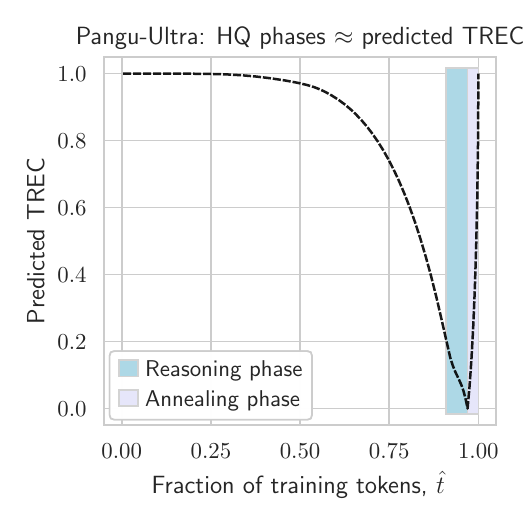}
    \end{minipage}
    \caption{\textbf{Predicted TRECs for the Pangu-Ultra training recipe.}
      $\Leftfig$: LR schedule used in training the Pangu-Ultra
      model~\citep{yin2025pangu}, including distinct ``reasoning'' and
      ``annealing'' phases.
      $\Middlefig$: EMA coefficients computed from the LR schedule,
      batch size, and weight decay.
      $\Rightfig$: Predicted TREC.\@ Both the reasoning and annealing
      phases---where high-quality data is introduced---align with the
      predicted TREC valley, indicating the model is well-positioned
      to retain this curated data.
      \label{fig:pangu}}
\end{figure}

To enable TREC prediction for existing models, we reconstructed the
learning rate (LR) schedules from published training recipes. For
example, \cref{fig:pangu} ($\leftfig$) shows the full LR schedule for
Pangu-Ultra~\citep{yin2025pangu}, including the designated
``reasoning'' and ``annealing'' phases. From these schedules, and
given all these recipes used the AdamW optimizer, we computed the
corresponding AdamW EMA coefficients (e.g., \cref{fig:pangu},
$\middlefig$).

We also extracted dataset size, batch size, sequence length, and
weight decay values from the training documentation. If batch size or
sequence length varied over the course of training (as is sometimes
the case), we used the values applied for the majority of
training. Early-stage variations typically occur well before the
predicted TREC valley and thus have minimal impact on our
prediction. Moreover, in continual-time interpretation, such
variations do not materially affect the TREC prediction once
progress is normalized to training fraction $\trainfrac$.

To determine the appropriate training-fraction exponent $m$, we
applied the power-law fit described in \cref{eqn:optimal_m}, using
models trained with a $\cosine$ decay schedule
(\cref{sec:lr_schedule_generalization}). This was appropriate because
the prior work considered here---Llama-3, Feng et al., OLMo-2,
Pangu-Ultra, and Nemotron-4---all used $\cosine$ LR schedules during
most of pre-training.  However, special mid-training phases that hold
the LR constant or rapidly decay it to zero (as in \cref{fig:pangu})
means that it is not exactly a $\cosine$ schedule that is ultimately
used over all of pre-training.  Yet, while TREC prediction accuracy
can benefit from a schedule-matched $m^\star$, recall that our earlier
analysis has shown:
\begin{itemize}
    \item $m^\star$ is relatively robust to schedule type,
    \item TREC prediction remains accurate even when $m^\star$ is slightly off, and
    \item our goal here is to assess general placement trends, rather than fit precise minima.
\end{itemize}

\subsection{Pangu-Ultra results}\label{sec:pangu}

\cref{fig:pangu} ($\rightfig$) illustrates the predicted TREC
for the Pangu-Ultra training recipe~\citep{yin2025pangu}, based on its
published learning rate schedule and training phases.

The Pangu-Ultra recipe includes two late-stage phases specifically
intended to improve model reasoning and instruction-following
capabilities. As described in the paper:

\begin{quote}
``In the second reasoning phase, we increase the proportion of
  high-quality and diverse mathematical and coding data---raising it
  to over 60\% of the corpus to enhance the reasoning capabilities of
  Pangu Ultra. [...] Moreover, LLM-generated synthetic data is widely
  incorporated to enrich the corpus.

The third annealing phase is designed to help the model consolidate
and effectively apply the knowledge and reasoning skills acquired in
the previous stages. Therefore, we place greater emphasis on
instruction data, [...] [including] carefully refined ... short and
long chain-of-thought responses.''
\end{quote}

As shown in \cref{fig:pangu} ($\rightfig$), both the reasoning and
annealing phases occur during the TREC valley predicted by our
framework. In some sense, this alignment between real-world data
placement and our predicted optimal locations supports the validity of
our framework. However, note the TRECs rise quickly during the
reasoning phase, suggesting that shifting this HQ phase slightly
earlier may have been beneficial. That is, a small timing adjustment
could have helped avoid placing expensively-curated data in the region
where the model has already begun to stabilize, minimizing the risk of
diminished impact.

\subsection{Analysis of Nemotron-CPT~\citep{parmar2024reuse} strategy}\label{sec:nemotron}

A particularly informative paper to interpret through our framework is
the recent work by \citet{parmar2024reuse}. While their study includes
a number of thoughtful ablations, it lacks a unifying theory of data
placement. As such, their findings can benefit from reinterpretation
in light of our insights into TRECs and training curriculums.

The starting point of their study is the Nemotron-4 15B base model,
pre-trained on 8T tokens (533~TPP). Aiming to enhance this model
without restarting training from scratch, they explore a continual
pre-training (CPT) strategy---adding 300B further tokens, including
2.8B high-quality QA tokens mixed into a ``high-quality'' blend.  The
HQ blend is placed at the \emph{end} of the CPT phase. Their main
research questions concern: (1)~the optimal onset point for
introducing the HQ blend during CPT, and (2)~the learning rate
schedule to apply during CPT.

Although this work is framed as CPT, we first note that the original
model was trained using a $\tenx$ LR decay schedule, which is
suboptimal relative to decay-to-zero
($\dtoz$)~\citep{bergsma2025straight}. Indeed, their own results show
$\dtoz$ is a better schedule: continuing with the same pre-training
data distribution but using $\dtoz$ during the 300B token extension
improves downstream accuracy from 48.9\% to 51.5\%---a 5.3\% relative
gain over the base model, using less than 4\% extra data.  In other
words, decaying to zero raises accuracy much more than would be
expected by instead simply extending the $\tenx$ schedule over the
same number of tokens.
While placing HQ data boosts performance further, the total
improvement naturally reflects the \emph{combined} effects of both
$\dtoz$ and HQ data placement.
Crucially, if \citeauthor{parmar2024reuse}\ were to train further on
an additional 300B tokens, they would not benefit again from applying
$\dtoz$---that benefit would already have been consumed.
This is precisely why we define methods that place data \emph{before}
the LR has fully decayed to be \emph{mid-training} rather than true
CPT (\cref{sec:demo}): true CPT methods can be applied repeatedly
(e.g., repeatedly re-warming and re-decaying the LR), while
mid-training benefits, like those in \citet{parmar2024reuse}, can
only be obtained once.

Second, by placing HQ data at the very end of training, their
ablations of LR schedules are intrinsically confounded. For example,
they find that $100\times$ decay performs marginally better than
$\dtoz$ during the ``CPT'' phase. But as we have shown, $\dtoz$
TRECs rise sharply near the end, so late-stage data is unlikely to
be retained.  By not decaying fully, they may make better use of the
(sub-optimally) placed data. Similarly, they observe that the WSD
schedule~\citep{hu2024minicpm,hagele2024scaling} underperforms
compared to $\cosine$ decay, and ``hypothesize that in continued
pre-training, switching the decay schedule from the one used during
pre-training is harmful.''  However, this explanation is doubtful:
different schedules yield different EMA coefficients, and thus
different TRECs. Their effectiveness therefore depends not on
matching schedule shapes, but on whether the HQ data is placed near
the TREC minimum.

Third, their experiments are affected by data repetition, which
further complicates interpretation. Repetition is known to degrade
pre-training quality~\citep{hernandez2022scaling}, and this likely
applies even to high-quality data~\citep{tiifalconh1}. While the
subsequent work by the same team~\citep{feng2024maximize} controlled
for repetition more carefully across training scales, the original
Nemotron-CPT study in \citet{parmar2024reuse} allows repetition to
grow significantly as the size of the CPT phase increases from 100B to
300B to 1T tokens. Specifically, the 2.8B QA tokens represent 10\% of
the HQ blend, implying roughly $10.7\times$ repetition if used for the
full the 300B-token phase and $35.7\times$ in the 1T-token case. Their
results confirm the downside of this: training with HQ data for the
full CPT phase yields 53.6\% accuracy at 100B tokens, but only 52.8\%
at 300B---i.e., longer training \emph{hurts} when it induces excessive
repetition.

For all these reasons, it is unlikely that some of the paper's
specific conclusions will generalize. For instance, their
recommendation that ``the switch [to the high-quality] data
distribution should occur at $\nicefrac{\eta_{\mathrm{max}}}{5}$ in
the learning rate schedule'' is not robust across LR schedules or
training regimes. More generally, our framework offers a principled
alternative for determining optimal data placement that accounts for
optimizer dynamics.

\section{Further details on 3.9B CPT experiments}
\label{sec:app_demo}

\subsection{Experimental details}

\cref{tab:model_info} and \cref{tab:train_steps} provide high-level
architectural and dataset details for the 3.9B model experiments.
This model generally follows the experimental setup of
\cref{sec:experimental_details}, with a few noteworthy exceptions
detailed here.
The 3.9B model is a GPT2-style LLM~\citep{radford2019gpt2} using
ALiBi~\citep{press2022alibi} position embeddings and a
squared-ReLU~\citep{so2021searching} activation function.
As shown in \cref{tab:model_info}, it employs an unusually wide FFN
dimension (8$\times \dmodel$), which we found to be effective in early
experiments.

Both pre-training and CPT phases use the AdamW optimizer with
$\beta_1=0.9$, $\beta_2=0.95$, and $\epsilon=10^{-16}$. The
pre-training warmup spans 375M tokens (not 10\% of total tokens),
followed by linear decay to zero learning rate---also used in
the CPT phase. Training is parameterized via the CompleteP variant of
$\mup$~\citep{dey2025dont}, with hyperparameters tuned using a
depth-32 proxy model.
The model uses a context length of 8192 tokens and a batch size of 672
sequences. Weight decay is set to $\lambda = 7.9 \times 10^{-4}$ based
on a projection for optimal $\tema$ following
\citet{bergsma2025power}. We also apply the layerwise weight decay
correction from \citet[Table~1]{dey2025dont}.
The peak learning rate during pre-training is $\eta = 0.15$ (selected
via proxy tuning), and we compare three different CPT learning rates
(10\%, 3\%, and 1\% of the peak PT learning rate); the specific
results for each are noted in the corresponding figures.

\begin{table}[]
\centering
  \caption{SlimPJ and other sources used in training the 3.9B model.
    General Blend (GB) is used in pre-training and continual
    pre-training phases, while Math Blend (MB), serving as
    high-quality data via up-weighting of OpenWebMath, is evaluated
    after placing it at particular locations during
    CPT.\label{tab:mathblend}}
\begin{tabular}{@{}ccc@{}}
\toprule
Subset         & General Blend (GB) & Math Blend (MB) \\ \midrule
GitHub         & 3.83\%             & 3.07\%          \\
Books          & 3.62\%             & 2.89\%          \\
ArXiv          & 4.21\%             & 3.37\%          \\
Wikipedia      & 3.16\%             & 2.53\%          \\
StackExchange  & 2.67\%             & 2.14\%          \\
Fineweb-Edu    & 64.75\%            & 51.80\%         \\
Cosmopedia     & 4.66\%             & 3.73\%          \\
OpenWebMath    & 1.88\%             & ~~28.18\%$^{\uparrow}$         \\
UltraTextBooks-2.0 & 0.42\%             & ~~2.29\%$^{\uparrow}$          \\
StarCoder      & 10.79\%            & 0.0\%           \\ \bottomrule
\end{tabular}
\end{table}

The data blends used in these experiments are listed in
\cref{tab:mathblend}. The general blend (GB) is used exclusively
during pre-training and as background data in CPT.\@ The math blend
(MB), which heavily up-weights OpenWebMath, serves as our designated
high-quality (HQ) data and is inserted at different positions during
CPT to assess TREC-guided placement strategies, as noted in
\cref{sec:demo}.  The MB phase is 234 steps, which comprise the final
234 steps of the 3303-step CPT phase when placing at the \emph{End},
while placement at \emph{Half} starts at step 1980.  Placement results
were given previously in \cref{fig:celerity_barchart}.

\subsection{Effect of learning rate on TREC loss during CPT}

\begin{figure}[ht]
    \centering
    \begin{minipage}{0.33\textwidth}
        \includegraphics[trim={0.3cm 0.4cm 0.264cm 0.3cm}, clip, width=\linewidth]{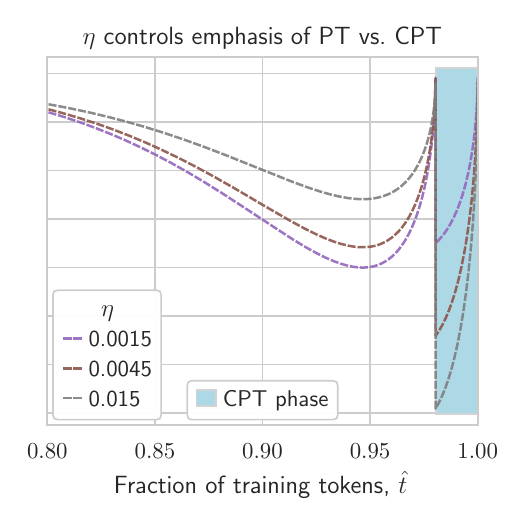}
    \end{minipage}
    \hspace{\twofighspace}
    \begin{minipage}{0.33\textwidth}
        \includegraphics[trim={0.3cm 0.4cm 0.264cm 0.3cm}, clip, width=\linewidth]{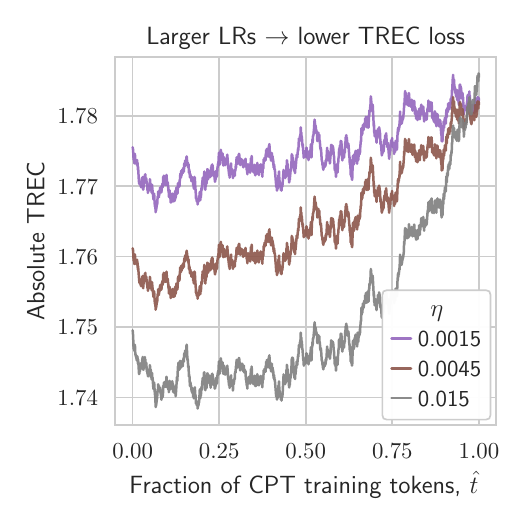}
    \end{minipage}
    \caption{\textbf{3.9B CPT: higher $\eta$ in CPT leads to lower
        TREC loss.} $\Leftfig$: Predicted TRECs for combined
      PT+CPT (zoomed into final 20\%).  $\Rightfig$: Actual TRECs
      in CPT phase: $\eta = 0.015$ has largest drop.
      \label{fig:celerityreeval}}
\end{figure}

\cref{fig:celerityreeval} ($\leftfig$) shows the predicted TRECs for
the 3.9B model (PT+CPT), zoomed-in to show the final portion of
pre-training and including the CPT phase (shown in blue shading).
The depth of the TREC valley varies with the CPT learning rate: as
$\eta$ increases, the minimum TREC loss becomes deeper.
For both SFT and CPT contexts, we hypothesize that TREC predictions
can help suggest the optimal post-training \emph{learning rate} (or at
least, the optimal range over which LR should be swept).  We are
exploring this further in ongoing work.

The observed TRECs from the actual training runs
(\cref{fig:celerityreeval}, $\rightfig$) match the predicted
trajectories (in blue shaded area) closely.

Interestingly, although $\eta = 0.015$ produces the deepest TREC
valley, it results in the \emph{worst} validation performance
(\cref{fig:celerity_barchart}). This illustrates a key failure mode
for \cref{hyp:crossschedule}: while TREC-guided placement is robust
\emph{within} a given training schedule, it does not reliably
generalize \emph{across} learning rate schedules.

\begin{figure}[ht]
    \centering
    \begin{minipage}{0.33\textwidth}
        \includegraphics[trim={0.3cm 0.4cm 0.264cm 0.3cm}, clip, width=\linewidth]{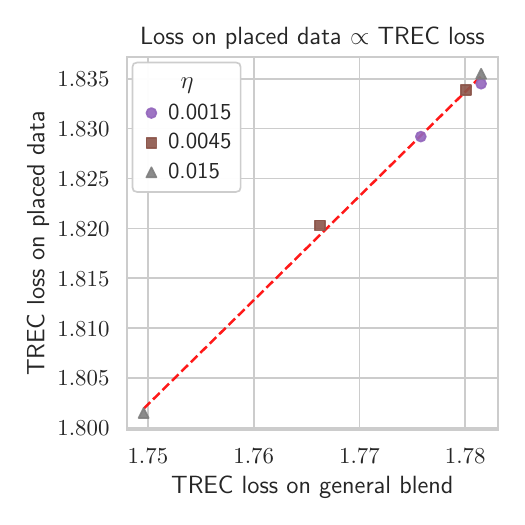}
    \end{minipage}
    \hspace{\twofighspace}
    \begin{minipage}{0.33\textwidth}
        \includegraphics[trim={0.3cm 0.4cm 0.264cm 0.3cm}, clip, width=\linewidth]{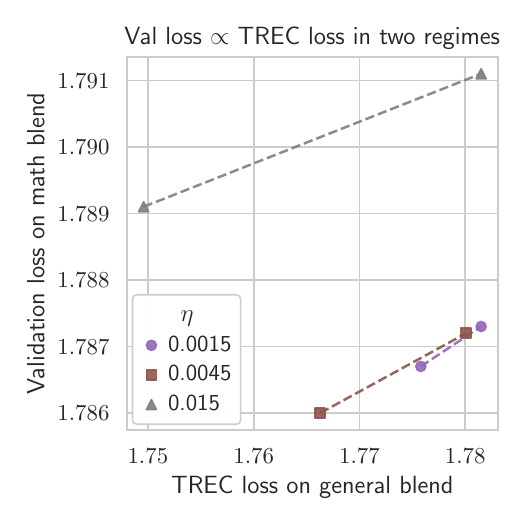}
    \end{minipage}
    \caption{\textbf{3.9B CPT: Predictive power of TRECs \emph{across}
        LR schedules.}
      $\Leftfig$: TREC loss on positions in general blend (training
      without placement) correlates very well with TREC loss on the
      \emph{inserted/training} math blend (i.e., when this math blend
      data is inserted at those positions, i.e., \emph{Half} and
      \emph{End} placement).
      $\Rightfig$: TREC loss on positions in general blend
      correlates with \emph{validation} loss on math blend data (i.e.,
      when separate training math blend data is inserted at those
      positions)---when $\eta <= 0.0045$; large $\eta$ seems to induce
      another TREC loss regime, with different correlation with
      validation performance.
      \label{fig:celeritycorrs}}
\end{figure}

We investigate this issue further in \cref{fig:celeritycorrs}, where
we now also assess the TREC losses on the \emph{placed data
segments}---i.e., the HQ data that was positioned at specific CPT
points (either halfway through CPT or at the end).
\cref{fig:celeritycorrs} ($\leftfig$) shows that the TREC loss on
the general blend (in the same segment) correlates nearly perfectly
with the loss on the placed math blend, regardless of the LR or where
the data was inserted. This confirms that TREC behavior generalizes
from \emph{homogeneous} to \emph{heterogeneous} data schedules: the
curve retains its predictive structure even when the inserted data
differs in content.

Given this strong correlation, the critical question becomes whether
performance on the \emph{training} HQ data (math blend) generalizes to
the \emph{validation} HQ data. This is assessed in
\cref{fig:celeritycorrs}, $\rightfig$.
Here, we observe two distinct regimes. For CPT runs using lower peak
learning rates ($\eta \leq 0.0045$), there is a strong linear
relationship between TREC loss and validation loss on the math
blend. However, at $\eta = 0.015$, validation performance deviates
from this trend entirely.

We speculate that high learning rates may push the model into a
different region of the optimization landscape---possibly one that
emphasizes memorization or shallower features---disrupting the
correspondence between training and validation loss.
This phenomenon is intriguing and merits further study in future work.

\section{Theoretical analysis of TRECs}\label{sec:theory}
\subsection{Motivation and key concepts}

\Cref{sec:predict} showed that TREC shapes are predicted well
by the AdamW EMA coefficients, but only after adjusting for training
fraction.  Why is this adjustment needed?

Our key hypothesis is that the \emph{effectiveness} of a gradient
update depends on \emph{where in parameter space} the update is
implicitly applied.  An update that was useful \emph{when computed}
may be less effective later in training if the local minimum for its
batch has shifted.  We argue that the pace of this minimizer drift is
largely scale-invariant and governed by the training fraction.

\newcommand{\xgrad}{-1.4}
\pgfmathsetmacro{\grad}{2.4*\xgrad*((\xgrad)^2 - 1) + 0.75*cos(deg(3*\xgrad)) - 0.2}
\def\dx{0.27}
\pgfmathsetmacro{\dy}{\grad*\dx}

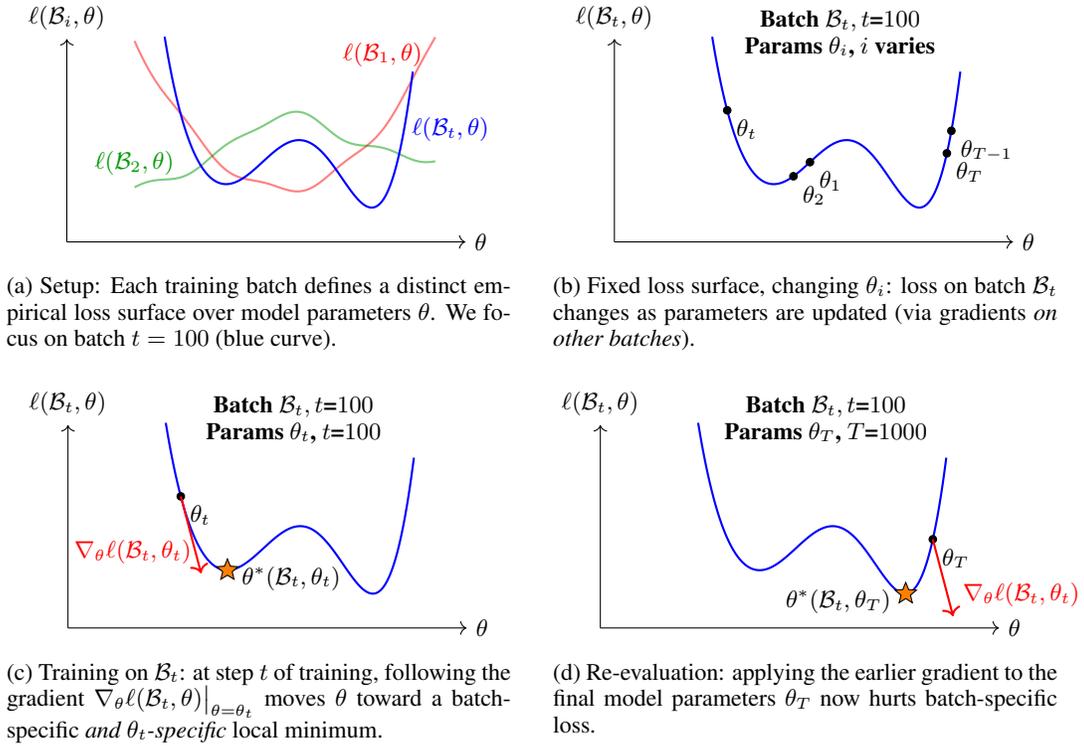
\begin{figure}[ht]
  \centering

\begin{subfigure}[t]{0.48\linewidth}
  \centering
  \begin{tikzpicture}[every node/.style={font=\small}]

    \draw[->] (-2.9, -0.7) -- (2.4, -0.7) node[right] {$\theta$};
    \draw[->] (-2.9, -0.7) -- (-2.9, 2.0) node[above] {$\ell(\mathcal{B}_i, \theta)$};

    \draw[thick, opacity=0.5, red, domain=-2.0:2.0, samples=100, smooth]
      plot(\x, {0.5*(\x)^2 - 0.05*sin(deg(6*\x))});
    \node[red] at (1.3, 1.8) {$\ell(\mathcal{B}_1, \theta)$};

    \draw[thick, opacity=0.5, green!60!black, domain=-2.0:2.0, samples=100, smooth]
      plot(\x, {1.0/(1 + (\x)^2) + 0.1*\x + 0.05*sin(deg(6*\x))});
    \node[green!60!black] at (-2.0, 0.35) {$\ell(\mathcal{B}_2, \theta)$};

    \draw[thick, blue, domain=-1.6:1.7, samples=100, smooth]
      plot(\x, {0.6*(\x*\x - 1)^2 + 0.25*sin(deg(3*\x)) - 0.2*\x});
    \node[blue] at (2.2, 0.8) {$\ell(\mathcal{B}_t, \theta)$};

  \end{tikzpicture}
  \caption{Setup: Each training batch defines a distinct empirical
    loss surface over model parameters $\theta$.  We focus on batch
    $t=100$ (blue curve).\label{fig:loss_surface:a}}
\end{subfigure}
  \hfill
  \begin{subfigure}[t]{0.48\linewidth}
    \centering
    \begin{tikzpicture}[every node/.style={font=\small}]

      \draw[->] (-2.9, -0.7) -- (2.4, -0.7) node[right] {$\theta$};
      \draw[->] (-2.9, -0.7) -- (-2.9, 2.0) node[above] {$\ell(\mathcal{B}_t, \theta)$};

      \draw[thick, blue, domain=-1.6:1.7, samples=100, smooth]
        plot(\x, {0.6*(\x*\x - 1)^2 + 0.25*sin(deg(3*\x)) - 0.2*\x});

      \def\xval{-0.3}
      \pgfmathsetmacro\yval{0.6*((\xval)^2 - 1)^2 + 0.25*sin(deg(3*\xval)) - 0.2*\xval}
      \filldraw[black] (\xval, \yval) circle (0.05);
      \node[below right] at (\xval, \yval) {$\theta_1$};

      \def\xval{-0.52}
      \pgfmathsetmacro\yval{0.6*((\xval)^2 - 1)^2 + 0.25*sin(deg(3*\xval)) - 0.2*\xval}
      \filldraw[black] (\xval, \yval) circle (0.05);
      \node[below right] at (\xval, \yval) {$\theta_2$};

      \def\xval{-1.4}
      \pgfmathsetmacro\yval{0.6*((\xval)^2 - 1)^2 + 0.25*sin(deg(3*\xval)) - 0.2*\xval}
      \filldraw[black] (\xval, \yval) circle (0.05);
      \node[below right] at (\xval, \yval) {$\theta_t$};

      \def\xval{1.58}
      \pgfmathsetmacro\yval{0.6*((\xval)^2 - 1)^2 + 0.25*sin(deg(3*\xval)) - 0.2*\xval}
      \filldraw[black] (\xval, \yval) circle (0.05);
      \node[below right] at (\xval, \yval) {$\theta_{T-1}$};

      \def\xval{1.52}
      \pgfmathsetmacro\yval{0.6*((\xval)^2 - 1)^2 + 0.25*sin(deg(3*\xval)) - 0.2*\xval}
      \filldraw[black] (\xval, \yval) circle (0.05);
      \node[below right] at (\xval, \yval) {$\theta_T$};
      
      \node at (0.1, 2.06) {\begin{tabular}{c}\textbf{Batch $\mathcal{B}_t, t$=$100$}\\\textbf{Params $\theta_i$, $i$ varies}\end{tabular}};
    \end{tikzpicture}
    \caption{Fixed loss surface, changing $\theta_i$: loss on batch $\mathcal{B}_t$ changes as parameters are updated (via gradients \emph{on other batches}).\label{fig:loss_surface:b}}
  \end{subfigure}

  \vspace{1em} 
  \begin{subfigure}[t]{0.48\linewidth}
    \centering
    \begin{tikzpicture}[every node/.style={font=\small}]

      \draw[->] (-2.9, -0.7) -- (2.4, -0.7) node[right] {$\theta$};
      \draw[->] (-2.9, -0.7) -- (-2.9, 2.0) node[above] {$\ell(\mathcal{B}_t, \theta)$};

      \draw[thick, blue, domain=-1.6:1.7, samples=100, smooth]
        plot(\x, {0.6*(\x*\x - 1)^2 + 0.25*sin(deg(3*\x)) - 0.2*\x});

      \def\xval{-1.4}
      \pgfmathsetmacro\yval{0.6*((\xval)^2 - 1)^2 + 0.25*sin(deg(3*\xval)) - 0.2*\xval}
      \filldraw[black] (\xval, \yval) circle (0.05);
      \node[below right] at (\xval, \yval) {$\theta_t$};

      \draw[->, thick, red] (\xval, \yval) -- ({\xval + \dx}, {\yval + \dy}) node[above left] {$\nabla_{\theta} \ell(\mathcal{B}_t, \theta_t)$};

      \def\xleft{-0.78}
      \pgfmathsetmacro\yleft{0.6*((\xleft)^2 - 1)^2 + 0.25*sin(deg(3*\xleft)) - 0.2*\xleft}
      \node[star, star points=5, star point ratio=2.25, fill=orange, draw=black, scale=0.4] at (\xleft, \yleft) {};
      \node[anchor=west, xshift=2pt, yshift=-3pt] at (\xleft, \yleft) {$\theta^*(\mathcal{B}_t, \theta_t)$};

      \node at (0.1, 2.06) {\begin{tabular}{c}\textbf{Batch $\mathcal{B}_t, t$=$100$}\\\textbf{Params $\theta_t$, $t$=$100$}\end{tabular}};
    \end{tikzpicture}
    \caption{Training on $\mathcal{B}_t$: at step $t$ of training, following the gradient $\nabla_{\theta} \ell(\mathcal{B}_t, \theta)\big|_{\theta = \theta_t}$ moves $\theta$ toward a batch-specific \emph{and $\theta_t$-specific} local minimum.\label{fig:loss_surface:c}}
  \end{subfigure}
  \hfill
  \begin{subfigure}[t]{0.48\linewidth}
    \centering
    \begin{tikzpicture}[every node/.style={font=\small}]

      \draw[->] (-2.9, -0.7) -- (2.4, -0.7) node[right] {$\theta$};
      \draw[->] (-2.9, -0.7) -- (-2.9, 2.0) node[above] {$\ell(\mathcal{B}_t, \theta)$};

      \draw[thick, blue, domain=-1.6:1.7, samples=100, smooth]
        plot(\x, {0.6*(\x*\x - 1)^2 + 0.25*sin(deg(3*\x)) - 0.2*\x});

      \def\xval{1.52}
      \pgfmathsetmacro\yval{0.6*((\xval)^2 - 1)^2 + 0.25*sin(deg(3*\xval)) - 0.2*\xval}
      \filldraw[black] (\xval, \yval) circle (0.05);
      \node[below right] at (\xval, \yval) {$\theta_T$};

      \draw[->, thick, red] (\xval, \yval) -- ({\xval + \dx}, {\yval + \dy}) node[above right] {$\nabla_{\theta} \ell(\mathcal{B}_t, \theta_t)$};

      \def\xright{1.16}
      \pgfmathsetmacro\yright{0.6*((\xright)^2 - 1)^2 + 0.25*sin(deg(3*\xright)) - 0.2*\xright}
      \node[star, star points=5, star point ratio=2.25, fill=orange, draw=black, scale=0.4] at (\xright, \yright) {};
      \node[anchor=east, xshift=-2pt, yshift=-3pt] at (\xright, \yright) {$\theta^*(\mathcal{B}_t, \theta_T)$};

      \node at (0.1, 2.06) {\begin{tabular}{c}\textbf{Batch $\mathcal{B}_t, t$=$100$}\\\textbf{Params $\theta_T$, $T$=$1000$}\end{tabular}};
    \end{tikzpicture}
    \caption{Re-evaluation: applying the earlier gradient to the final
      model parameters $\theta_T$ now hurts batch-specific
      loss.\label{fig:loss_surface:d}}
  \end{subfigure}

  \caption{Illustration of how parameters and gradients affect
    batch-specific loss: (a) batch $\mathcal{B}_t$ defines its own
    unique loss surface $\ell(\mathcal{B}_t, \theta)$, with two local
    minima.  (b) Each step of training defines a new set of parameters
    $\theta_i$, which determines the batch-specific loss.  Now, the
    \emph{effectiveness} of a gradient update to parameters $\theta$
    depends on $\theta$'s current position: (c) during training, an
    update in the gradient direction (red arrow) is effective---it
    moves $\theta_t$ toward the local minimum $\theta^*(\mathcal{B}_t,
    \theta_t)$; (d) after training, when we re-evaluate
    $\mathcal{B}_t$, incorporating that same gradient update is
    \emph{ineffective}---it now moves us away from the (new) local
    optimum, $\theta^*(\mathcal{B}_t, \theta_T)$.  \emph{Updates from
    batches seen later in training, when $t$ is closer to $T$, are
    more likely to improve loss on batch
    $\batcht$.}\label{fig:loss_surface}}
\end{figure}

More formally, each training batch \( \mathcal{B}_t \), as an
empirical sample, defines a corresponding loss
surface \( \ell(\batcht, \theta) \) over model parameters $\theta$.
This surface is fixed by the batch itself (the data, the model
architecture, and the parameter space do not change), but our position
on the surface, $\theta_i$, changes over training.
\Cref{fig:loss_surface} illustrates: at $t=100$, a gradient on
\( \mathcal{B}_t \) moves $\theta$ toward that batch's minimum.  
At $T=1000$, reapplying the same update from a different position can
increase loss, because the optimal direction has changed.

This perspective follows naturally from the EMA view of AdamW (and
SGD): the final parameters \( \theta_T \) are a weighted sum of
earlier updates, effectively ``applying'' each update to the final
model state, even if the update was computed long before.  Batches
nearer the end of training tend to have gradients more aligned with
$\theta_T$, and hence the gradients on these batches retain more of
their original usefulness.  That is, when applied at step $T$, these
gradients are more effective in terms of lowering loss on their
original batches, which manifests as lower loss upon re-evaluation
(using parameters $\theta_T$).

In the remainder of this section, we formalize this idea using a
simple quadratic model that allows us to isolate the factors---EMA
weighting and minimizer drift---that govern TREC loss. We proceed
in three steps:
\begin{enumerate}
  \item Begin with SGD, showing how it accumulates local minimizers via EMA-like dynamics.
  \item Derive TREC loss as a function of EMA weights and minimizer drift.
  \item Extend to AdamW, where the EMA is over preconditioned gradient updates.
\end{enumerate}

We then explain why the training-fraction exponent in our predictive
form should be scale-invariant, and conclude by discussing how this
viewpoint aligns with the empirical evidence presented elsewhere in
the paper.

\subsection{Setup and preliminaries}

To gain insight into TREC loss, we adopt a simplified
analytical model similar in spirit to the work
of \citet{zhang2019algorithmic}, who derived closed-form convergence
and loss expressions for various optimizers under different batch
sizes. As in their setup, we assume the optimizer dynamics are
invariant to rotation and translation, allowing us to model loss as
locally quadratic and separable across dimensions.

Specifically, we assume that each training batch \( \mathcal{B}_t \)
defines an empirical loss surface \( L(\batcht, \boldsymbol{\theta})
\) over model parameters $\theta$. We approximate this surface as
\emph{locally quadratic} at the current point in training.\footnote{
While LLM training minimizes cross-entropy loss, it is common to
perform a local quadratic approximation, i.e., a second-order Taylor
expansion in the parameters, with the constant Hessian replaced by the
instantaneous Hessian along the training
trajectory~\citep{lecun1989optimal}.  Thus conclusions drawn from
quadratic models often generalize to large, realistic
networks~\citep{zhang2019algorithmic}.}
Let $\boldsymbol{\theta}^*(\batcht)$ denote the local minimum. In $D$
dimensions, the loss at step $t$, given model parameters
$\boldsymbol{\theta}_t$, is:
\begin{equation}
L(\batcht, \boldsymbol{\theta}_t) = \frac{1}{2} (\boldsymbol{\theta}_t
- \boldsymbol{\theta}^*(\batcht))^\top \cdot H_t \cdot (\boldsymbol{\theta}_t
- \boldsymbol{\theta}^*(\batcht))
= \sum_{d=1}^D \ell^{(d)}(\batcht, \theta_t^{(d)}),
\end{equation}
where \( H_t = \mathrm{diag}(h_t^{(1)}, \ldots, h_t^{(D)}) \) is a
diagonal positive semi-definite Hessian (reflecting the batch-specific
loss curvature), and
\begin{equation}
\ell^{(d)}(\batcht, \theta_t^{(d)}) = \frac{1}{2} \cdot h_t^{(d)} \cdot \left( \theta_t^{(d)} - \theta^{*,(d)}(\batcht) \right)^2
\end{equation}
is the per-dimension contribution to the loss.

\paragraph{The key idea: ``optimal'' parameters can change.}
To streamline the analysis, we now focus on a single dimension and
drop the superscript notation. Unlike \citet{zhang2019algorithmic}, we
explicitly consider that \textbf{the locally-optimal parameter
  $\theta^*$ for a batch $\mathcal{B}_t$ may depend on the model state
  at a given training step}, i.e., what's optimal depends on where we
are on the loss surface. We therefore denote this local optimum as
$\theta^*(\batcht, \theta_s)$: the local minimizer for batch
$\mathcal{B}_t$ as computed at step $s$; think of $\mathcal{B}_t$ as
defining the loss surface, while $\theta_s$ defines our current
position on it---and the local minimizer depends on our position. This
is illustrated in \cref{fig:loss_surface}, where the loss surface is
fixed, but the local minimizer is different at step \( s = 100 \)
(\cref{fig:loss_surface:c}) and step \( s = 1000 \)
(\cref{fig:loss_surface:d}), as our position on the loss surface
changes.  Furthermore, for simplicity, we assume the batch-specific
curvature \( h_t \) remains fixed across steps.

At each step of training, the loss is evaluated at the \emph{current}
model parameters \( \theta_t \) (our current position on the loss
surface), so the \emph{standard training loss curve} is:
\begin{equation}\label{eq:train_loss}
\ell(\batcht, \theta_t) = \frac{1}{2} \cdot h_t \cdot \left( \theta_t - \theta^*(\batcht, \theta_t) \right)^2.
\end{equation}

However, for \emph{TREC}, we compute loss using the
\emph{final} model parameters \( \theta_T \) (our final position on
the loss surface), i.e., using the same empirical loss surface, but
positioned near a (potentially) different local minimizer.  Thus the
\emph{TREC} is:
\begin{equation}\label{eq:reeval_loss}
\reeval(t) := \ell(\batcht, \theta_T) = \frac{1}{2} \cdot h_t \cdot \left( \theta_T - \theta^*(\batcht, \theta_T) \right)^2.
\end{equation}

We have endeavored to make this key concept clear because this
distinction---between what's optimal at the time a batch was seen
(\cref{eq:train_loss}) and what's optimal when re-evaluating it later
using the final model parameters (\cref{eq:reeval_loss})---is central
to understanding TREC dynamics. We will now examine how minibatch
SGD accumulates these position-specific minimizers and how temporal
shifts between \( \theta^*(\batcht, \theta_t) \) and \(
\theta^*(\batcht, \theta_T) \) affect the TREC loss.

\subsection{SGD: TREC loss and EMA coefficients}

We now derive the TREC loss under vanilla SGD and show how it relates to the EMA coefficients and shifts in local optima over time.

The gradient of the quadratic loss surface at step $t$ is:
\begin{equation}
\nabla_{\theta} \ell(\batcht, \theta_t) = h_t \cdot (\theta_t - \theta^*(\batcht, \theta_t)),
\end{equation}
and the parameter update under SGD is:
\begin{equation}
\theta_{t+1} = \theta_t - \eta_t \nabla_{\theta} \ell(\batcht, \theta_t) = \theta_t - \eta_t h_t \cdot (\theta_t - \theta^*(\batcht, \theta_t)),
\end{equation}
where $\eta_t$ is the learning rate at step $t$.

This can be rearranged as:
\begin{equation}\label{eq:sgd_ema_update}
\theta_{t+1} = (1 - \eta_t h_t) \cdot \theta_t + \eta_t h_t \cdot \theta^*(\batcht, \theta_t),
\end{equation}
which is equivalent to an exponential moving average (EMA):
\begin{equation}
y_t = (1 - \alpha_t) y_{t-1} + \alpha_t x_t,
\end{equation}
where $y_t = \theta_t$, $\alpha_t = \eta_t h_t$, and $x_t
= \theta^*(\batcht, \theta_t)$ (assuming the LR $\eta_t$ is chosen so
that the sum is stable, i.e., $0 \le \alpha_t \le 1$).

Just as in the AdamW EMA case (\cref{sec:predict}), we can unroll the
recursion and explicitly compute the contribution of each step to the
final parameter value:
\begin{equation}\label{eq:sgd_ema_unrolled}
\theta_T = \sum_{i=1}^T \left( \alpha_i \prod_{j=i+1}^T (1 - \alpha_j) \right) \theta_i^*(i) = \sum_{i=1}^T c_i \theta_i^*(i),
\end{equation}
where $c_i = \alpha_i \prod_{j=i+1}^T (1 - \alpha_j)$.

As noted above, we are primarily interested in the TREC loss
of batch $\batcht$, evaluated at the final model parameters
$\theta_T$, on a loss surface with minimizer
$\theta^*(\batcht, \theta_T)$ (\cref{eq:reeval_loss}).  Substituting
our EMA expression for $\theta_T$ into \cref{eq:reeval_loss}:
\begin{equation}
\ell(\batcht, \theta_T) = \frac{1}{2} \cdot h_t \cdot \left( \left[\sum_{i=1}^T c_i \theta_i^*(i)\right] - \theta^*(\batcht, \theta_T) \right)^2.
\end{equation}

To isolate the contribution from $i = t$:
\begin{equation}
\ell(\batcht, \theta_T) = \frac{1}{2} \cdot h_t \cdot \left( \left[\sum_{i \ne t} c_i \theta_i^*(i)\right] + c_t \theta^*(\batcht, \theta_t) - \theta^*(\batcht, \theta_T) \right)^2.
\end{equation}

In the special case where $\theta^*(\batcht, \theta_t)
= \theta^*(\batcht, \theta_T)$ (i.e., the local minimizer has not
changed):
\begin{equation}\label{eq:nodrift}
\ell(\batcht, \theta_T) = \frac{1}{2} \cdot h_t \cdot \left( \left[\sum_{i \ne t} c_i \theta_i^*(i)\right] + (c_t - 1) \theta^*(\batcht, \theta_T) \right)^2.
\end{equation}

The loss in \cref{eq:nodrift} is fully minimized when $c_t = 1$ and
all other $c_i = 0$.  More generally, the steps with the highest EMA
coefficient will obtain the lowest loss: with a static local
minimizer, the EMA coefficients fully define the TREC trajectory.

However, in practice, the local optimum for batch $t$ may drift over
the course of training. We model this by assuming:
\begin{equation}
\theta^*(\batcht, \theta_t) = r_t \cdot \theta^*(\batcht, \theta_T),
\end{equation}
with $r_t \in \mathbb{R}$ a scaling factor (possibly $<0$). The
TREC loss becomes:
\begin{equation}
\ell(\batcht, \theta_T) = \frac{1}{2} \cdot h_t \cdot \left( \left[\sum_{i \ne t} c_i \theta_i^*(i)\right] + (c_t r_t - 1) \theta^*(\batcht, \theta_T) \right)^2.
\end{equation}

This expression emphasizes that the TREC loss is minimized when \(
c_t r_t = 1 \) (and the sum over \( i \ne t \) vanishes), i.e., when
the model fully incorporates the local minimizer for batch \( t \) via
a high \( c_t \), \emph{and} this minimizer is well-aligned with the
one at the final step \( T \). This analytical result mirrors our
empirical finding (\cref{sec:predict}) that EMA coefficients alone are
insufficient to predict TREC loss: early gradients may have high
coefficients \( c_t \), but the local minimizers of their
corresponding batches can drift substantially over training (changing
$r_t$), leading to poor retention.

\subsection{AdamW: TREC loss and preconditioned updates}\label{subsec:adamw_analysis}

We obtain a similar result when analyzing AdamW with the same
locally-quadratic loss surface. As derived in \cref{sec:predict},
AdamW parameters at the final step \( T \) can be expressed as a
convex combination of weight \emph{updates}:
\begin{equation}
\theta_T = \sum_{i=1}^T c_i x_i,
\end{equation}
where each update \( x_i \) has the form:
\begin{equation}
x_t = -\frac{1}{\lambda} \cdot \frac{\hat{m}_t}{\sqrt{\hat{v}_t} + \epsilon},
\end{equation}
and each coefficient is defined as:
\begin{equation}\label{eq:adamw_coeffs_appendix}
c_i = \left( \prod_{j=i+1}^{T} (1 - \eta_j \lambda) \right) \eta_i \lambda.
\end{equation}

Substituting this decomposition into the TREC loss expression, we
obtain:
\begin{equation}\label{eq:adamw_reeval}
\ell(\batcht, \theta_T) = \frac{1}{2} \cdot h_t \cdot \left( \left[\sum_{i \ne t} c_i x_i\right] + c_t x_t - \theta^*(\batcht, \theta_T) \right)^2.
\end{equation}

TREC loss is minimized when \( c_t \) is large and the update \(
x_t \) points in the same direction as the final-step local
minimizer \( \theta^*(\batcht, \theta_T) \). In other words, it is primarily
the \emph{sign alignment} of \( x_t \) and \( \theta^*(\batcht, \theta_T) \) that
determines whether the update helps or hurts TREC loss.

In the quadratic model (and optimization invariant to translation), we
can assume without loss of generality that \( \theta_t = 0 \) at the
time of the update. The gradient becomes:
\begin{equation}
\nabla_{\theta} \ell(\batcht, \theta_t) = h_t \cdot (\theta_t - \theta^*(\batcht, \theta_t)) = -h_t \theta^*(\batcht, \theta_t),
\end{equation}
and assuming no momentum for simplicity, the update is:
\begin{equation}\label{eq:kappa_t}
x_t = \frac{1}{\lambda} \cdot \frac{h_t \theta^*(\batcht, \theta_t)}{\sqrt{v_t} + \epsilon} = \kappa_t \theta^*(\batcht, \theta_t),
\end{equation}
where \( \kappa_t \ge 0 \) absorbs the preconditioning, curvature, and
scaling terms.


To model drift in the loss surface, we again assume that the local
minimizer for batch \( t \) changes over time and satisfies:
\begin{equation}\label{eq:adamw_rt}
\theta^*(\batcht, \theta_t) = r_t \cdot \theta^*(\batcht, \theta_T).
\end{equation}
Substituting the change in local minimizer (\cref{eq:adamw_rt}) into
our simplified expression for the update (\cref{eq:kappa_t}), and
using this update in the TREC loss equation
(\cref{eq:adamw_reeval}) yields:
\begin{equation}
\ell(\batcht, \theta_T) = \frac{1}{2} \cdot h_t \cdot \left( \left[\sum_{i \ne t} c_i x_i\right] + (c_t \kappa_t r_t - 1) \theta^*(\batcht, \theta_T) \right)^2.
\end{equation}

This loss is reduced when the AdamW update contributes a value that
matches the current minimizer direction, where $r_t$ controls the
update direction (since $c_t \ge 0$ and $k_t \ge 0$). As before, loss
is maximally reduced when $c_t = 1$ and the weight on all other
updates vanishes.  Loss is also maximally reduced when the product \(
c_t \kappa_t r_t \approx 1 \). In practice, if the local minimizer has
shifted substantially---especially if \( r_t < 0 \)---then the early
update may push in the wrong direction, \emph{increasing} TREC loss
despite having a high EMA coefficient.

This again echoes the central insight: an update's value is determined
not only by its EMA weight \( c_t \), but by its alignment with
$\theta_T$'s position in the loss landscape.

\subsection{Local minimizer drift and the training-fraction clock}\label{sec:tf_clock}

In the preceding analysis, the term $r_t$ captures the effect of
\emph{local minimizer drift}---the change in the batch-specific optimum
$\theta^*(\mathcal{B}_t, \cdot)$ between when the gradient was computed
(at step $t$) and when it is effectively applied (in the final model
$\theta_T$).  
We now explain why the natural coordinate for this drift is the
\emph{training fraction} $\trainfrac = t/T$.

In the SGD case, the final parameters $\theta_T$ can be written as an
EMA with time-varying smoothing $\alpha_t = \eta_t h_t$, where
$\eta_t$ is the learning rate and $h_t$ the curvature along the
current parameter direction
(\cref{eq:sgd_ema_update,eq:sgd_ema_unrolled}).  When $\alpha_t$
values are larger, more weight is put on recent minimizers (shrinking
the EMA timescale), and parameters evolve faster from step to step.
In particular, the pace at which the final parameters $\theta_T$ move
away from earlier parameters $\theta_t$ is directly governed by the
sequence $\{\eta_s h_s\}_{s=t}^T$.  We naturally assume that larger
movement in \emph{parameters} coincides with larger movement
in \emph{local optimizers}, i.e., that the distance between
$\theta^*(\batcht, \theta_t)$ and $\theta^*(\batcht, \theta_T)$ is
also governed by this sequence.

In AdamW, the same qualitative picture holds: the learning rate enters
directly into the EMA coefficients $c_t$
(\cref{eq:adamw_coeffs_appendix}), while the curvature appears as a
scaling factor in the updates (\cref{eq:kappa_t}).  Although
preconditioning partly normalizes curvature variation, the product of
learning rate and effective curvature still governs the rate at which
parameters update, and hence the rate at which local minimizers drift.

Recent work by \citet{noci2024super} shows that, under
maximal update parameterization ($\mup$), curvature statistics such as
the largest Hessian eigenvalues evolve in a nearly
\emph{scale-independent} way, a phenomenon they term
\emph{super-consistency}.
\citet{qiu2025scaling} build on this observation, hypothesizing, and
providing empirical evidence, that a related curvature proxy depends
only on training fraction (what they term ``normalized compute'') and
is largely scale-independent under $\mup$.
That is, if we align two models of different sizes by their training
fraction $\trainfrac = t/T$, their curvature trajectories
$h(\trainfrac)$ are nearly identical.

If both the curvature $h(\trainfrac)$ and the learning rate schedule
$\eta(\trainfrac)$ are scale-independent functions of $\trainfrac$,
then their product $\eta(\trainfrac) h(\trainfrac)$---which governs
the pace of parameter drift in the quadratic model---is also
scale-independent.  Consequently, the sequence of
curvature/learning-rate conditions experienced by a small model at
$\trainfrac = 0.3$ is essentially the same as that experienced by a
large model at $\trainfrac = 0.3$.
Based on the prior work in $\mup$ and our own empirical results
(discussed below, \cref{sec:empirical-m}), we speculate that in more
complex settings, other aspects of training dynamics also evolve with
training fraction.
For example, in $\mup$ neural networks trained with cross-entropy
loss, the extent of feature learning and the evolution of
representations in earlier layers should also evolve invariant to
model size.
This yields a \emph{training-fraction clock} for minimizer drift:
relative training progress, not absolute tokens, is the natural
coordinate for $r_t$ in re-evaluation dynamics.  In this view, $r_t$
should be written as $r(\trainfrac)$.

\subsection{From curvature to functional form}
Our quadratic-model analysis suggests that the product
$\eta(\trainfrac)h(\trainfrac)$---learning rate times curvature---sets
the magnitude of parameter updates, and hence the rate at which local
minimizers move.  A gradient computed at training fraction
$\trainfrac$ will remain well aligned with the final model parameters
$\theta_T$ only if the subsequent minimizer motion between
$\trainfrac$ and the end of training is small.

Conceptually, we can think of a \emph{drift rate} at each point in
training, proportional to $\eta(\trainfrac)h(\trainfrac)$, which
measures how fast the optimizer ``forgets'' earlier minimizers.  The
cumulative drift that erodes the utility of a gradient from fraction
$\trainfrac$ is then the integral of this rate over the
\emph{remaining} training,
\[
r(\trainfrac) \;\propto\; \int_{\trainfrac}^1 \!\eta(s) h(s) \, ds.
\]
Empirically, we find that $r(\trainfrac)$ is well fit by a power law
$\trainfrac^m$, as used in our predictive form for the TREC
(\cref{eqn:prediction}):
\[
\hat{\reeval}(\trainfrac) = 1 - c(\trainfrac)^p \, \trainfrac^m,
\]
where $c(\trainfrac)$ is the AdamW EMA coefficient and $m$ captures
the cumulative effect of curvature- and LR-driven drift along the
training-fraction clock.

\subsection{Empirical evidence for variation in $m$}\label{sec:empirical-m}

Interpreting the $\trainfrac^m$ term as a cumulative drift means that
larger $m$ values (higher drift) correspond to gradients losing
effectiveness more quickly; higher drift affects the TRECs by
increasing the span of ``forgotten'' data---i.e., the data with
baseline higher TREC loss (regardless of $c(\trainfrac)$
coefficient).

From our analysis, $m$ should change under interventions that modify
either the curvature/learning-rate product $\eta(\trainfrac)
h(\trainfrac)$ or the optimizer's effective memory.  Consistent with
this prediction, we observe systematic variation in $m$ across
multiple settings presented earlier in the paper:

\paragraph{Lower EMA timescale.}
Shortening the AdamW EMA timescale $\tema$ decreases the optimizer's
effective memory, and therefore increases the rate-of-change of
parameters.  This should lead to faster reductions in gradient
effectiveness, with an effect on TRECs independent of the separate,
analytical effect of a smaller $\tema$ on the EMA coefficients
themselves.  Indeed, we observe higher optimal $m$ values for lower
timescales in our experiments (e.g., \cref{fig:fits}).  Moreover, our
$\mstar$ power law fitted at 111M scale predicts well across model
sizes: the $\tema$-governed drift is scale-invariant.

\paragraph{Higher tokens-per-parameter (TPP).}
We also see a small but consistent increase in optimal $m$ when
increasing the TPP ratio (also shown in \cref{fig:fits}).  Note the
normalized LR schedule $\eta(\trainfrac)$ does not change with
TPP.\@ We therefore hypothesize that the degree of overtraining (as
defined by TPP) slightly affects the pace of change in curvature (as a
function of training fraction), which drives the pace that earlier
gradients become ineffective.  Yet it remains notable that these
changes in $m$ are solely TPP-dependent and do not depend on the
absolute number of training steps (which scales with model size for a
given TPP).

\paragraph{Fewer MoE experts.}
In sparse Mixture-of-Experts (MoE) models, \emph{reducing} the number
of experts \emph{increases} the number of tokens processed by each
expert's parameters, raising the expert's effective TPP.\@ The impact
on $m$ closely matches the shift observed when directly changing TPP
in dense models, suggesting that MoE layers inherit the same drift
scaling in TPP as non-expert dense layers.

\paragraph{CPT vs.\ from-scratch training.}
Recall \cref{fig:pt_cpt_test} comparing two 3.9B models trained with
identical configurations---same data, batch size, weight decay, and an
identical learning-rate schedule $\eta(t)$---but different starts: one
from random initialization and one from a near-optimal checkpoint
(i.e., this variant undergoes continual pre-training, CPT). Although
$\eta(\trainfrac)$ and EMA coefficients $c(\trainfrac)$ were identical
in both runs, the two trajectories occurred at different points in
training fraction $\trainfrac$ and thus likely under different
curvature statistics $h(\trainfrac)$.
Beyond curvature variation, changes in the extent of feature learning
and evolving representations in earlier layers may also play a
role. In a deep network, early updates are large compared to weight
magnitudes (despite preconditioning)~\citep{kosson2024analyzing}, and
thus optimization steps can produce disproportionately large shifts in
the parameters, and thus the optimal batch minimizers.  Empirically,
the CPT model, starting in a region already well aligned with the
final parameters, exhibited little minimizer drift: its $\trainfrac^m$
term remains close to~1, producing a TREC prediction closely
following the EMA coefficients. In contrast, the scratch model
traversed a much larger region of the loss landscape, with substantial
drift that reduced alignment to early updates; its $\trainfrac^m$ term
rises gradually from 0 to~1, with a corresponding TREC valley only
aligning to the EMA coefficients near the very end of training.

\paragraph{WSD vs. Cyclic LR schedules.}
In \cref{sec:wsd_cyclic_alignment}, we compared the $\wsd$ and the
$\cyclic$ schedules.  Training runs with the two schedules completely
align in step-wise LR, weight decay, and batch size---in the final
20\% of training.  In a way, this is a similar test to the CPT
vs. from-scratch comparison above: two similar training runs, but
initializing from different start points.
In this case, however, the two models have undergone a similar total
amount of training during their initial phases, and consequently the
initial gradients in the final phase have similar alignment with the
final model parameters.  Consequently, the final period spans the same
range of training fraction, and their TRECs align very well
over this final period.
For the same peak LR, the $\wsd$ schedule does appear to drop slightly
lower at around, e.g., $\trainfrac = 0.8$, which may reflect a more
mature model at that stage (perhaps because this schedule, unlike
$\cyclic$, does not have the periods where LR returns to zero).

\paragraph{Different $\mstar$ for different learning rate schedules.}

While the minor differences between $\wsd$ and $\cyclic$ are unlikely
to be consequential in terms of data placement strategy, differences
in the final stages of training have a greater impact.

As we noted in \cref{sec:lr_schedule_generalization}, optimal $m$
tends to change when the learning rate decay pattern changes
(schedule-specific fits yield the best predictive performance).  We
also separately observed that $\constant$ learning rate schedules
produce the largest optimal $m$ values. This is consistent with our
drift interpretation: without decay, $\eta(\trainfrac)$ remains high
even late in training, inducing substantial movement in $\theta_T$ and
its associated local minimizer.  Earlier gradients are therefore
misaligned more quickly, requiring a larger $m$ to match the observed
TRECs.

Across these cases, the observed shifts in $m$ are consistent with
changes in the effective $\eta(\trainfrac) h(\trainfrac)$ product.
Together, these results suggest that a functional form may exist for
local minimizer drift that not only normalizes across scale (as does
the training fraction term), but also across LR schedules.  That is,
the functional form for this drift could directly incorporate the LR
schedule itself.


\end{document}